\documentclass[journal,twoside,web]{ieeecolor}
\usepackage{generic}
\usepackage{cite}
\usepackage{amsmath,amssymb,amsfonts}
\usepackage{algorithmic}
\usepackage{graphicx}
\usepackage{algorithm,algorithmic}
\usepackage{xr-hyper}
\makeatletter
\newcommand*{\addFileDependency}[1]{
  \typeout{(#1)}
  \@addtofilelist{#1}
  \IfFileExists{#1}{}{\typeout{No file #1.}}
}
\makeatother
\newcommand*{\myexternaldocument}[1]{%
    \externaldocument{#1}%
    \addFileDependency{#1.tex}%
    \addFileDependency{#1.aux}%
}
\usepackage{hyperref}
\hypersetup{hidelinks}
\usepackage{textcomp}
\usepackage{caption}
\usepackage{subcaption}
\usepackage{textcomp}
\usepackage{multirow}
\usepackage{booktabs}
\usepackage{bbding}
\usepackage{caption}
\usepackage{makecell}
\usepackage{url}
\usepackage[table]{xcolor}
\def\BibTeX{{\rm B\kern-.05em{\sc i\kern-.025em b}\kern-.08em
    T\kern-.1667em\lower.7ex\hbox{E}\kern-.125emX}}
\myexternaldocument{../supplementary/supplementary}

\begin{document}
\title{MCSeg: Pre-training and Fine-tuning Volumetric Pyramid Transformer for Multi-modal Cardiac Image Segmentation}
\author{Zhiyu Ye, Hairong Zheng, \IEEEmembership{Senior Member, IEEE}, and Tong Zhang, \IEEEmembership{Member, IEEE}
\thanks{The manuscript is first submitted on 14 November 2024.}
\thanks{This work was supported in part by the PCL Major Key Project (grant PCL2026A01-2 and PCL2025A09-1), and the Guangdong Provincial Science and Technology Department under Grant No. 2024A1515010245.}
\thanks{Z. Ye is with State Key Laboratory of Biomedical Imaging Science and System, Shenzhen Institute of Advanced Technology, Chinese Academy of Sciences, Shenzhen, China, Pengcheng Laboratory, Shenzhen, China, and University of Chinese Academy of Sciences, Beijing, China. (e-mail: yezhy@pcl.ac.cn)}
\thanks{H. Zheng is with State Key Laboratory of Biomedical Imaging Science and System, Shenzhen Institute of Advanced Technology, Chinese Academy of Sciences, Shenzhen, China, and University of Chinese Academy of Sciences, Beijing, China. (e-mail: hr.zheng@siat.ac.cn, corresponding author)}
\thanks{T. Zhang is with Pengcheng Laboratory, Shenzhen, China. (e-mail: zhangt02@pcl.ac.cn, corresponding author)}
}

\maketitle

\begin{abstract}
Automatic cardiac image segmentation is pivotal for diagnosing and treating cardiac diseases. 
In this work, we introduce MCSeg, a volumetric transformer-based network tailored for multi-modal cardiac segmentation. 
To overcome the architectural mismatch inherent in existing hybrid networks, we propose a novel Scaling Feature Pyramid (SFP). Unlike conventional skip connections, the SFP effectively bridges the single-scale 3D Vision Transformer (ViT) encoder and the multi-scale CNN decoder by transforming the ViT's output into a hierarchical feature pyramid, ensuring that global contextual information is effectively leveraged.
For the training paradigm, the ViT encoder first undergoes self-supervised pre-training via masked image modeling. Subsequently, the network is fine-tuned on downstream tasks, during which a regional mutual information (RMI) loss is integrated to improve boundary segmentation accuracy.
In experiments, MCSeg consistently outperforms eleven SOTA methods on CT dataset ImageCHD, multi-modal dataset MM-WHS, MRI dataset HVSMR-2.0 and MSD Heart, highlighting the effectiveness of our MCSeg for multi-modal cardiac segmentation tasks.
Furthermore, MCSeg's superior performance in few-shot experiment showcases its significant potential in adapting to limited data scenarios. Codes and pre-trained ViT-B weights are open-sourced at \url{https://openi.pcl.ac.cn/OpenMedIA/MCSeg}.
\end{abstract}


\section{Introduction}
\label{sec:introduction}
\IEEEPARstart{M}{edical} image segmentation, an important application of artificial intelligence in healthcare~\cite{wang2022medical}, has been extensively researched, encompassing various organs and diseases~\cite{bilic2023liver, heller2021state, wang2018automatic, menze2014multimodal}.
Automated segmentation of cardiac images is a crucial step for non-invasive assessment of cardiac structures and functions. It offers indispensable support for diagnosis, disease monitoring, treatment planning, and prognosis~\cite{chen2020deep}.

\begin{figure}[!t]
\centering
\begin{minipage}{\linewidth}
    \centering
    \tabcolsep=0.02cm
    \scriptsize
    \begin{tabular}{cccccc}
     \makecell[c]{Image} & \makecell[c]{GT} & \makecell[c]{MCSeg \\ Dice=0.8474} &  \makecell[c]{CardiacSeg \\ Dice=0.7488}&\makecell[c]{SwinUNETR \\ Dice=0.7531} & \makecell[c]{UNETR \\ Dice=0.8174}
    \vspace{-0.02in} \\ 
    \makecell[c]{\includegraphics[width=0.15\linewidth]{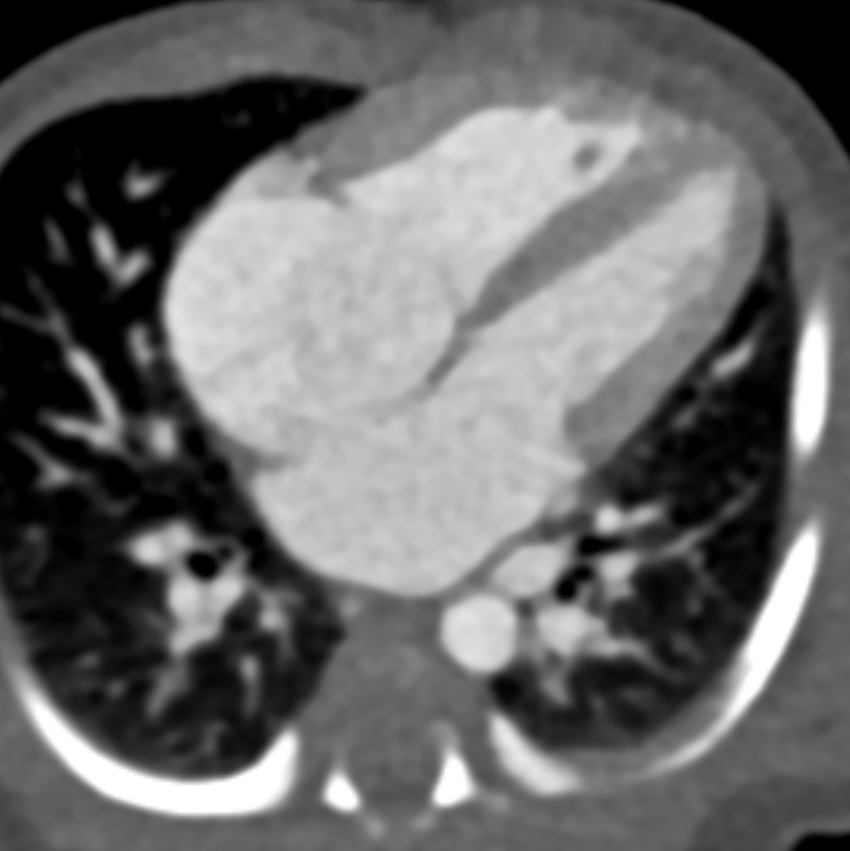}}
    & \makecell[c]{\includegraphics[width=0.15\linewidth]{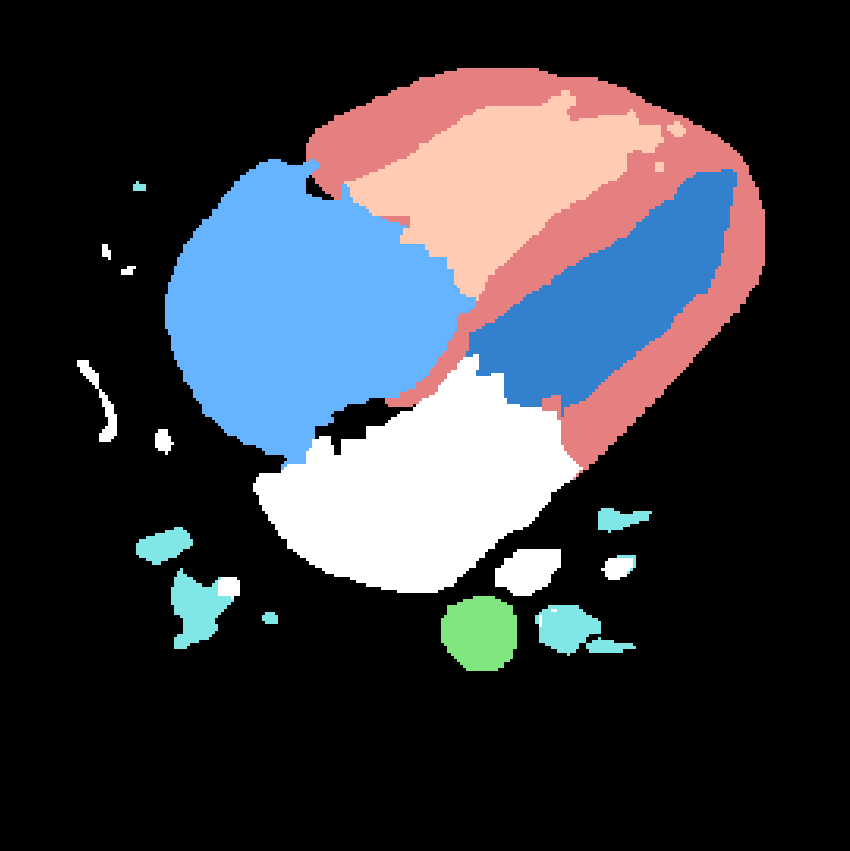}}
    & \makecell[c]{\includegraphics[width=0.15\linewidth]{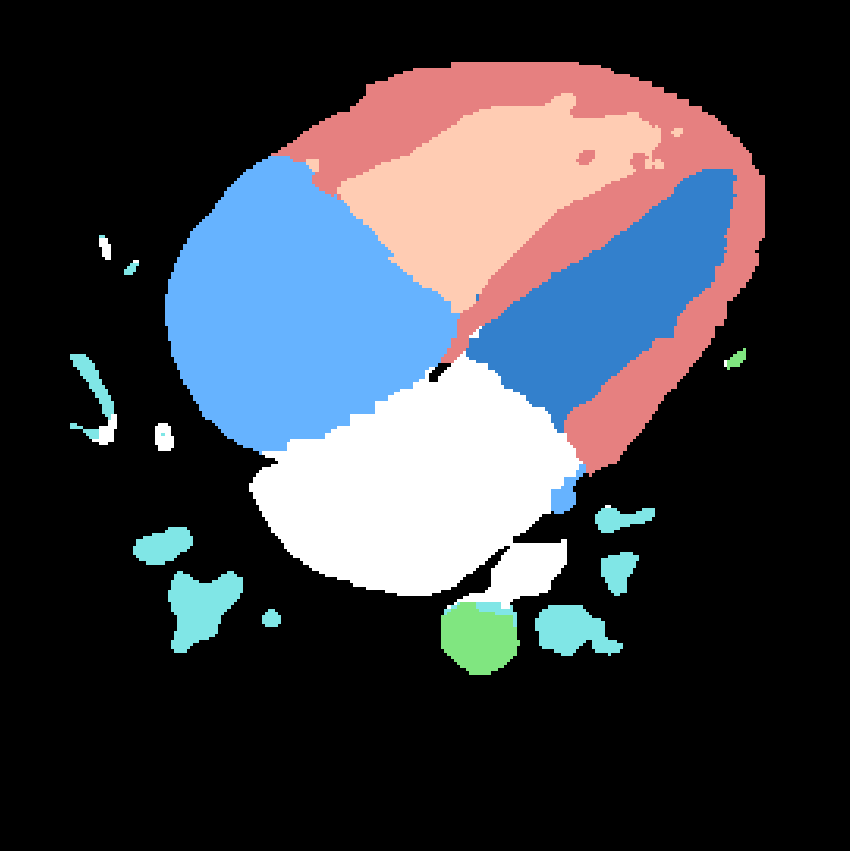}}
    &  \makecell[c]{\includegraphics[width=0.15\linewidth]{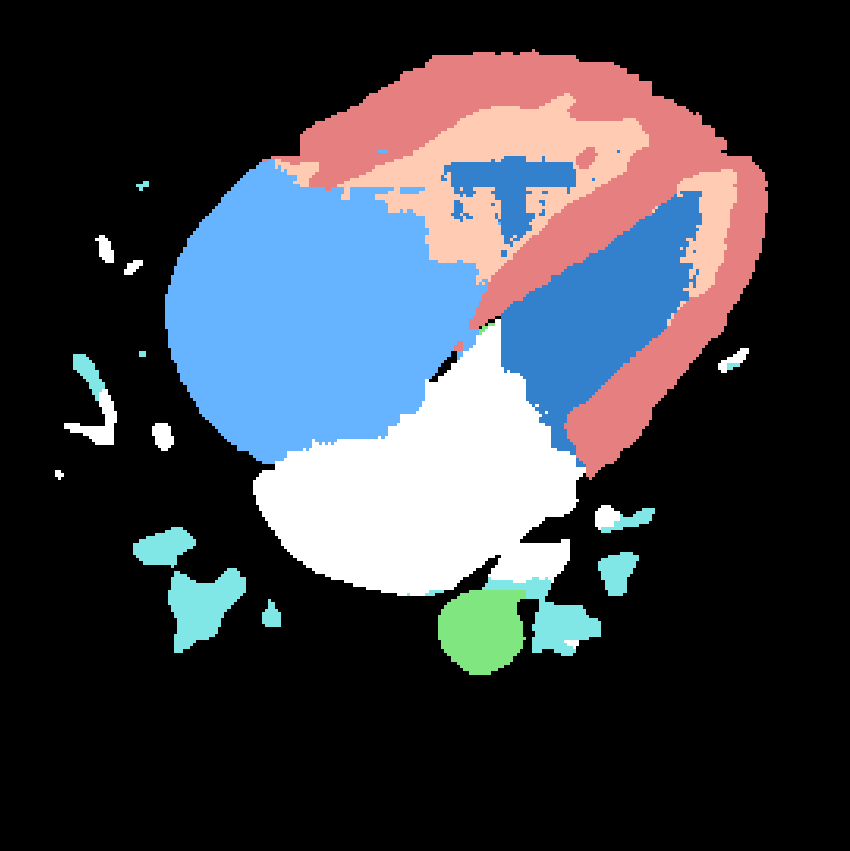}} 
    &\makecell[c]{\includegraphics[width=0.15\linewidth]{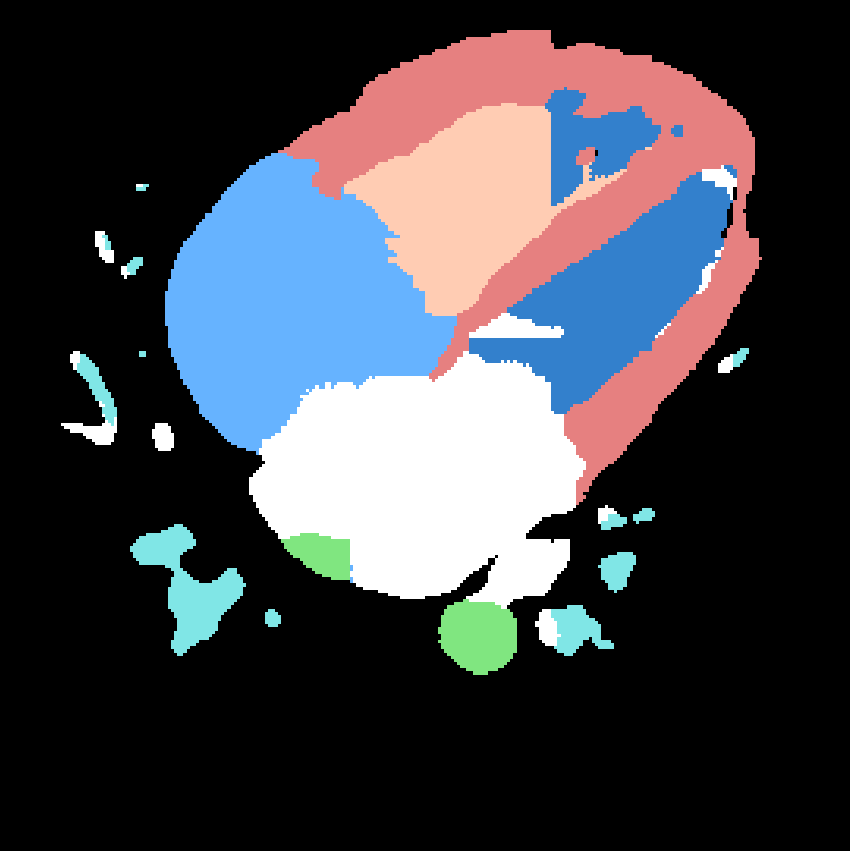}} 
    & \makecell[c]{\includegraphics[width=0.15\linewidth]{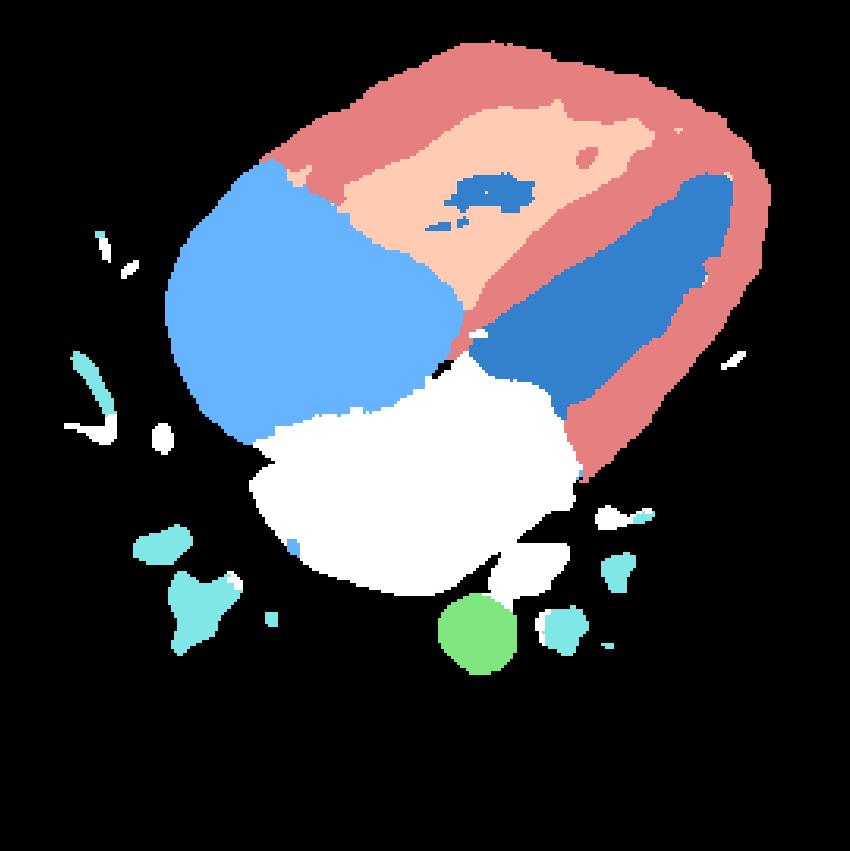}} 
    \vspace{-0.02in} \\  
    \makecell[c]{\includegraphics[width=0.15\linewidth]{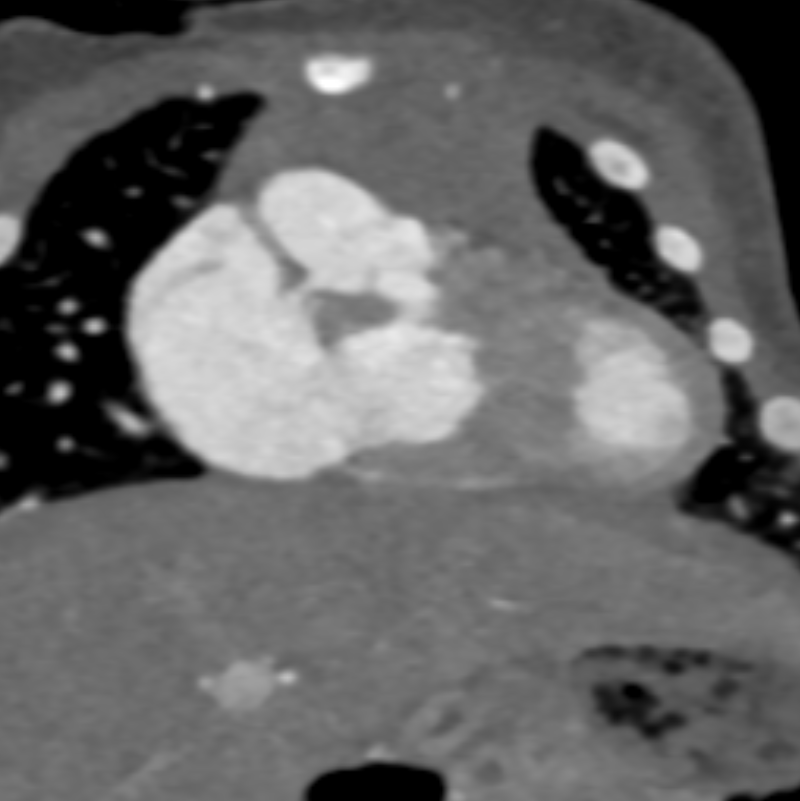}} 
    & \makecell[c]{\includegraphics[width=0.15\linewidth]{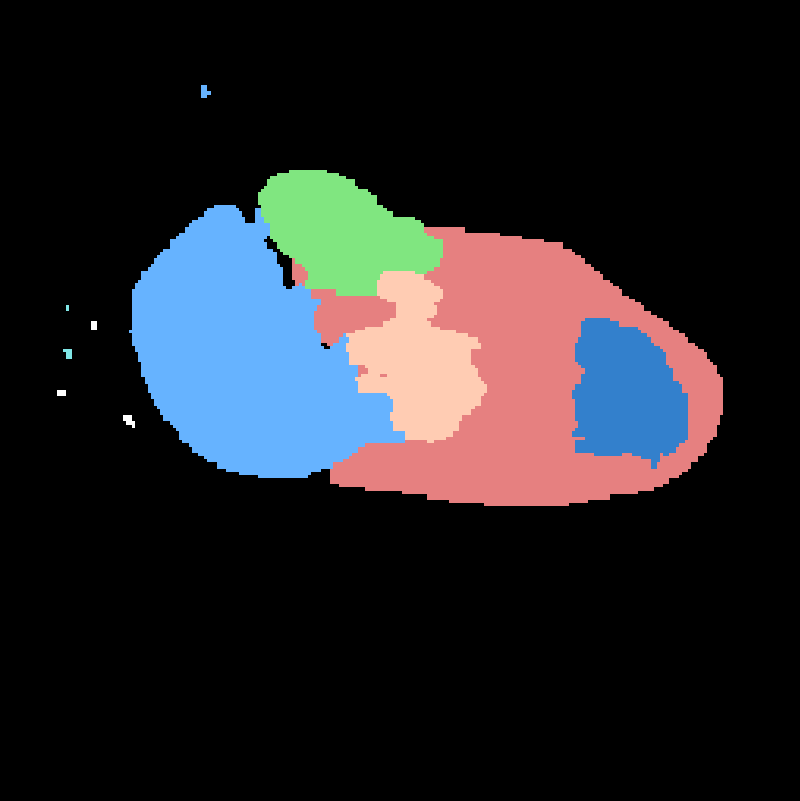}} 
    & \makecell[c]{\includegraphics[width=0.15\linewidth]{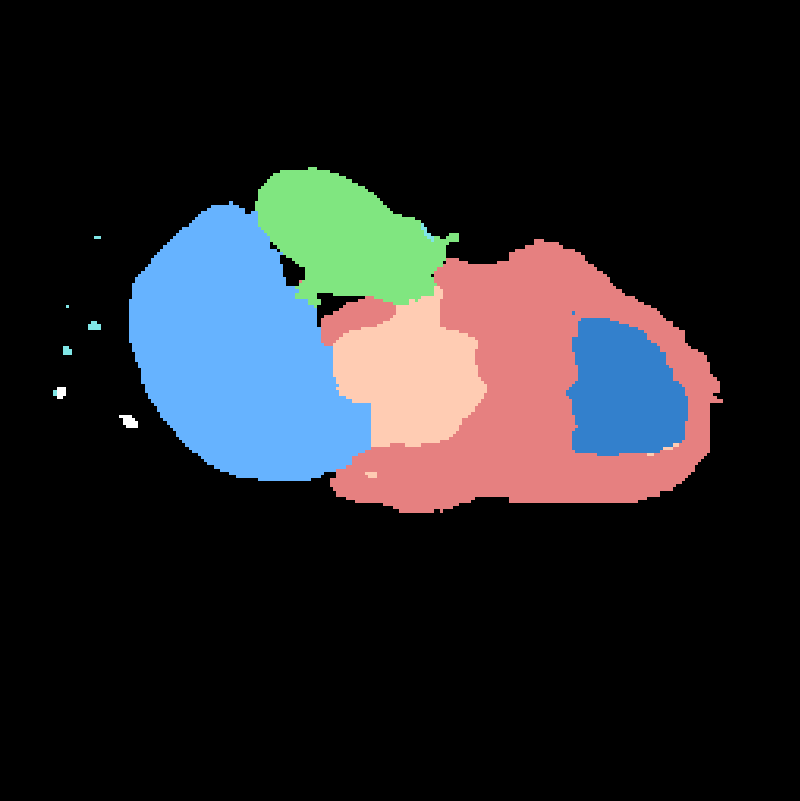}} 
    & \makecell[c]{\includegraphics[width=0.15\linewidth]{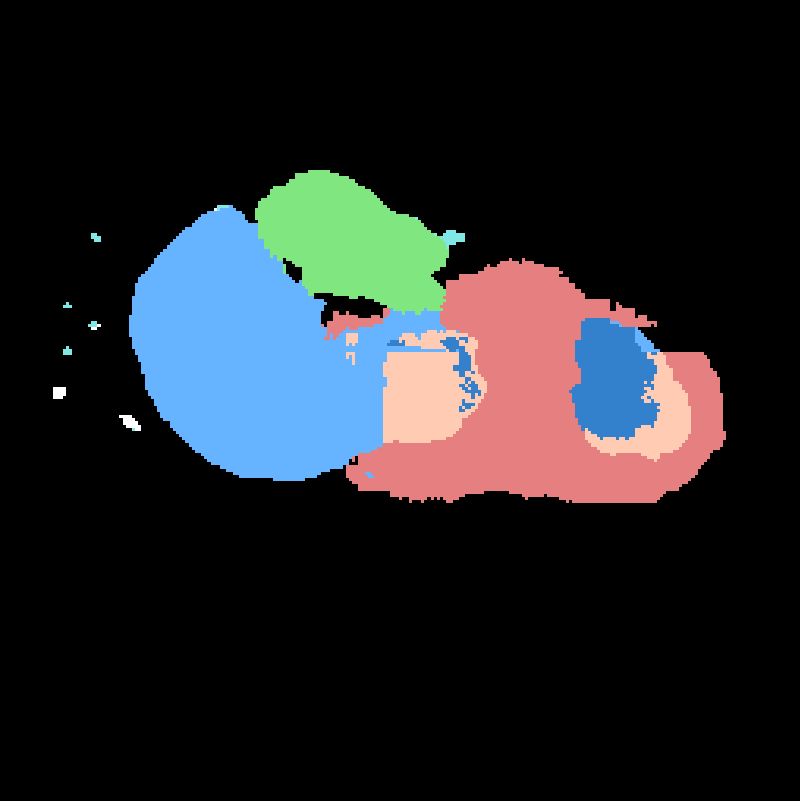}} 
    &\makecell[c]{\includegraphics[width=0.15\linewidth]{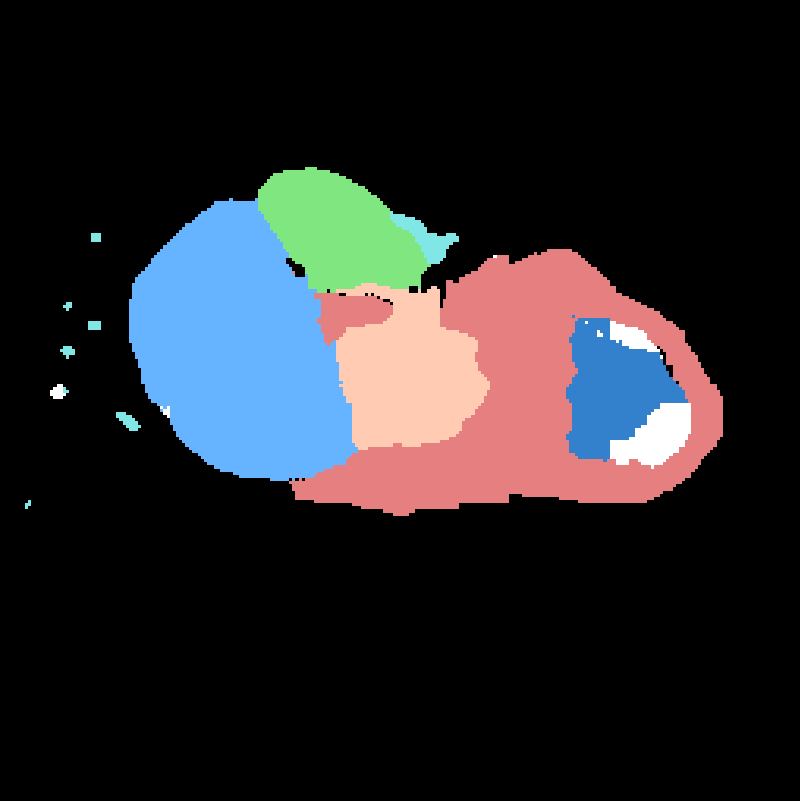}} 
    & \makecell[c]{\includegraphics[width=0.15\linewidth]{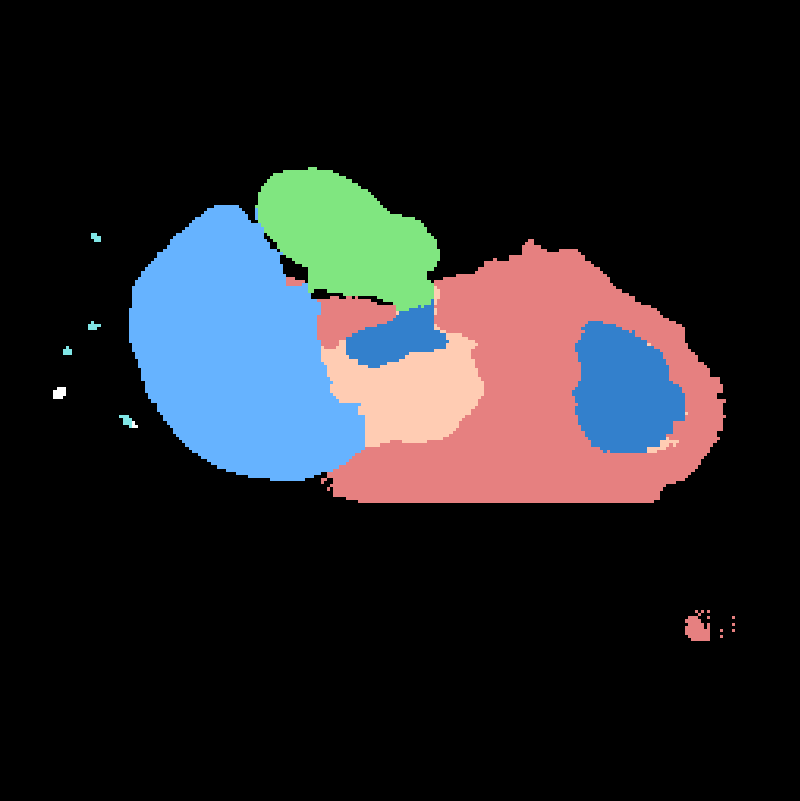}}
    \vspace{-0.02in} \\ 
    \makecell[c]{\includegraphics[width=0.15\linewidth]{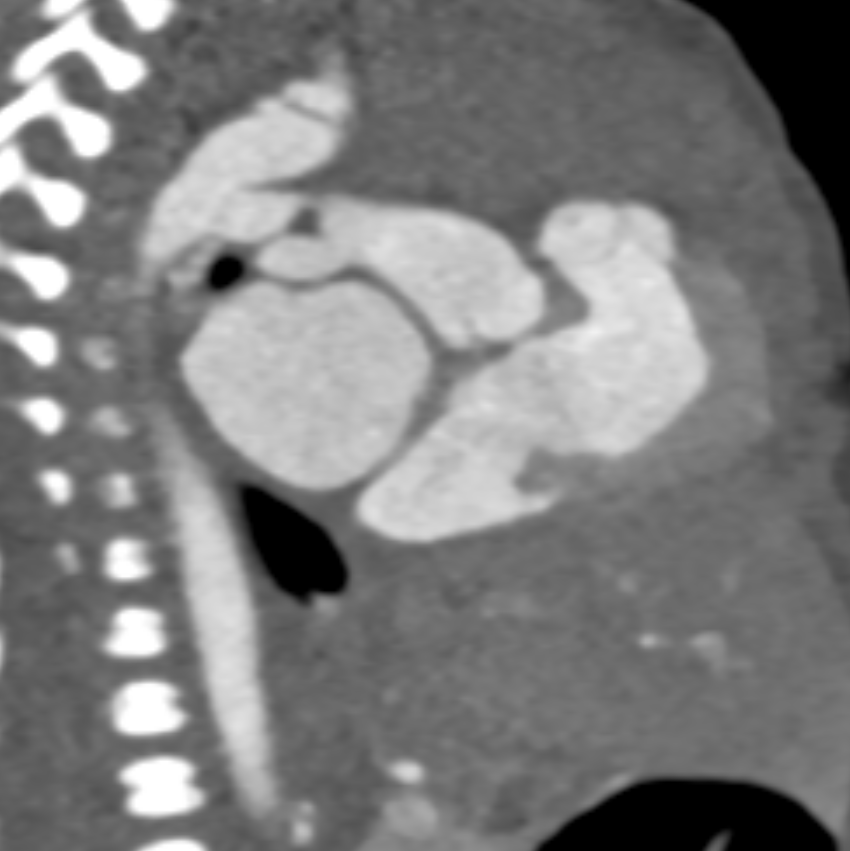}} 
    & \makecell[c]{\includegraphics[width=0.15\linewidth]{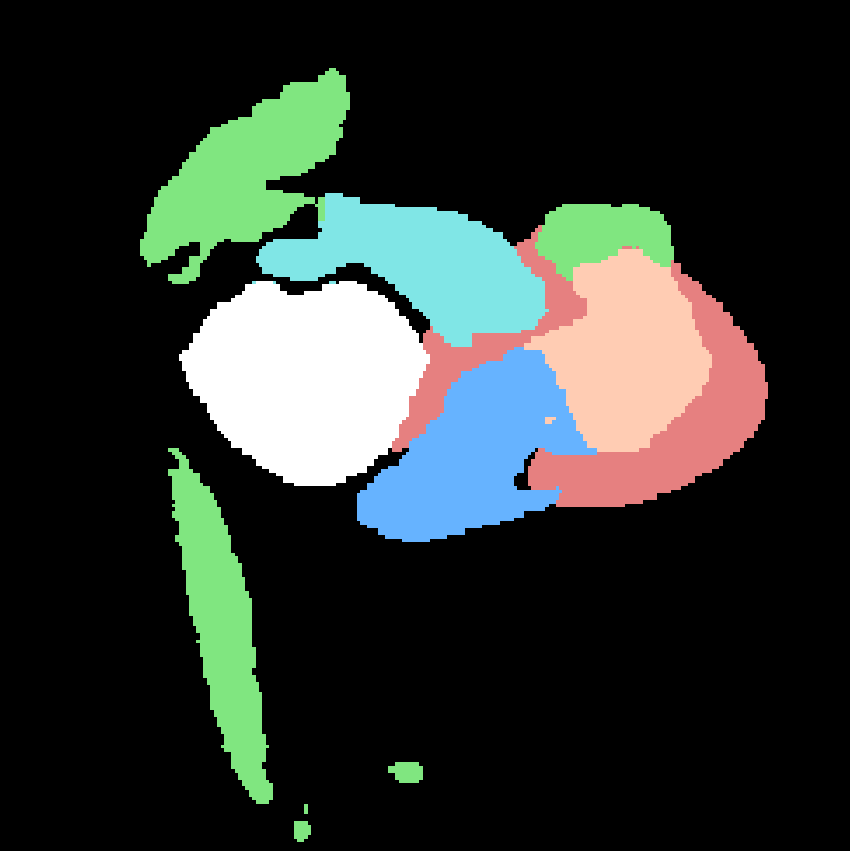}} 
    & \makecell[c]{\includegraphics[width=0.15\linewidth]{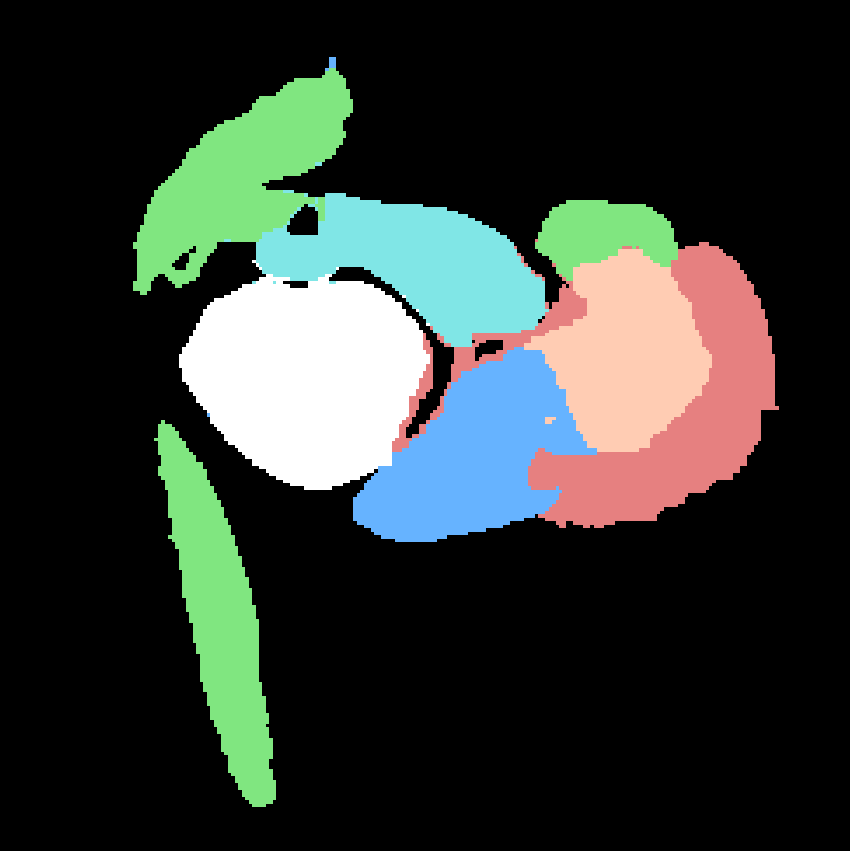}} 
    &  \makecell[c]{\includegraphics[width=0.15\linewidth]{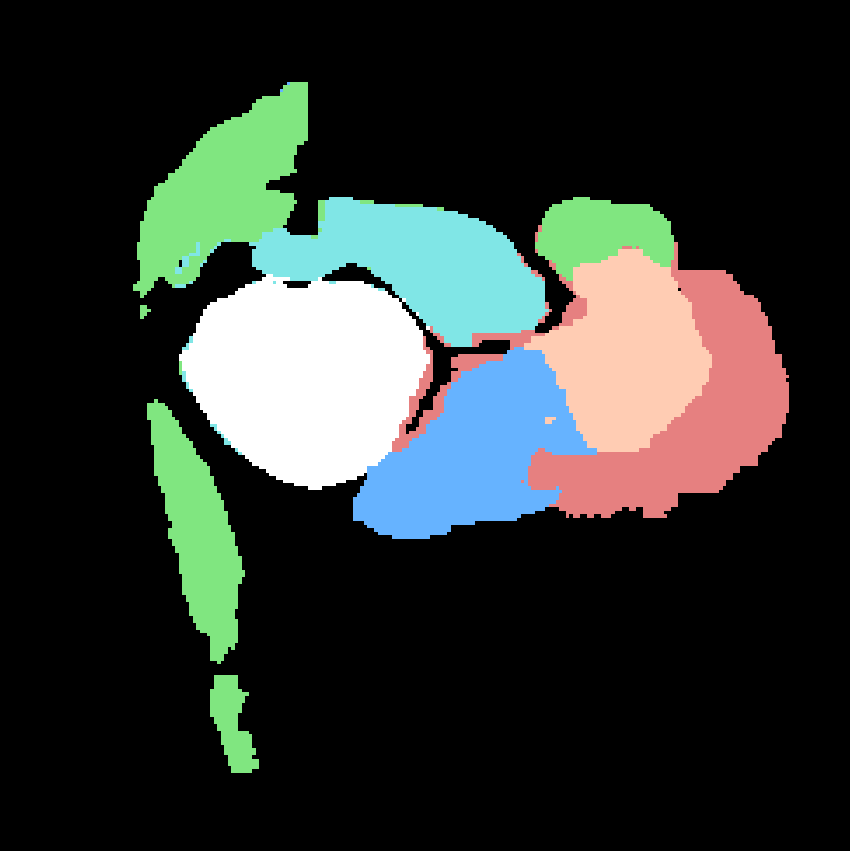}} &\makecell[c]{\includegraphics[width=0.15\linewidth]{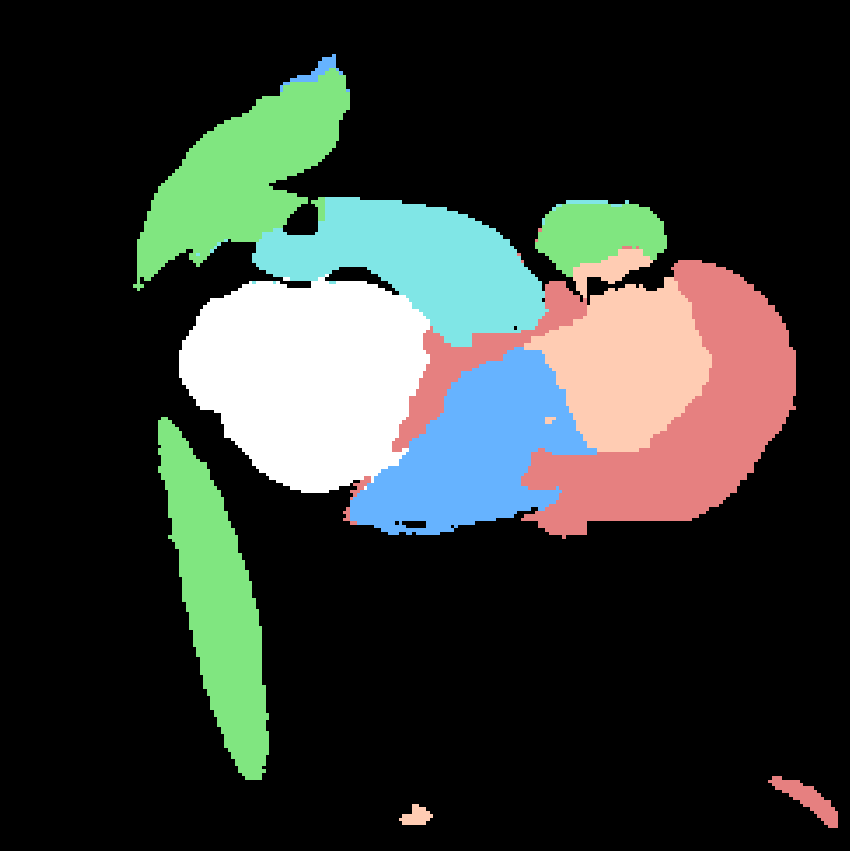}} 
    & \makecell[c]{\includegraphics[width=0.15\linewidth]{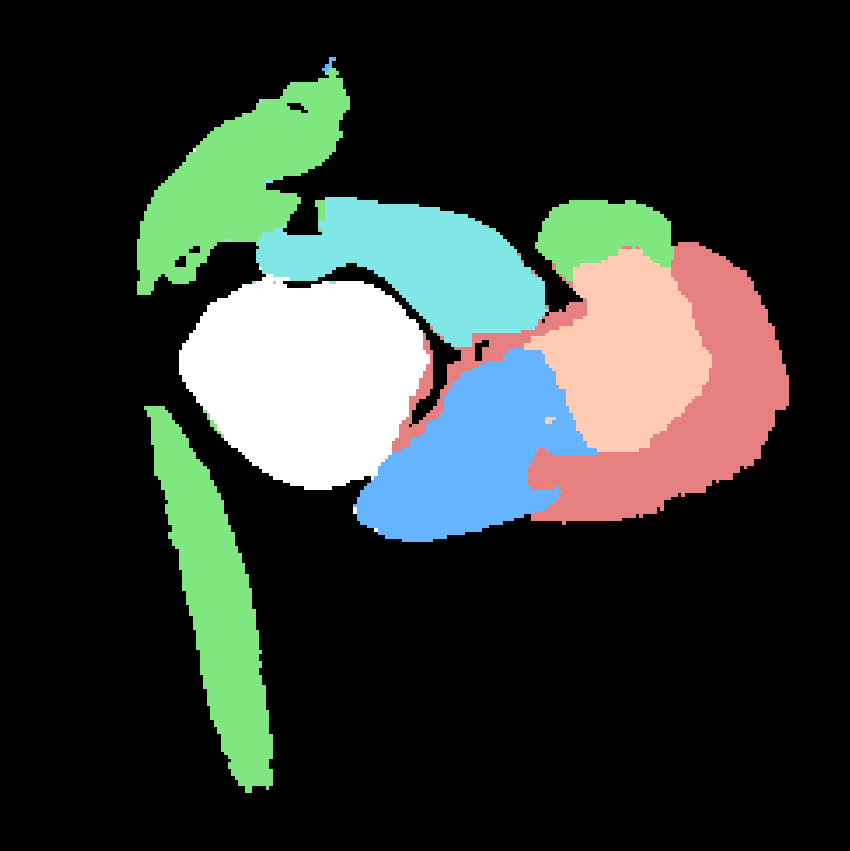}}
    \\
\end{tabular}
\end{minipage}
\caption{Qualitative visualizations of models in few-shot experiment on image with ID ``1085" from ImageCHD dataset. Results are presented in axial, coronal and sagittal views.
}
\label{fig:case_fewshot}
\end{figure}

Cardiac imaging examinations commonly involve multi-modal imaging techniques such as echocardiography, computed tomography (CT) and magnetic resonance imaging (MRI). 
Unlike echocardiography, which is highly operator-dependent, CT and MRI examinations can generate comprehensive and consistent 3D visual representations of the cardiac structure. Such capability enables the diagnosis and quantification of cardiac structural abnormalities and facilitates precise surgical planning. Consequently, whole-heart segmentation is routinely conducted on CT and MRI images.

Conventional methods for cardiac image segmentation, such as probability atlases~\cite{zhuang2016multi, ghosh2021multi} and early machine learning methods~\cite{xu2022mrdff}, have largely been superseded by deep learning, particularly convolutional neural network (CNN) based approaches~\cite{yan2021novel, cui2023improved}.
Whereas CNNs excel at learning hierarchical features on specific small datasets, their inductive biases often limit domain generalization across diverse datasets.
Furthermore, medical image segmentation inherently suffers from a severe scarcity of finely annotated 3D data and a long-tailed distribution of rare diseases. Unlike natural image datasets with millions of samples, medical datasets typically contain only dozens to hundreds of cases, posing a significant obstacle to training robust and universally applicable models.

To tackle these challenges, especially in scenarios with limited annotated data and relatively abundant unlabeled images, self-supervised learning (SSL) — particularly masked image modeling (MIM) — has emerged as a powerful paradigm.
Within this context, the Vision Transformer (ViT)~\cite{vit} has demonstrated exceptional pre-training capabilities and an innate ability to model long-range dependencies.
Our objective is to leverage these advantages in developing a network that can be universally applied to diverse 3D cardiac segmentation tasks. Additionally, we aim to enhance the network's adaptability to images that are rarely or even never encountered.

Building upon our prior work CardiacSeg~\cite{ye2023cardiacseg}, we introduce MCSeg, a volumetric transformer-based network tailored for multi-modal cardiac segmentation. 
MCSeg employs a ViT encoder pre-trained via masked autoencoders (MAE)~\cite{mae} on unannotated cardiac volumes to establish prior anatomical knowledge. Notably, this pre-training stage scales up the data volume by five times compared to our previous work. 
For downstream fine-tuning, this pre-trained encoder is integrated with our customized Scaling Feature Pyramid (SFP), which explicitly bridges single-scale ViT and multi-scale decoder. Furthermore, we incorporate the regional mutual information (RMI) loss~\cite{RMIloss} during the fine-tuning process to further enhance boundary segmentation accuracy. 
As illustrated in the few-shot experiment (Fig. 1), MCSeg significantly outperforms existing baselines, highlighting its exceptional capability in overcoming medical data scarcity.

The main contributions of this paper are outlined as follows:
\begin{itemize}
    \item We introduce MCSeg, a volumetric transformer-based network tailored for multi-modal cardiac segmentation. The key innovation of this architecture is a novel Scaling Feature Pyramid (SFP), which explicitly bridges single-scale ViT and multi-scale decoder, thereby better exploiting the capabilities of the pre-trained 3D ViT.
    \item We adopt a two-stage training paradigm that combines self-supervised 3D MAE pre-training with downstream task fine-tuning. during the fine-tuning, we incorporate a regional mutual information (RMI) loss to improve boundary segmentation accuracy. 
    \item We achieve state-of-the-art (SOTA) performance across four diverse cardiac datasets (ImageCHD, HVSMR-2.0, MM-WHS, and MSD). Extensive experiments demonstrate that MCSeg exhibits cross-modality adaptability and few-shot learning potential. 
\end{itemize}

\section{Related Works}
\label{sec:RelatedWorks}
\subsection{Volumetric Medical Image Segmentation}
\label{ssec:ViTforMIS}
U-Net~\cite{unet} and its 3D variants~\cite{3dunet, vnet} established the typical symmetric encoder-decoder paradigm for medical image segmentation. Recently, volumetric transformers have emerged to model long-range dependencies, broadly categorized into pure transformer and hybrid CNN-transformer architectures. Pure transformers, such as nnFormer~\cite{zhou2023nnformer} and VT-UNet~\cite{VT-UNet}, utilize self-attention mechanisms in both encoding and decoding stages but often require substantial computational resources. Conversely, hybrid models integrate transformer encoders with CNN decoders to balance global context modeling with local detail sensitivity. For instance, CoTr~\cite{cotr} connects a deformable transformer to CNN decoder, whereas UNETR~\cite{unetr} and SwinUNETR~\cite{tang2022self} bridge ViT and Swin Transformer encoders to CNN decoders via multi-resolution skip connections. Concurrently, modernized CNN baselines like MedNeXt~\cite{roy2023mednext} and foundation models like SAM-Med3D~\cite{wang2024sam} have demonstrated strong baseline performance. 
Whereas these architectures demonstrate strong capabilities, a dedicated pre-training and fine-tuning paradigm tailored specifically to cardiac-specific tasks may better align with practical clinical requirements. To this end, MCSeg couples a tailored hybrid network design with specialized large-scale cardiac pre-training, constructing prior contextual information for highly accurate and clinically reliable segmentation.

\subsection{Masked Image modeling}
\label{ssec:MIM}
Whereas the massive parameter count of transformers enables them to achieve excellent performance, it also makes them highly dependent on large-scale training data. To alleviate this heavy data dependency, MIM has been widely adopted for self-supervised ViT pre-training. Adapted from natural language processing (NLP)~\cite{devlin2018bert}, visual approaches like BEiT~\cite{bao2021beit} and SimMIM~\cite{xie2022simmim} reconstruct masked image patches to learn semantic representations. Notably, MAE~\cite{mae} employs an asymmetric encoder-decoder architecture with a high masking ratio, significantly reducing training computation, a paradigm subsequently extended to 3D medical imaging~\cite{med3dmae}. 
By leveraging 3D MAE on unannotated cardiac datasets, our MCSeg effectively overcomes the data scarcity challenge, establishing comprehensive anatomical priors that directly benefits downstream fine-tuning.

\subsection{Feature Pyramid Networks}
\label{ssec:FPNs}
For pixel‑level vision tasks such as object detection and semantic segmentation, preserving original input resolution throughout training is computationally prohibitive and impedes the extraction of high‑level semantic information.
Feature Pyramid Networks (FPNs)~\cite{lin2017fpn} leverage the naturally occurring multi-scale feature maps of CNN backbones by combining a top‑down pathway with lateral connections to construct semantically rich feature pyramids. 
In segmentation frameworks such as Panoptic FPN~\cite{kirillov2019panoptic}, FPNs further facilitate multi‑scale feature integration during decoder up‑sampling.
Since standard ViTs output only single-scale token sequences, frameworks like ViTDet~\cite{vitdet} incorporate feature pyramids to adapt ViT backbones for dense prediction tasks. 
In volumetric medical segmentation, simple skip-connection schemes struggle to efficiently bridge the scale and semantic gap between the ViT's single-scale representations and the CNN decoder's multi-scale feature requirements. To resolve this, MCSeg employs a customized SFP that explicitly transforms the deepest well-encoded ViT features into a multi-scale pyramid, ensuring optimal cross-scale contextual utilization for the decoder.

\begin{figure*}[!t]
\centering
\resizebox{0.75\textwidth}{!}{\includegraphics{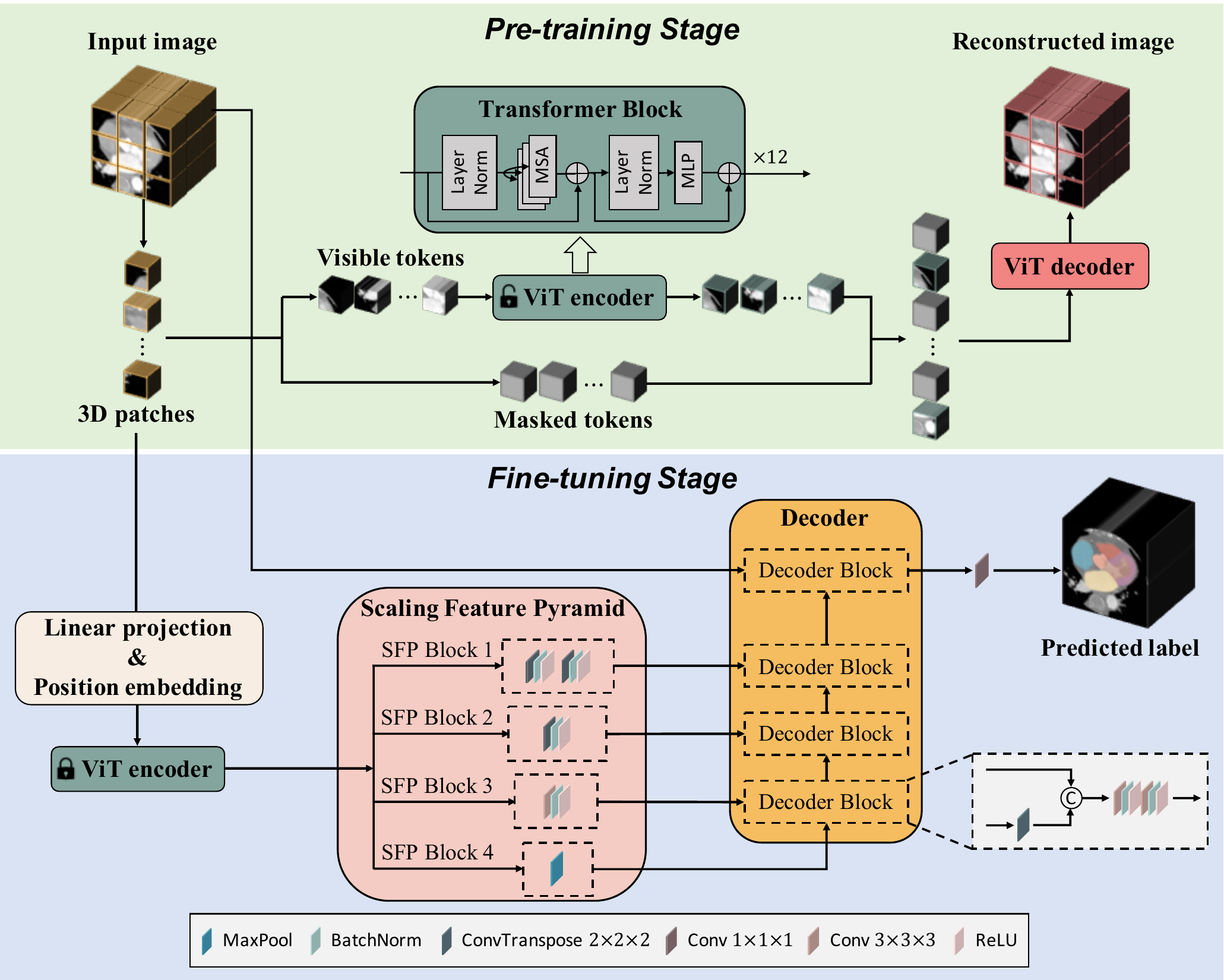}}
\caption{The schematic overview of MCSeg. During the pre-training stage, the ViT encoder undergoes training using the MAE method. Subsequently, in the fine-tuning stage, the ViT encoder loads the pre-trained weights and freezes its parameters. The training process in this stage exclusively focuses on the SFP and the decoder.}
\label{fig:architecture}
\end{figure*}

\section{Method}
\label{sec:Method}
\subsection{Network Architecture}
\label{ssec:Architecture}

Our MCSeg comprises three primary components: a volumetric transformer encoder, an SFP, and a CNN decoder (Fig. \ref{fig:architecture}). Unlike conventional U-shaped networks that rely on hierarchical encoders, MCSeg leverages a powerful, non-hierarchical pre-trained ViT backbone to extract global contextual representations, which are subsequently remapped into a multi-scale pyramid via the SFP to facilitate precise segmentation.

\subsubsection{Volumetric Embedding and Encoder}
\label{sssec:Encoder}

Given a 3D input volume $x \in \mathbb{R}^{H \times W \times D}$ with height $H$, width $W$, and depth $D$, the initial step involves partitioning the volume into $N$ non-overlapping cubic patches $p_i \in \mathbb{R}^{P \times P \times P}$, where $P=16$ denotes the fixed patch size and $N = \frac{H \times W \times D}{P^3}$. Each patch $p_i$ is flattened and projected into an $E$-dimensional latent space through a linear mapping function $f: p_i \rightarrow e_i \in \mathbb{R}^E$. This process yields a sequence of patch embeddings $\mathbf{x_e} = [e_1, e_2, \dots, e_N] \in \mathbb{R}^{N \times E}$
To preserve spatial information within the 1D sequence, learnable positional embeddings $\mathbf{x_p} \in \mathbb{R}^{N \times E}$ are added to the sequence. The combined representation, $\mathbf{z_0} = \mathbf{x_e} + \mathbf{x_p}$, serves as the input to the ViT encoder. We adopt the ViT-B configuration consisting of $L=12$ successive transformer blocks, where each block models long-range volumetric dependencies through multi-head self-attention. The encoder maintains a constant latent resolution of $\frac{H}{P} \times \frac{W}{P} \times \frac{D}{P}$ across all layers. This structure is strategically chosen to align with the requirements of large-scale MIM pre-training, ensuring that the global semantic features are robustly encoded before being passed to the decoder.

\subsubsection{Scaling Feature Pyramid (SFP) and Decoder}
\label{sssec:SFP&decoder}

The integration of SFP with the CNN decoder enables segmentation generation from the single-scale outputs of the ViT encoder. To enrich the feature representations for progressive up-sampling, the SFP transforms the ViT’s single-scale outputs into multi-scale features, allowing the decoder to access both high-level contextual information and scale-appropriate spatial details.
The SFP transforms the final outputs $\mathbf{z_{12}}$ of the encoder, which have a size of $\left(\frac{H}{16}, \frac{W}{16}, \frac{D}{16}, E\right)$ (\textit{i.e.}, a scale of $\frac{1}{16}$), into feature maps at scales $\frac{1}{4}$, $\frac{1}{8}$, $\frac{1}{16}$, and $\frac{1}{32}$ using four SFP blocks (see Fig. \ref{fig:architecture}).
The decoder consists of four CNN decoder blocks. In the bottom three decoder blocks, feature representations of relatively small resolution are up-sampled using transpose convolution layers. These up-sampled feature maps are then concatenated with the corresponding feature maps from the SFP. 
The last decoder block is similar to the other decoder blocks, except that it concatenates the up-sampled lower-resolution feature maps with the original input image. After these decoder blocks, the segmentation result is obtained through a fully connected layer.

\subsection{Fine-tuning for Cardiac Segmentation}
\label{ssec:finetune}

\subsubsection{Training Strategy}
\label{sssec:TrainingStrategy}
In the training stage, the pre-trained ViT backbone was adopted as the encoder and frozen during training. In contrast, only the SFP and decoder were trained to construct the segmentation results in this stage.

\subsubsection{Loss Function}
\label{sssec:LossFunction}
During training, we optimize the network using a combination of soft Dice loss~\cite{vnet}, cross-entropy (CE) loss, and regional mutual information (RMI) loss~\cite{RMIloss}. 
Whereas Dice and CE losses strictly assess overall class-level and voxel-level accuracy respectively, they lack local spatial restrictions. 
Since accurate and smooth boundaries are crucial in medical image segmentation~\cite{clough2020topological}, we incorporate RMI loss. 
By maximizing the mutual information between the predicted and ground-truth regions, RMI loss essentially enhances local consistency and improves boundary segmentation accuracy.
Therefore, the RMI loss is defined as described in~\cite{RMIloss}:
\begin{equation}
    L_{rmi} \left(G, P \right) = \frac{1}{B} \sum_{b=1}^{B} \sum_{c=1}^{C} \left(-I^{b,c} \left(G, P \right) \right)
\end{equation}
where $B$ is the number of regions in the input, which is determined by the down-sampling factor $d_s$ and square region size $r$. $I \left(G, P \right)$ represents the mutual information between $G$ and $P$, calculated according to the formula in: 
\begin{equation}
    I \left(G, P \right) = \sum_{x \in \mathcal{G}} \sum_{y \in \mathcal{P}} p\left(x, y\right) \log \frac{p\left(x, y\right)}{p\left(x\right) p\left(y\right)}.
\end{equation}

In summary, the goal of the training stage is to minimize the compound loss function as follows:
\begin{equation}
\begin{split}
    L \left(G, P \right) = \  &  L_{dice} \left(G, P \right) + \lambda_{ce}  L_{ce} \left(G, P \right) \\
    & + \sigma_{ep_{rmi}} \lambda_{rmi} L_{rmi}\left(G, P \right),
\end{split}
\end{equation}
where $\lambda_{ce}$ and $\lambda_{rmi}$ are hyper-parameters. In our experiments, $\lambda_{ce}=0.5$ and $\lambda_{rmi}=0.1$. $\sigma_{ep_{rmi}}$ is a smooth step function designed for regulating the timing of incorporating RMI loss into the compound loss function during training, it is defined as
\begin{equation}
    \sigma_{ep_{rmi}} = \frac{1}{1 + e^{ep_{rmi} - ep_{cur}}}, 
\end{equation}
where $ep_{rmi}$ is a hyper-parameter, $ep_{cur}$ and $ep_{max}$ represent the current and maximum epoch during training, $ep_{cur} \in \left[0, ep_{max}\right] \cap \mathbb{Z}$.

\section{Experiments}
\label{sec:Experiments}
\subsection{Datasets}
\label{ssec:Datasets}

\subsubsection{ImageCAS}
\label{sssec:ImageCAS}
ImageCAS~\cite{zeng2023imagecas} is a large-scale coronary artery segmentation dataset consisting of 1000 CTA images. As the complete heart structure is visible in coronary CTA images, we utilize the entire dataset solely for pre-training.

\subsubsection{ImageCHD}
\label{sssec:CHD}
ImageCHD~\cite{ImageCHD} is a comprehensive dataset dedicated to the segmentation of congenital heart diseases (CHDs). It consists of 110 3D CT images, covering 16 types of CHDs. Each image in the dataset is annotated with labels for seven cardiac substructures, including the left ventricle (LV), right ventricle (RV), left atrium (LA), right atrium (RA), myocardium (Myo), aorta (AO), and pulmonary artery (PA). 
During fine-tuning, the dataset was divided into training, validation, and test sets, consisting of 77, 11, and 22 cases, respectively, based on disease types. Images in the training and validation sets were also utilized for pre-training.

\subsubsection{HVSMR-2.0}
\label{sssec:HVSMR}
HVSMR-2.0~\cite{pace2024hvsmr} is a 3D cardiovascular MRI dataset for whole-heart segmentation in CHDs, which consisting of 60 cardiovascular MRI scans with manual segmentation masks of the four cardiac chambers and four great vessels, \textit{i.e.}, LV, RV, LA, RA, AO, PA, superior vena cava (SVC), and inferior vena cava (IVC). 
The dataset was divided into training, validation, and test sets comprising 42, 6, and 12 samples, respectively, for fine-tuning, and all images were utilized for pre-training except for those in the test set.

\subsubsection{MM-WHS}
\label{sssec:MMWHS}
MM-WHS~\cite{mmwhs1,mmwhs2} is a 3D multi-modal whole-heart segmentation dataset, consisting of 60 CT and 60 MRI images. 
Each modality containing 20 annotated cases, all of which were utilized in comparative experiments. The remaining 40 unlabeled images in each modality were utilized for pre-training.
The cardiac substructures annotated in this dataset are identical to those in the ImageCHD dataset.

\subsubsection{MSD Task02 Heart}
\label{sssec:MSD}
The Task02 of Medical Segmentation Decathlon (MSD)~\cite{msd} 
is the left atrium segmentation for MRI images, including 20 labeled training data.
This data was employed for comparative experiment.

\begin{table*}[!t]
\caption{Quantitative comparison of full-data experimental results for different models on (a) ImageCHD and (b) HVSMR-2.0 datasets. (${\dag}$ indicates models adopt their original pre-trained weights. Subscripts ${ViT-B_{223}}$ and ${ViT-B_{723}}$ denote models utilized the ViT-B pre-trained on 223 and 723 CT images, respectively, in this study. Wilcoxon signed-rank test \textit{p}-values (each model vs. MCSeg): $^{*}(p < 0.05)$, $^{**}(p < 0.01)$, $^{***}(p < 0.001)$.)}
\label{tab:exp_full}
\centering
\vspace{-0.1cm}
\begin{subtable}{\linewidth}
    \caption{Results on ImageCHD dataset.}
    \vspace{-0.1cm}
    \label{tab:exp_chd}
    \centering
    \resizebox{0.98\textwidth}{!}{
    \renewcommand\arraystretch{1.1}
    \begin{tabular}{l|c|ccccccc}
        \toprule
        \textbf{Networks} &\textbf{Overall} & \textbf{LV} & \textbf{RV} & \textbf{LA} & \textbf{RA} & \textbf{Myo} & \textbf{AO} & \textbf{PA} \\
        \midrule
        \rowcolor{gray!20}
        \multicolumn{9}{l}{\textit{Pure CNN-based Methods}} \\
        UNet++$^{**}$ & $0.621 \pm 0.056$ & $0.715 \pm 0.092$ & $0.487 \pm 0.124$ & $0.797 \pm 0.057$ & $0.698 \pm 0.058$ & $0.460 \pm 0.104$ & $0.562 \pm 0.083$ & $0.619 \pm 0.114$ \\
        nnUNet$^{*}$ & $0.882 \pm 0.050$ & $\mathbf{0.925 \pm 0.022}$ & $\underline{0.882 \pm 0.039}$ & $\underline{0.916 \pm 0.022}$ & $0.896 \pm 0.025$ & $0.810 \pm 0.261$ & $\underline{0.887 \pm 0.066}$ & $\underline{0.857 \pm 0.089}$ \\
        MedNeXt$^{**}$ & $0.858 \pm 0.043$ & $0.866 \pm 0.084$ & $0.826 \pm 0.107$ & $0.895 \pm 0.029$ & $0.891 \pm 0.032$ & $0.849 \pm 0.091$ & $0.866 \pm 0.066$ & $0.815 \pm 0.081$ \\
        \midrule
        \rowcolor{gray!20}
        \multicolumn{9}{l}{\textit{Pure Transformer-based Methods}} \\
        nnFormer$^{**}$ & $0.850 \pm 0.033$&$ 0.893 \pm 0.046$&$ 0.851 \pm 0.079$&$ 0.875 \pm 0.032$&$ 0.894 \pm 0.029$&$ 0.872 \pm 0.049$&$ 0.814 \pm 0.079$&$0.759 \pm 0.089$\\
        VT-UNet$^{\dag}$$^{**}$ & $0.820 \pm 0.060$& $0.897 \pm 0.050$& $0.832 \pm 0.082$& $0.854 \pm 0.033$& $0.866 \pm 0.047$&$0.800 \pm 0.254$&$0.787 \pm 0.086$&$0.704 \pm 0.081$\\
        \midrule
        \rowcolor{gray!20}
        \multicolumn{9}{l}{\textit{CNN \& Transformer Hybrid Methods}} \\
        CoTr$^{**}$ & $0.852 \pm 0.040$& $0.902 \pm 0.037$& $0.818 \pm 0.129$& $0.886 \pm 0.026$& $0.864 \pm 0.043$& $ 0.849 \pm 0.050$& $ 0.845 \pm 0.068$&$ 0.803 \pm 0.095$ \\ 
        TransUNet$^{**}$ & $\underline{0.887 \pm 0.026}$ & $0.916 \pm 0.040$ & $0.877 \pm 0.076$ & $0.911 \pm 0.022$ & $\underline{0.907 \pm 0.026}$ & $0.879 \pm 0.053$ & $0.881 \pm 0.044$ & $0.840 \pm 0.065$ \\
        UNETR$_{ViT-B_{723}}$$^{**}$ & $0.879 \pm 0.036$ & $ 0.910 \pm 0.044$ & $0.867 \pm 0.069$ & $0.901 \pm 0.040$ & $ 0.896 \pm 0.033$ & $\mathbf{0.892 \pm 0.043}$& $0.876 \pm 0.057$ & $0.817 \pm 0.116$\\
        SwinUNETR$^{\dag}$$^{**}$ & $ 0.870 \pm 0.036$ & $0.895 \pm 0.044$ & $0.848 \pm 0.093$ & $0.898 \pm 0.029$ & $0.896 \pm 0.029$ & $0.875 \pm 0.054$ & $0.862 \pm 0.083$ & $0.822 \pm 0.099$\\
        CardiacSeg$^{\dag}$$^{**}$ & $0.875 \pm 0.031$& $0.918 \pm 0.032$& $0.861 \pm 0.055$& $0.899 \pm 0.029$ & $0.899 \pm 0.031$& $0.878 \pm 0.055$& $0.861 \pm 0.054$& $0.813 \pm 0.083$\\
        SAM-Med3D$^{\dag}$$^{**}$ & $0.544 \pm 0.137$ & $0.763 \pm 0.118$ & $0.702 \pm 0.220$ & $0.489 \pm 0.240$ & $0.671 \pm 0.106$ & $0.475 \pm 0.219$ & $0.312 \pm 0.164$ & $0.396 \pm 0.162$ \\
        \textbf{MCSeg$\mathbf{_{ViT-B_{723}}}$ (Ours)} & $\mathbf{0.896 \pm 0.024}$&$\underline{0.923 \pm 0.033}$&$\mathbf{0.888 \pm 0.044}$&$\mathbf{0.918 \pm 0.021}$&$\mathbf{0.909 \pm 0.026}$&$\underline{0.886 \pm 0.038}$&$\mathbf{0.896 \pm 0.055}$&$\mathbf{0.857 \pm 0.077}$\\
        \bottomrule
    \end{tabular}}
\end{subtable}
\vspace{0cm} \\
\vspace{-0.08cm}
\begin{subtable}{\linewidth}
    \caption{Results on HVSMR-2.0 dataset.}
    \vspace{-0.1cm}
    \label{tab:exp_hvsmr}
    \centering
    \resizebox{0.98\textwidth}{!}{
    \renewcommand\arraystretch{1.1}
    \begin{tabular}{l|c|cccccccc}
        \toprule
        \textbf{Networks} &\textbf{Overall} & \textbf{LV} & \textbf{RV} & \textbf{LA} & \textbf{RA} & \textbf{AO} & \textbf{PA}  & \textbf{SVC}&\textbf{IVC}\\
        \midrule
        \rowcolor{gray!20}
        \multicolumn{10}{l}{\textit{Pure CNN-based Methods}} \\
        UNet++$^{***}$ & $0.783 \pm 0.083$ & $0.885 \pm 0.089$ & $0.765 \pm 0.222$ & $0.786 \pm 0.172$ & $0.838 \pm 0.052$ & $0.844 \pm 0.090$ & $0.696 \pm 0.187$ & $0.738 \pm 0.115$ & $0.727 \pm 0.066$ \\
        nnUNet$^{*}$ & $0.816 \pm 0.043$ & $0.872 \pm 0.034$ & $0.827 \pm 0.122$ & $\underline{0.838 \pm 0.023}$ & $0.778 \pm 0.259$ & $0.870 \pm 0.024$ & $\mathbf{0.797 \pm 0.060}$ & $0.778 \pm 0.049$ & $0.771 \pm 0.046$ \\
        MedNeXt$^{***}$ & $0.807 \pm 0.079$ & $0.889 \pm 0.099$ & $0.803 \pm 0.222$ & $0.809 \pm 0.099$ & $\underline{0.850 \pm 0.081}$ & $0.860 \pm 0.090$ & $0.709 \pm 0.170$ & $0.758 \pm 0.124$ & $0.794 \pm 0.092$ \\
        \midrule
        \rowcolor{gray!20}
        \multicolumn{10}{l}{\textit{Pure Transformer-based Methods}} \\
        nnFormer$^{***}$ & $0.727 \pm 0.102$ & $0.818 \pm 0.151$ & $0.725 \pm 0.235$ & $0.712 \pm 0.154$ & $0.789 \pm 0.128$ & $0.818 \pm 0.082$ & $0.634 \pm 0.172$ & $0.643 \pm 0.124$ & $0.701 \pm 0.157$ \\
        VT-UNet$^{\dag}$$^{**}$ & $0.816 \pm 0.073$ & $\underline{0.917 \pm 0.055}$ & $ \mathbf{0.843 \pm 0.221}$& $ 0.815 \pm 0.090$ & $ 0.795 \pm 0.246$ & $0.882 \pm 0.064$ & $0.750 \pm 0.144$ & $0.736 \pm 0.151$ & $0.789 \pm 0.044$ \\
        \midrule
        \rowcolor{gray!20}
        \multicolumn{10}{l}{\textit{CNN \& Transformer Hybrid Methods}} \\
        CoTr$^{**}$ & $0.810 \pm 0.089$ & $ 0.899 \pm 0.087$ & $ 0.832 \pm 0.190$ & $0.831 \pm 0.104$ & $ 0.784 \pm 0.258$ & $ 0.883 \pm 0.080$ & $ 0.754 \pm 0.148$ & $0.709 \pm 0.225$ & $0.800 \pm 0.062$ \\ 
        TransUNet$^{*}$ & $\underline{0.818 \pm 0.061}$ & $0.878 \pm 0.078$ & $0.812 \pm 0.187$ & $0.828 \pm 0.060$ & $0.837 \pm 0.097$ & $0.869 \pm 0.072$ & $0.750 \pm 0.145$ & $0.786 \pm 0.087$ & $0.787 \pm 0.064$ \\
        UNETR$_{ViT-B_{223}}$$^{*}$ & $0.807 \pm 0.083$ & $ 0.887 \pm 0.102$ & $ 0.806 \pm 0.218$ & $ \mathbf{0.841 \pm 0.070}$& $ 0.810 \pm 0.205$ & $\underline{0.887 \pm 0.072}$ & $ 0.705 \pm 0.188$ & $\underline{0.783 \pm 0.149}$ & $0.766 \pm 0.058$ \\
        SwinUNETR$^{\dag}$$^{**}$ & $0.795 \pm 0.073$ & $ 0.835 \pm 0.095$ & $ 0.777 \pm 0.243$ & $ 0.823 \pm 0.036$ & $\mathbf{0.851 \pm 0.071}$& $ 0.839 \pm 0.071$ & $ 0.735 \pm 0.142$ & $ 0.757 \pm 0.120$ & $0.750 \pm 0.071$ \\
        CardiacSeg$^{\dag}$$^{*}$ & $0.816 \pm 0.065$ & $ 0.870 \pm 0.103$ & $ 0.820 \pm 0.094$ & $ 0.808 \pm 0.097$ & $ 0.796 \pm 0.255$ & $ 0.881 \pm 0.062$ & $0.762 \pm 0.132$ & $ \mathbf{0.793 \pm 0.070}$& $\underline{0.805 \pm 0.067}$ \\
        SAM-Med3D$^{\dag}$$^{***}$ & $0.595 \pm 0.131$ & $0.841 \pm 0.095$ & $0.818 \pm 0.115$ & $0.559 \pm 0.141$ & $0.772 \pm 0.079$ & $0.417 \pm 0.218$ & $0.526 \pm 0.224$ & $0.460 \pm 0.142$ & $0.368 \pm 0.218$ \\
        \textbf{MCSeg$\mathbf{_{ViT-B_{223}}}$ (Ours)} & $\mathbf{0.832 \pm 0.071}$& $\mathbf{0.925 \pm 0.042}$& $\underline{0.840 \pm 0.175}$ & $0.825 \pm 0.125$ & $0.817 \pm 0.199$ & $\mathbf{0.894 \pm 0.058}$& $\underline{0.775 \pm 0.143}$& $0.775 \pm 0.146$ & $\mathbf{0.810 \pm 0.050}$\\
        \bottomrule
    \end{tabular}}
\end{subtable}
\vspace{-0.2cm}
\end{table*}

\subsection{Implementation Details}
\label{ssec:Implementation}
Our MCSeg framework was built on PyTorch\footnote{\url{https://pytorch.org/}} and MONAI\footnote{\url{https://monai.io/}}.
The model was pre-trained on four 80G A100 GPUs using the open-source implementation available at OpenMedIA\footnote{\url{https://openi.pcl.ac.cn/OpenMedIA}}\footnote{\url{https://openi.pcl.ac.cn/OpenMedIA/Transformer3DSeg}}~\cite{zhuang2022openmedia}. All subsequent fine-tuning and ablation studies were conducted on a single 32G V100 GPU.
The ViT backbone underwent pre-training for 3,200 epochs with batch size of 12 per GPU, using AdamW optimizer with a learning rate of $10^{-3}$.
During fine-tuning stage, models were trained for 300 epochs with batch size of 2, utilizing AdamW optimizer (learning rate of $10^{-3}$, weight decay of $10^{-5}$), complemented by a 20-epoch linear warm-up and a cosine annealing learning rate scheduler.
As part of the pre-processing steps, all input data was initially resampled to spacing of $1 \times 1 \times 1 \ mm^{3}$. For CT images, the HU values were clipped into $[500, 2000]$ (for ImageCHD) or $[-300, 700]$ (for MM-WHS) and normalized to the range $[0, 1]$. For MRI images, 1\% to 99\% of the image intensities were clipped and normalized to $[0, 1]$.
Subsequently, the input image was cropped into a bounding box containing only the foreground based on its label. This bounding box was then randomly cropped into four cubes, each of size $\left(128, 128, 128\right)$, serving as input samples. Data augmentations included random addition of Gaussian noise with standard deviation of 0.05, random scaling of intensity by factor of 0.1, and intensity shifting with randomly selected offset from $[-0.1, 0.1]$ for the images.

The Dice coefficient~\cite{vnet} is employed to measure the experimental results.

\subsection{Comparative Experiments}
MCSeg was compared with eleven volumetric medical image segmentation networks, including three pure CNN-based methods (UNet++~\cite{zhou2019unetplusplus}, nnUNet~\cite{isensee2021nnu}, MedNeXt~\cite{roy2023mednext}), two pure transformer-based methods (nnFormer~\cite{zhou2023nnformer}, VT-UNet~\cite{VT-UNet}), and six hybrid CNN–Transformer methods (CoTr~\cite{cotr}, TransUNet~\cite{transunet}, UNETR~\cite{unetr}, SwinUNETR~\cite{tang2022self}, CardiacSeg~\cite{ye2023cardiacseg}, SAM-Med3D~\cite{wang2024sam}).
In the experiments, VT-UNet, SwinUNETR, CardiacSeg, and SAM-Med3D employed the pre-trained weights and fine-tuning strategies mentioned in their original papers.
For UNETR\footnote{Pre-trained weights were not utilized in the original UNETR paper. Since UNETR also adopts a ViT encoder, we included a version of UNETR initialized with the same pre-trained weights as MCSeg to ensure a fair comparison and better highlight the superiority of our proposed MCSeg.}~\cite{unetr} and MCSeg, we used the pre-trained ViT-B weights proposed in this study and kept their parameters frozen during training.

\subsubsection{Whole-heart Segmentation on ImageCHD Dataset}
\label{sec:exp_chd}
On the ImageCHD dataset, we evaluate model adaptability under both full-data and few-shot settings.

\begin{figure}[!t]
\begin{minipage}{0.5\linewidth}
\centering
\tabcolsep=0.02cm
\scriptsize
\begin{tabular}{ccccc}
\makecell[c]{Method \\ (Dice Score)} & Axial & Coronal & Sagittal & \makecell{3D \\ Renderings} \\
Image & 
\makecell{\includegraphics[width=0.4\linewidth]{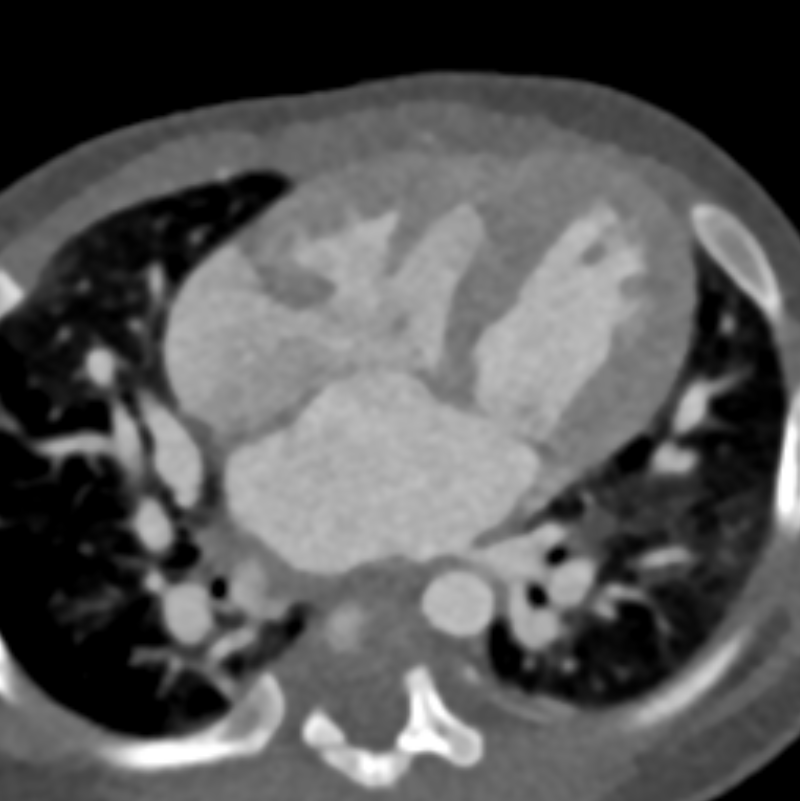}} &
\makecell{\includegraphics[width=0.4\linewidth]{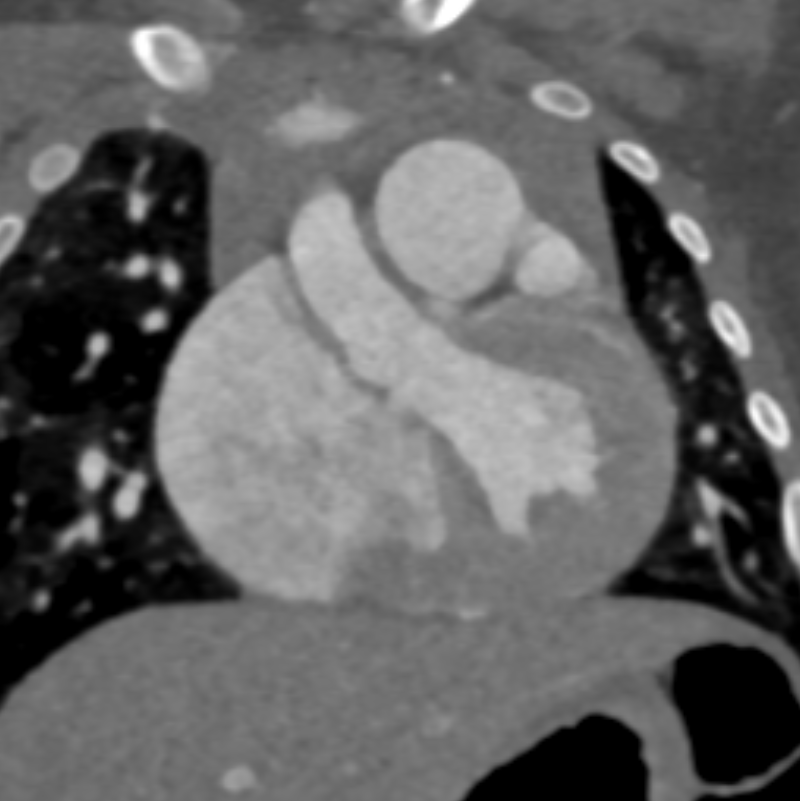}} &
\makecell{\includegraphics[width=0.4\linewidth]{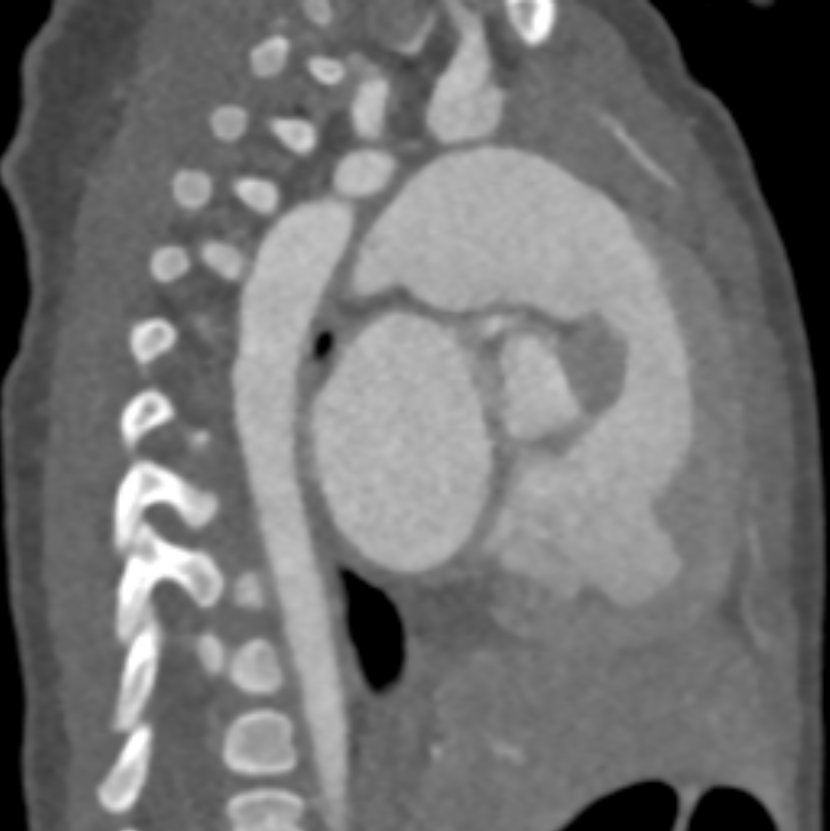}} &
\vspace{-0.03cm} \\

GT & 
\makecell{\includegraphics[width=0.4\linewidth]{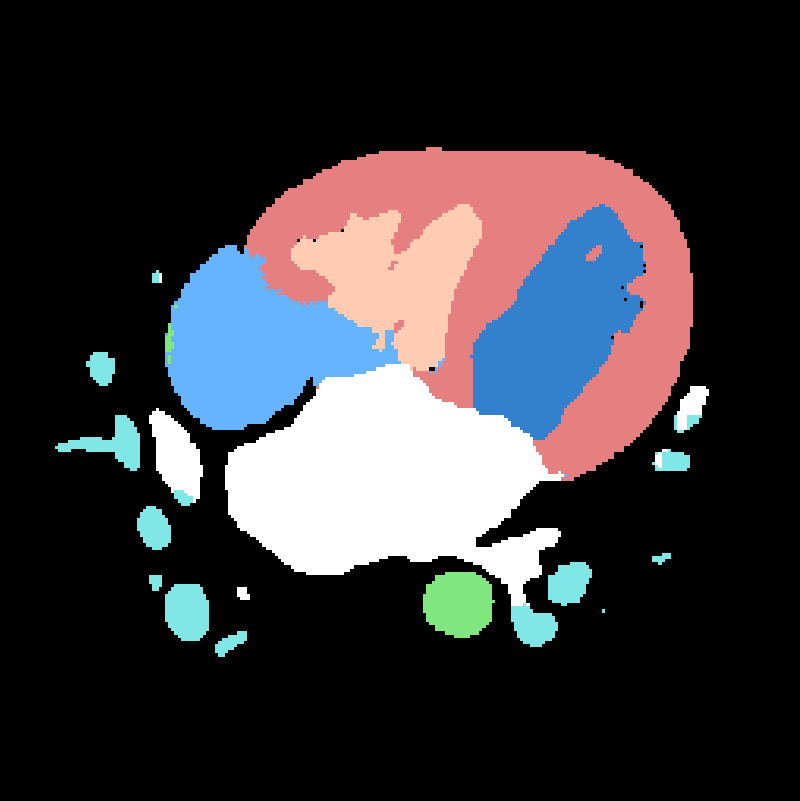}} &
\makecell{\includegraphics[width=0.4\linewidth]{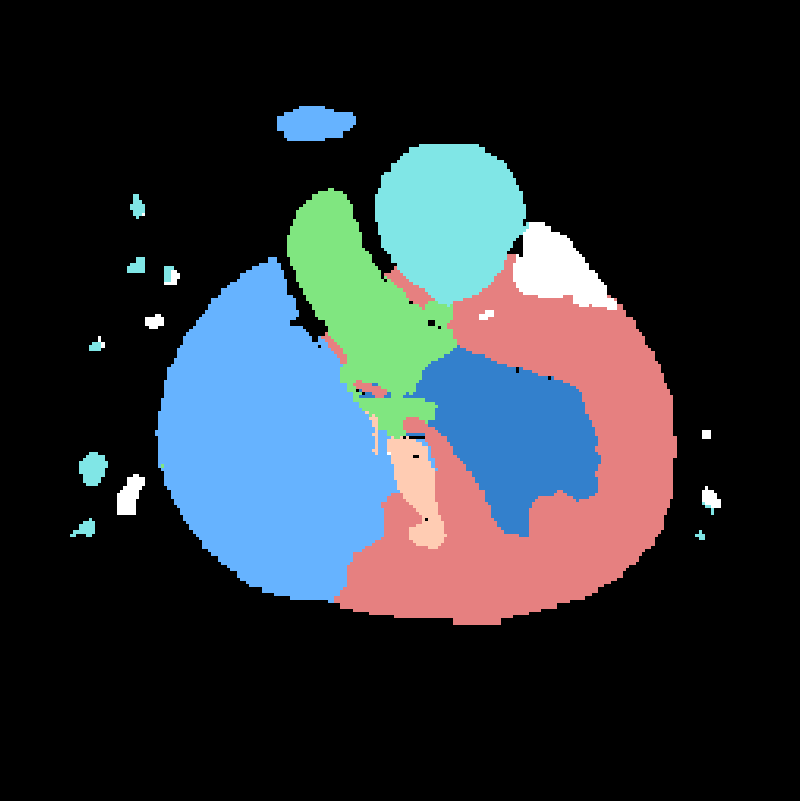}} &
\makecell{\includegraphics[width=0.4\linewidth]{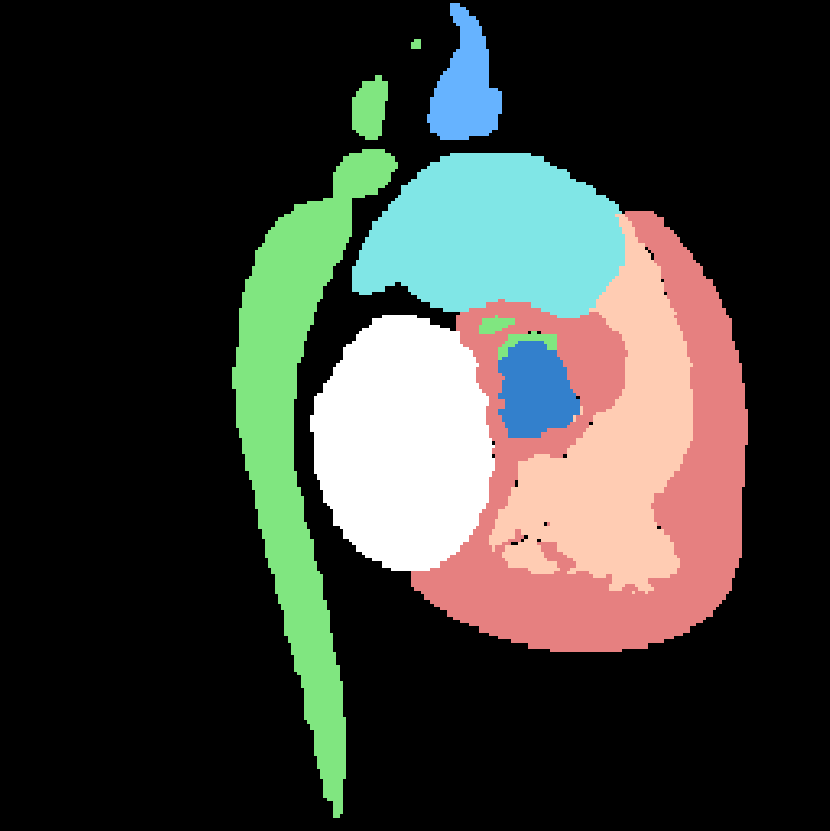}} &
\makecell{\includegraphics[width=0.4\linewidth]{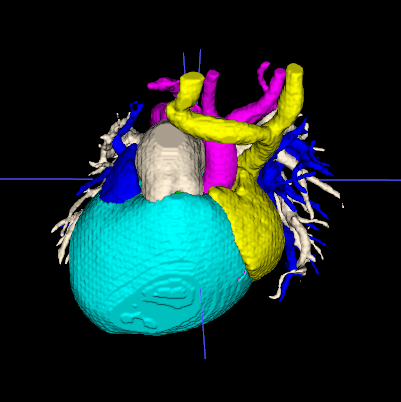}} 
\vspace{-0.03cm}\\

\makecell{MCSeg \\ (0.9247)} & 
\makecell{\includegraphics[width=0.4\linewidth]{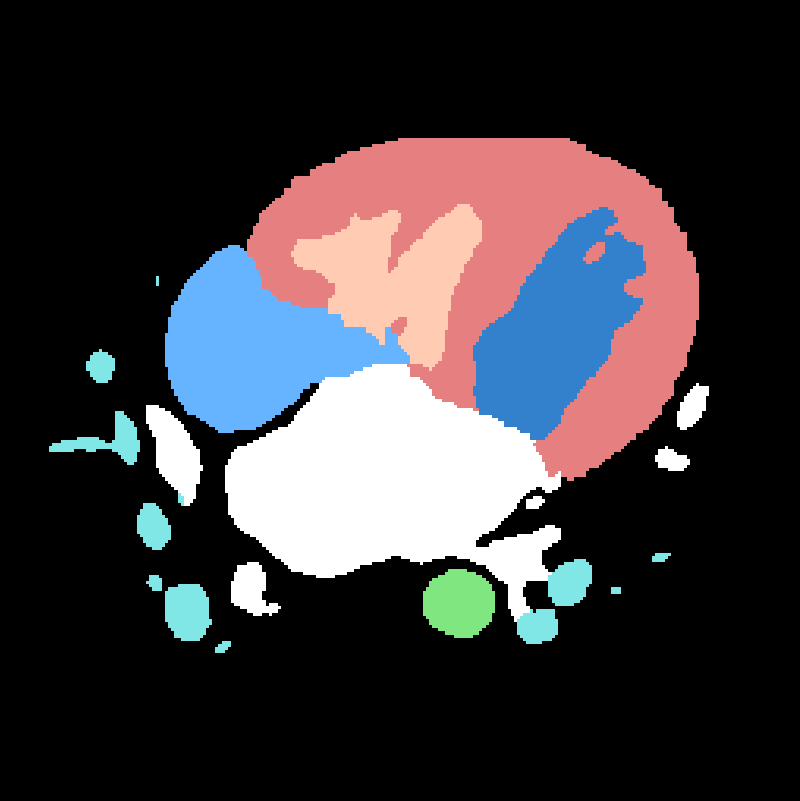}} &
\makecell{\includegraphics[width=0.4\linewidth]{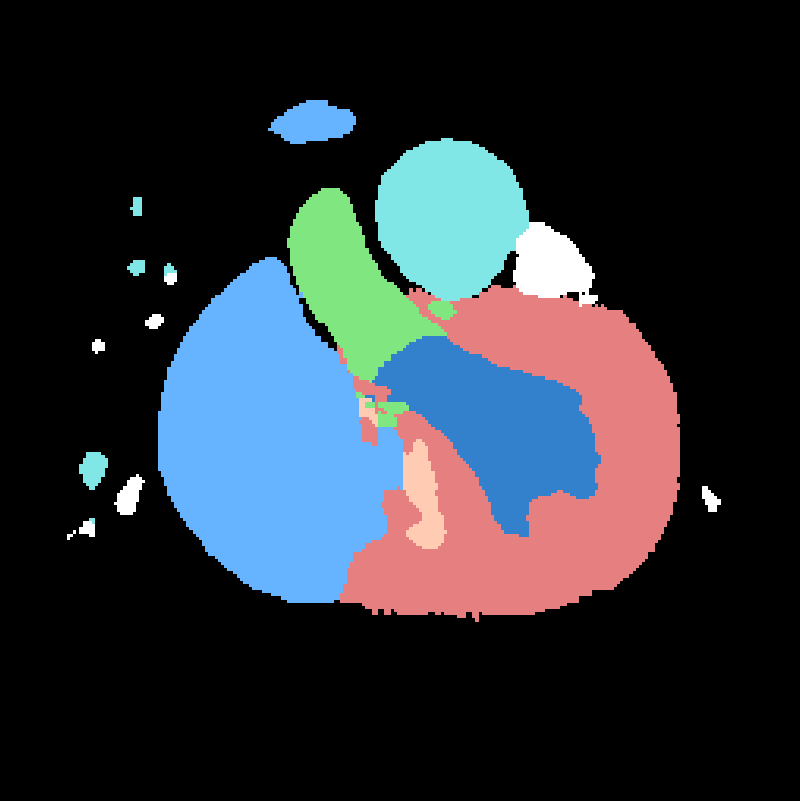}} &
\makecell{\includegraphics[width=0.4\linewidth]{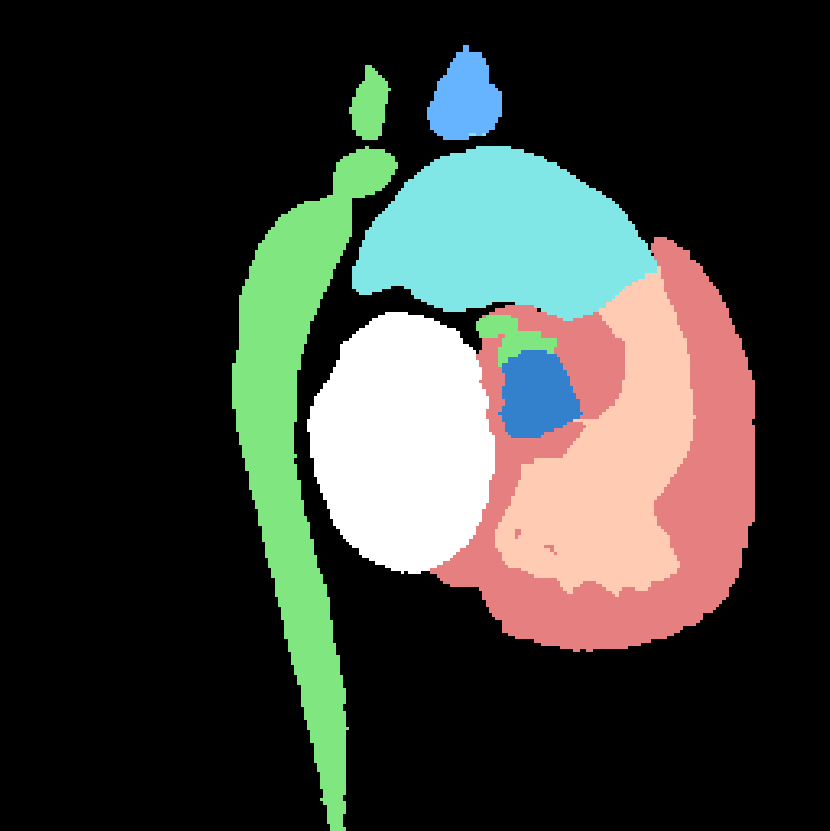}} &
\makecell{\includegraphics[width=0.4\linewidth]{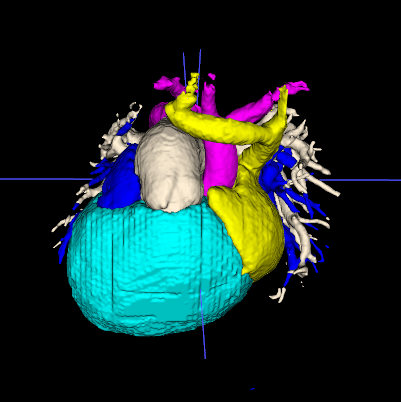}} 
\vspace{-0.03cm}\\

\makecell{CardiacSeg \\ (0.9125)} & 
\makecell{\includegraphics[width=0.4\linewidth]{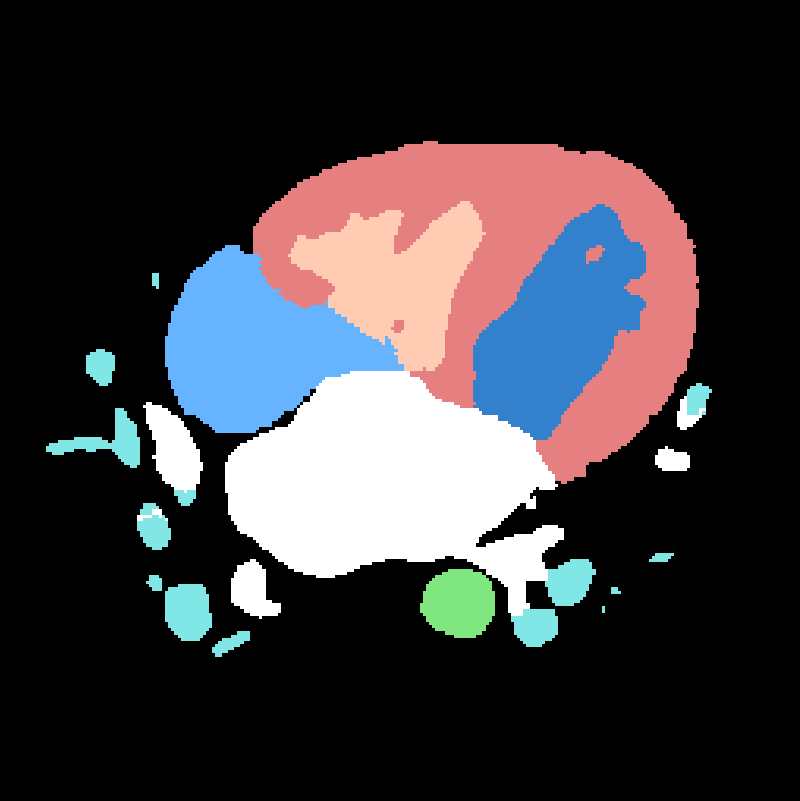}} &
\makecell{\includegraphics[width=0.4\linewidth]{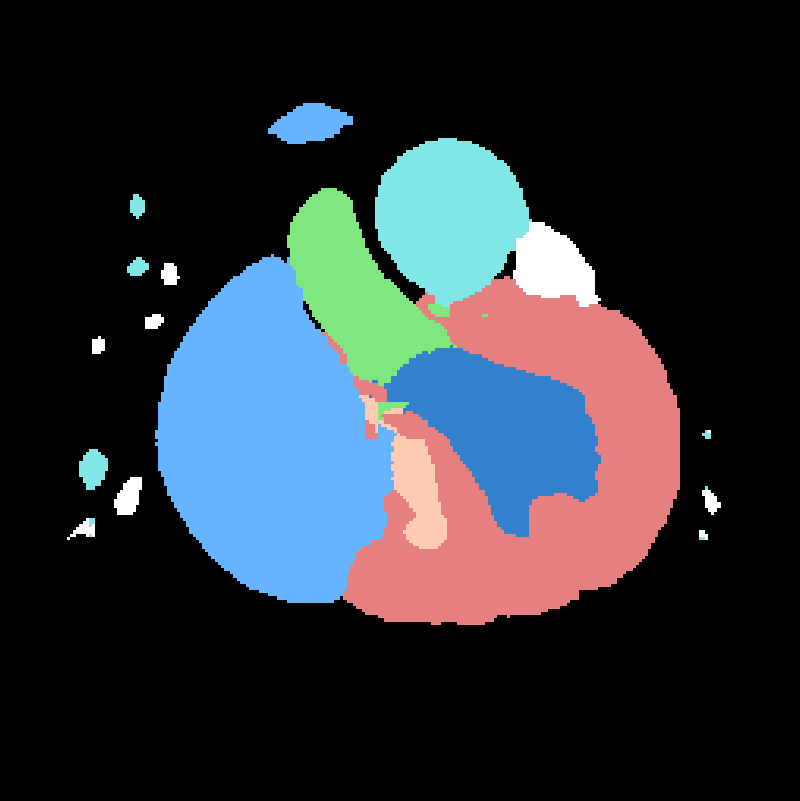}} &
\makecell{\includegraphics[width=0.4\linewidth]{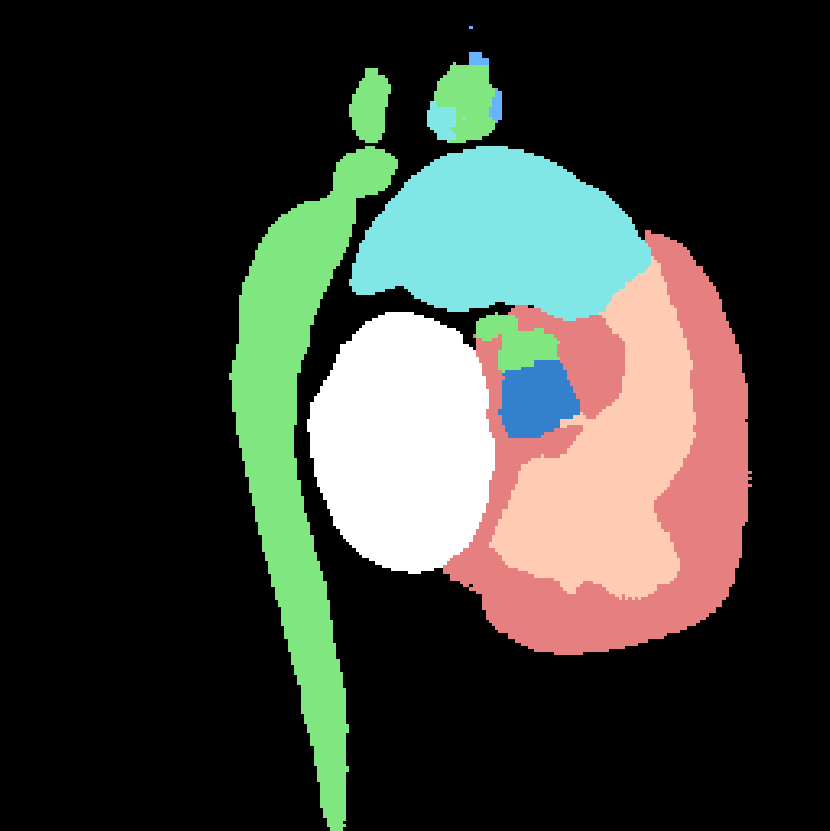}} &
\makecell{\includegraphics[width=0.4\linewidth]{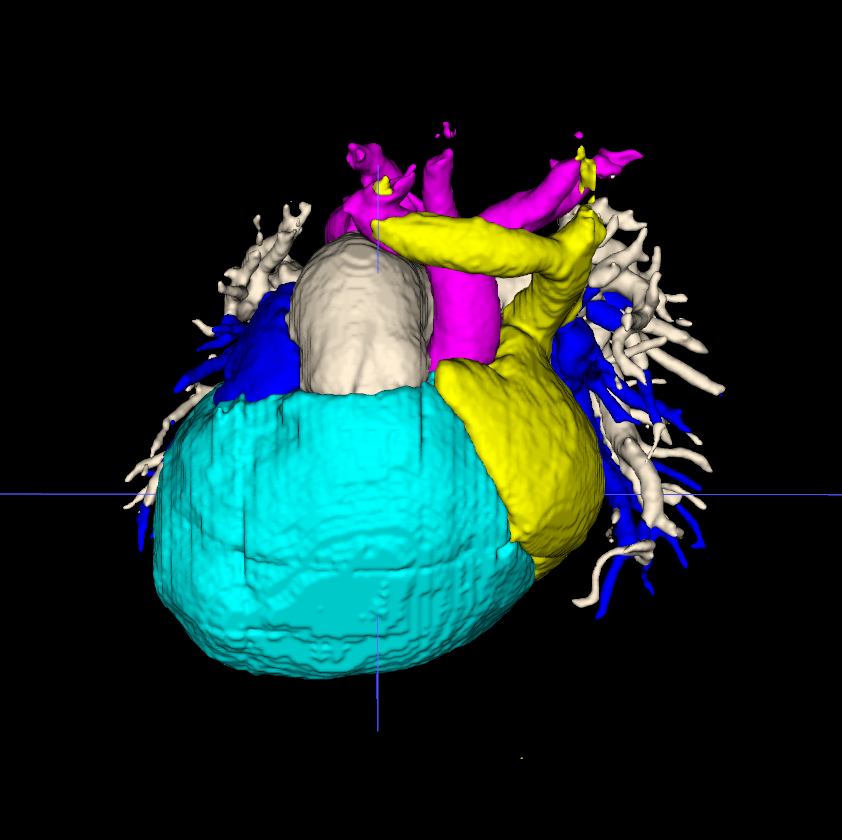}} 
\vspace{-0.03cm}\\

\makecell{MedNeXt \\ (0.8996)} &
\makecell{\includegraphics[width=0.4\linewidth]{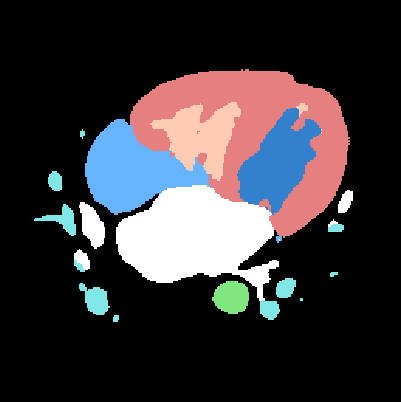}} &
\makecell{\includegraphics[width=0.4\linewidth]{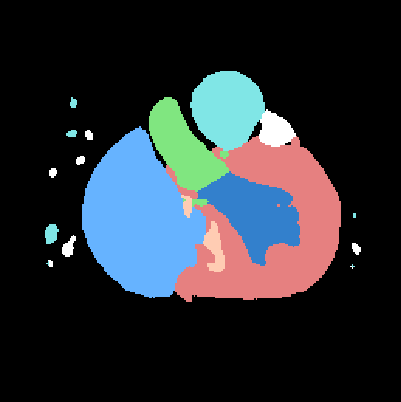}} &
\makecell{\includegraphics[width=0.4\linewidth]{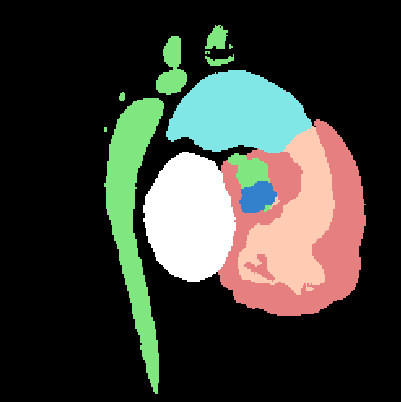}} &
\makecell{\includegraphics[width=0.4\linewidth]{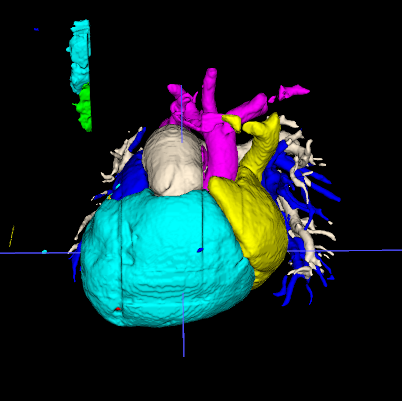}} 
\vspace{-0.03cm}\\

\makecell{SwinUNETR \\ (0.9041)} &
\makecell{\includegraphics[width=0.4\linewidth]{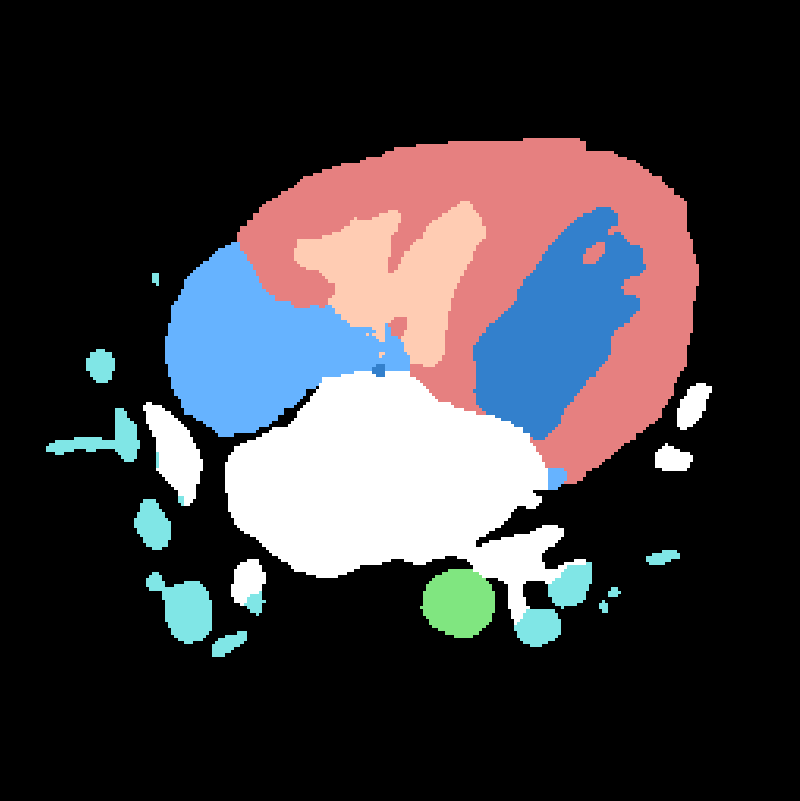}} &
\makecell{\includegraphics[width=0.4\linewidth]{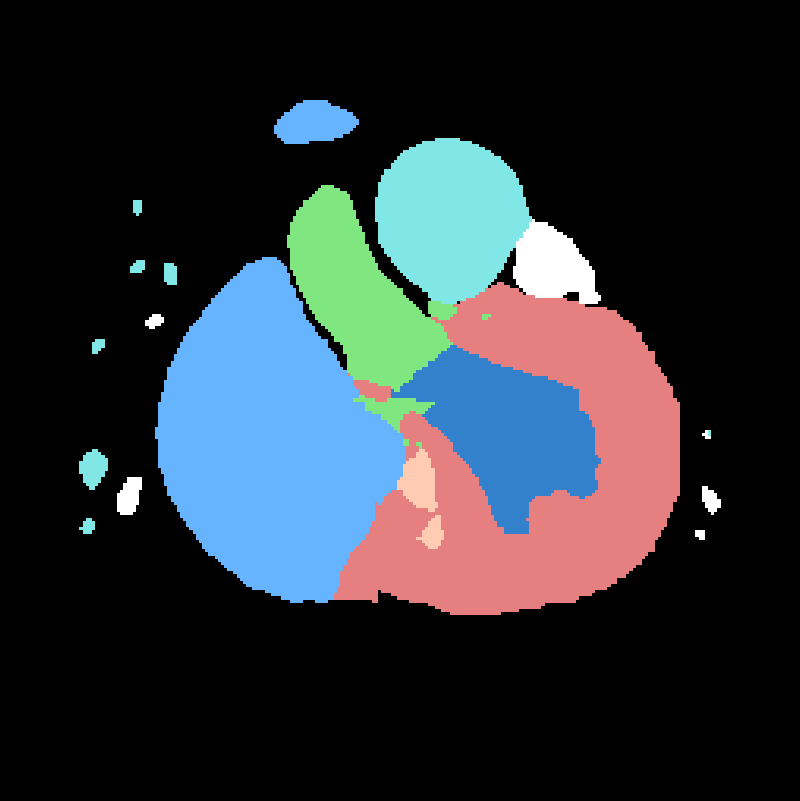}} &
\makecell{\includegraphics[width=0.4\linewidth]{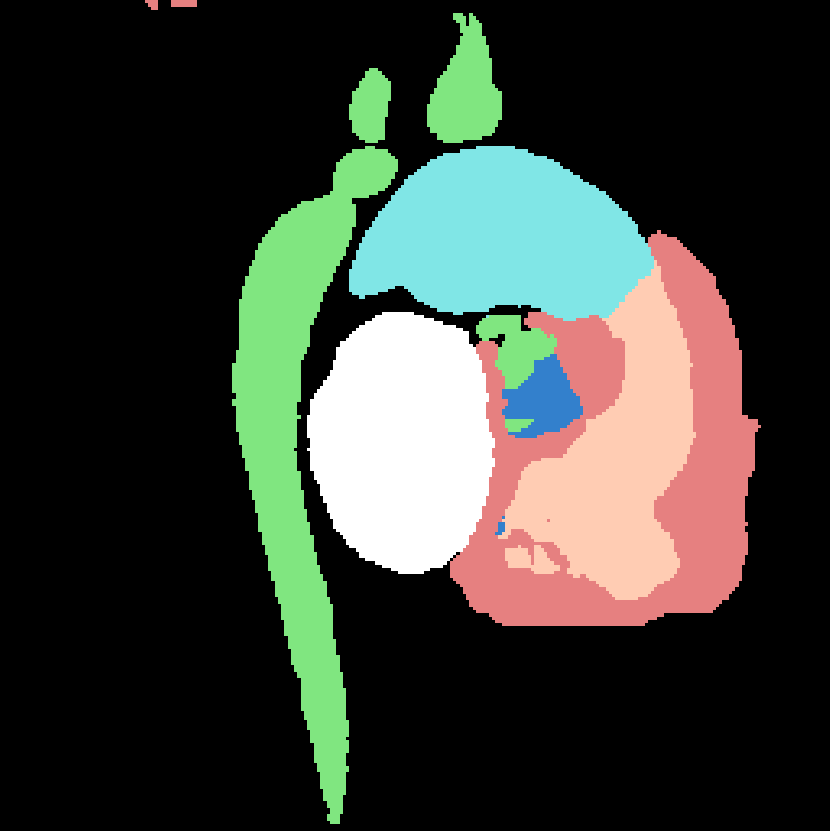}} &
\makecell{\includegraphics[width=0.4\linewidth]{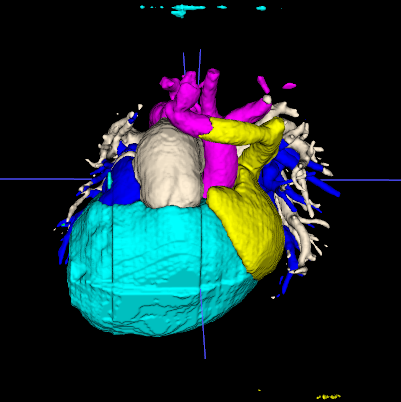}} 
\vspace{-0.03cm}\\

\makecell{CoTr \\ (0.8794)} &
\makecell{\includegraphics[width=0.4\linewidth]{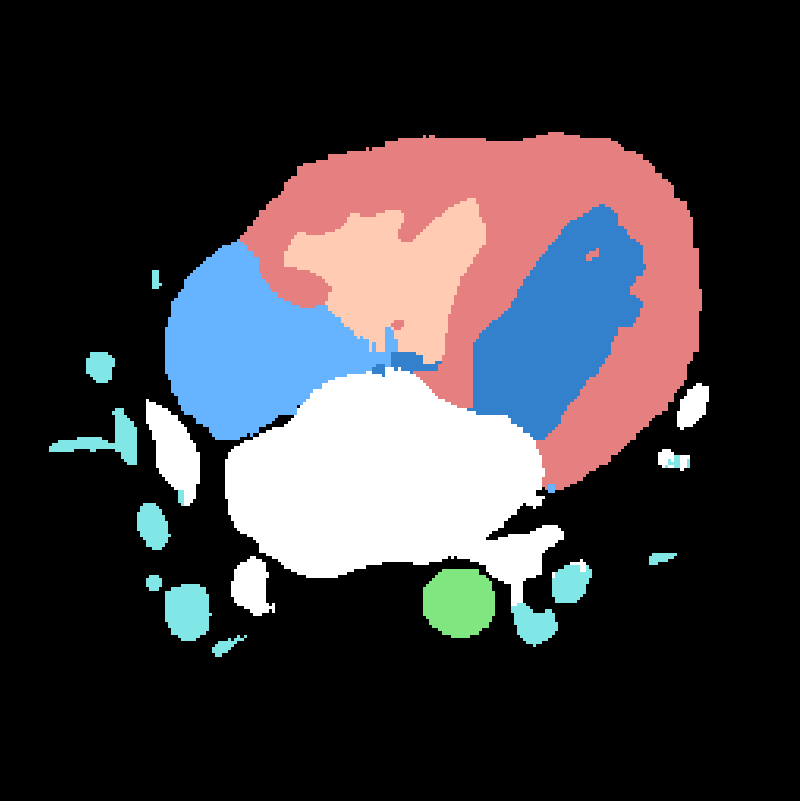}} &
\makecell{\includegraphics[width=0.4\linewidth]{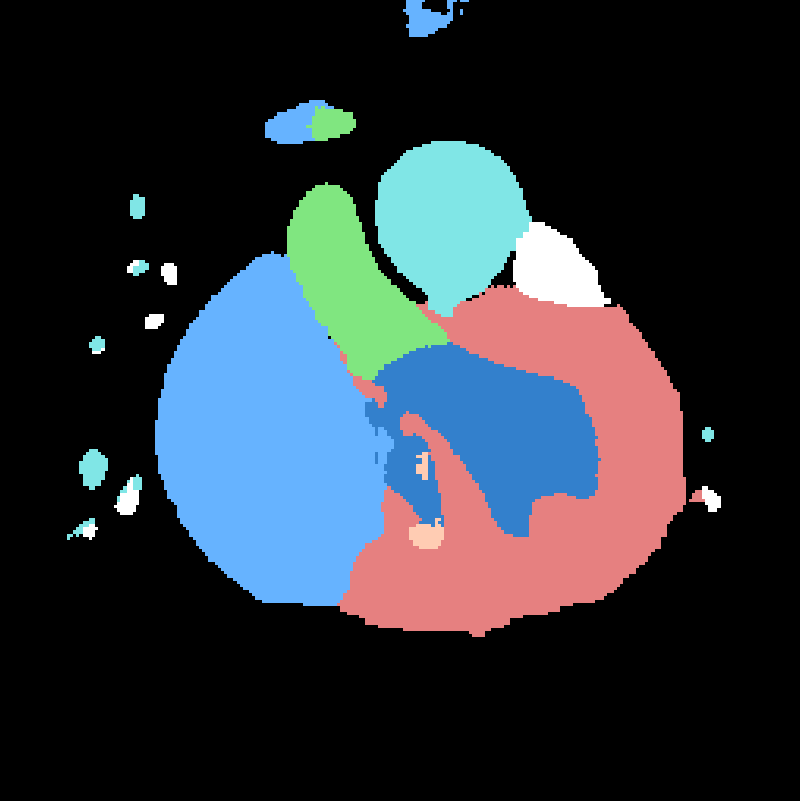}} &
\makecell{\includegraphics[width=0.4\linewidth]{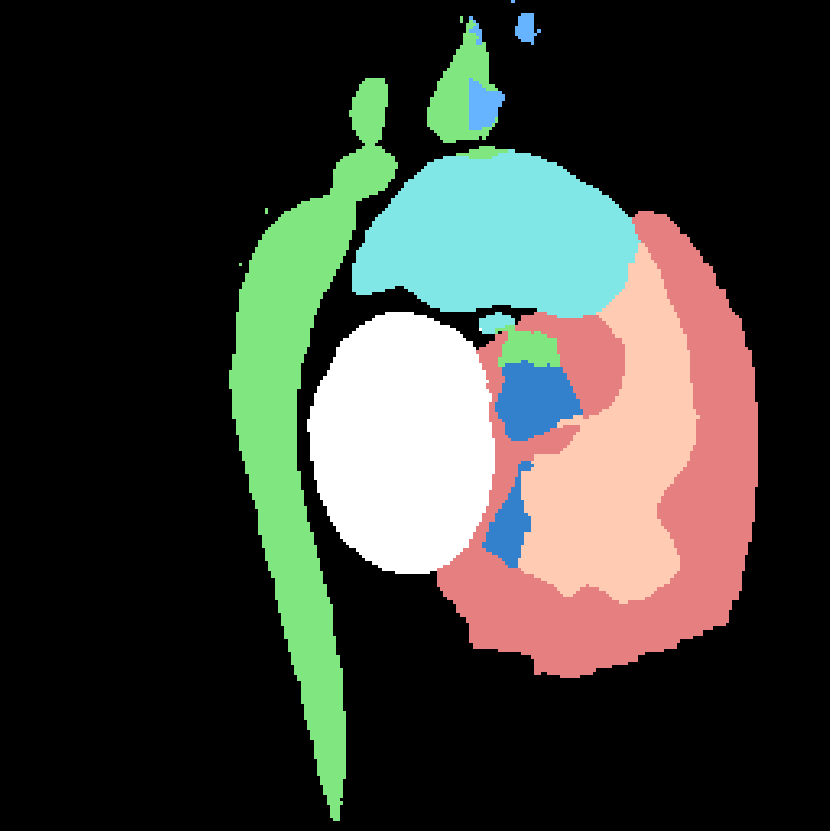}} &
\makecell{\includegraphics[width=0.4\linewidth]{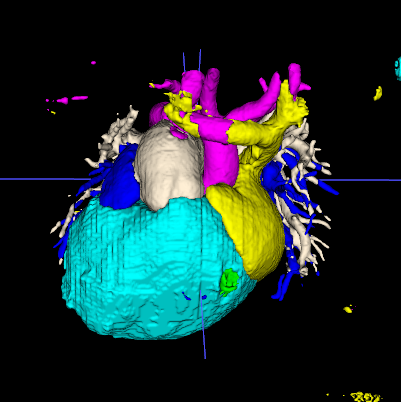}} 
\vspace{-0.03cm}\\

\end{tabular}
\end{minipage}
\vspace{-0.1cm}
\caption{Qualitative comparison for different models on image with ID ``1029" from ImageCHD dataset. 
}
\vspace{-0.4cm}
\label{fig:chd_cases}
\end{figure}

\textbf{Full-data Experiment.}
Table~\ref{tab:exp_chd} reports the overall and per-substructure average Dice scores, along with standard deviations, for each model. 
MCSeg demonstrates statistically superior performance, achieving the highest overall score and leading in five of the seven substructures. Notably, MCSeg maintains robust stability across all regions, particularly excelling on challenging substructures like the RV, AO, and PA, where competing models suffer from degraded accuracy. Qualitatively, Fig.~\ref{fig:chd_cases} shows that our method generates segmentations most closely aligned with the ground truth across three views and 3D renderings.

\begin{figure}[!t]
\begin{minipage}{0.5\linewidth}
    \centering
    \tabcolsep=0.02cm
    \scriptsize
\begin{tabular}{ccccc}
\makecell[c]{Method \\ (Dice Score)} & Axial & Coronal & Sagittal & \makecell{3D \\ Renderings} \\
Image & 
\makecell{\includegraphics[width=0.4\linewidth]{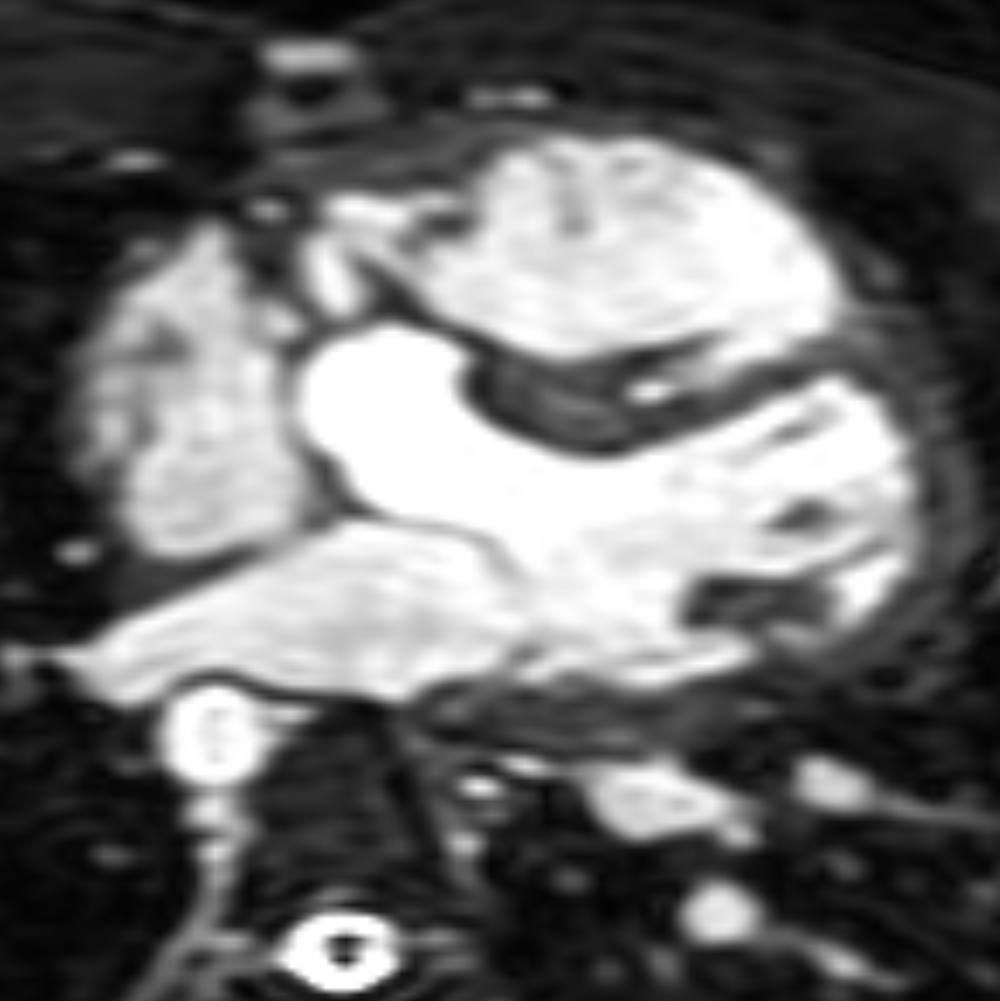}} &
\makecell{\includegraphics[width=0.4\linewidth]{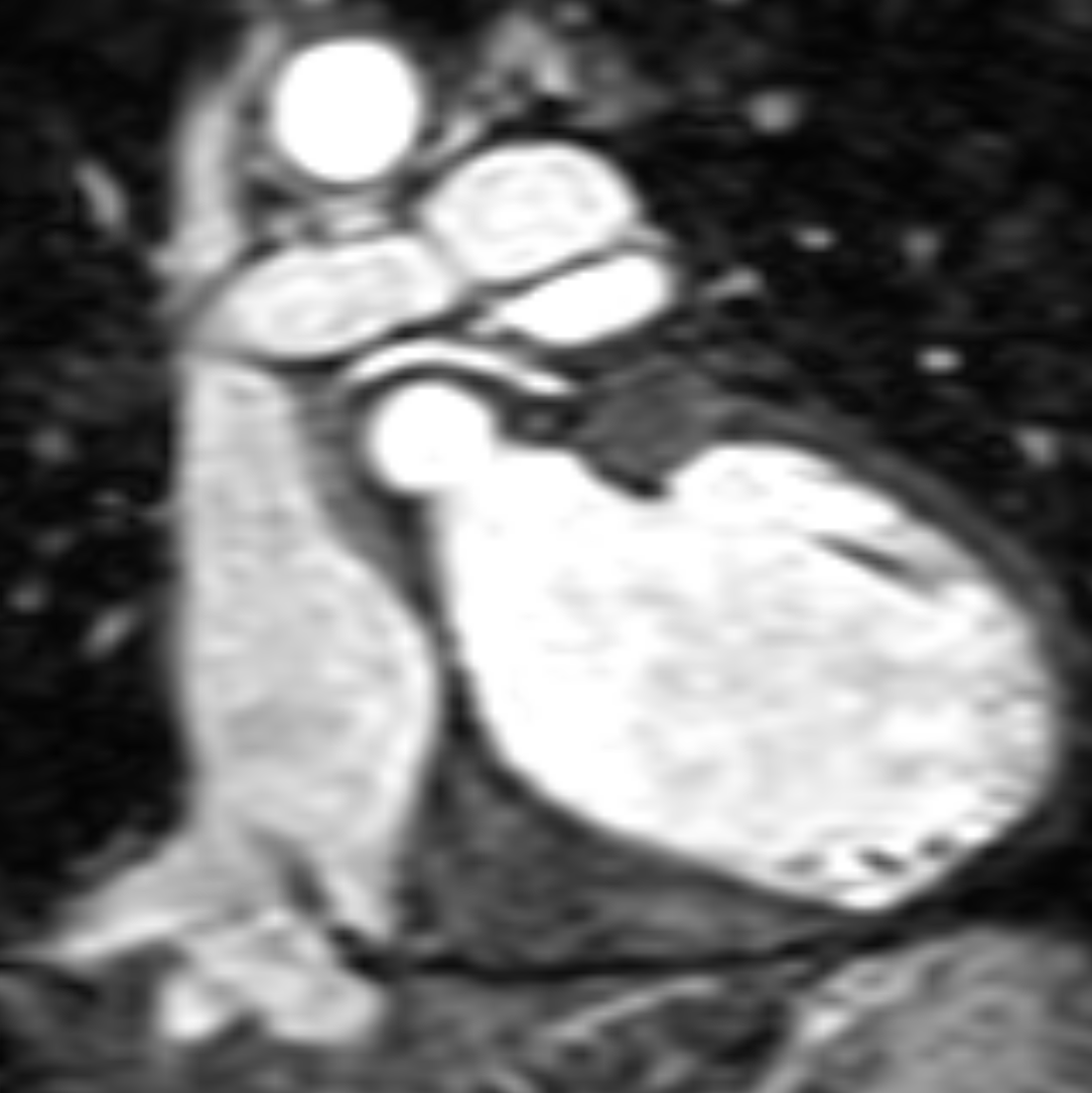}} &
\makecell{\includegraphics[width=0.4\linewidth]{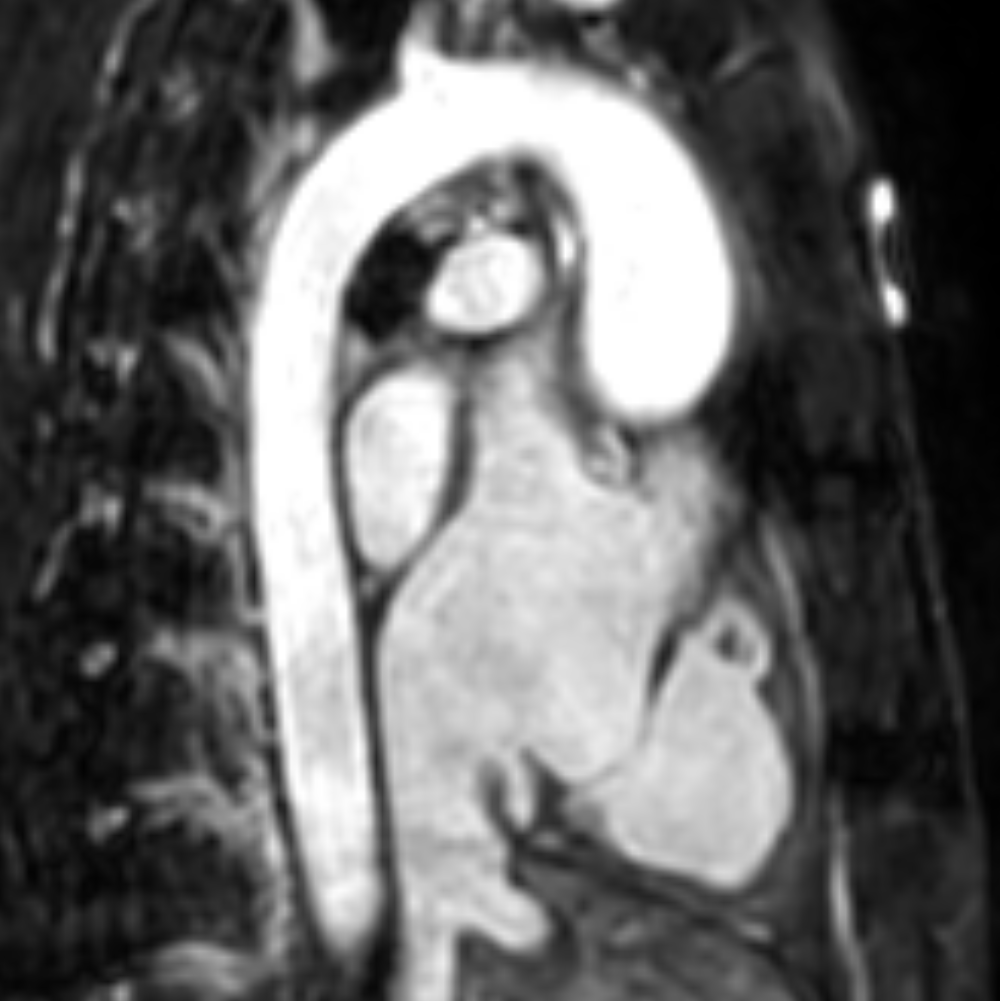}} &
\vspace{-0.03cm} \\

GT & 
\makecell{\includegraphics[width=0.4\linewidth]{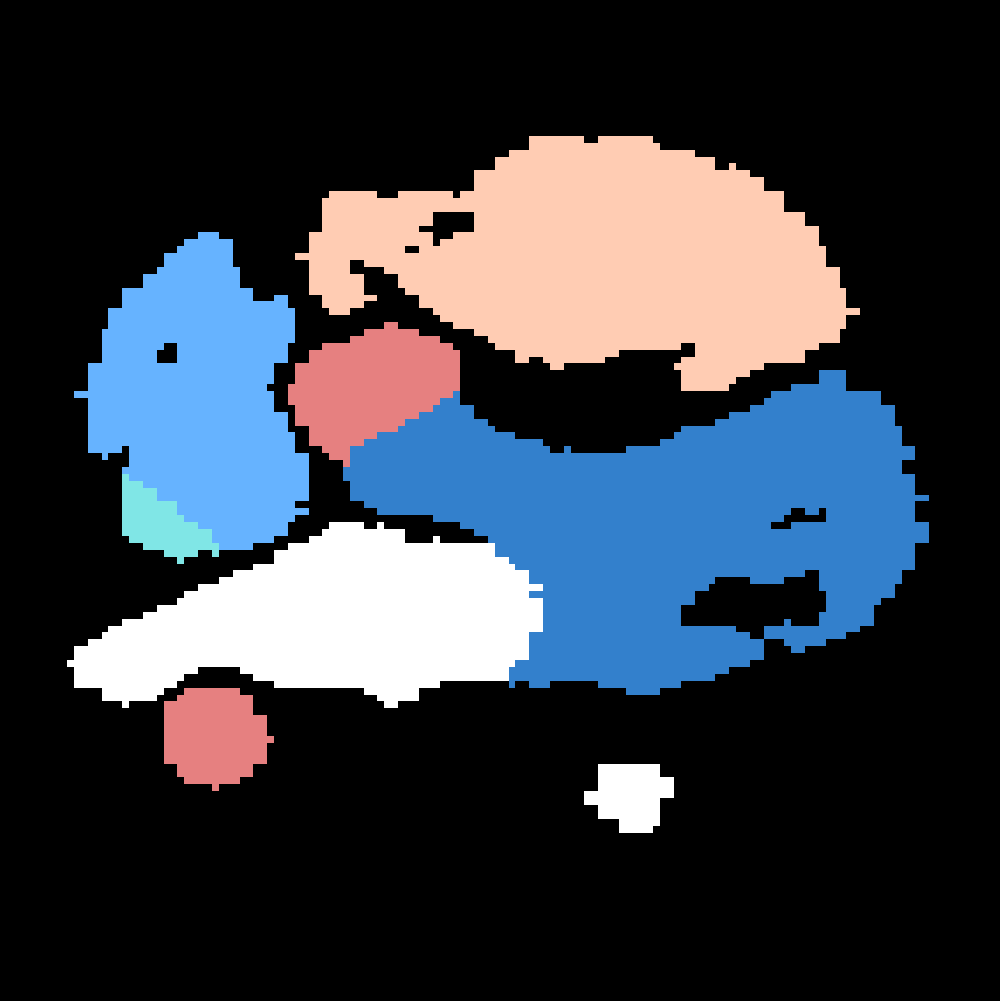}} &
\makecell{\includegraphics[width=0.4\linewidth]{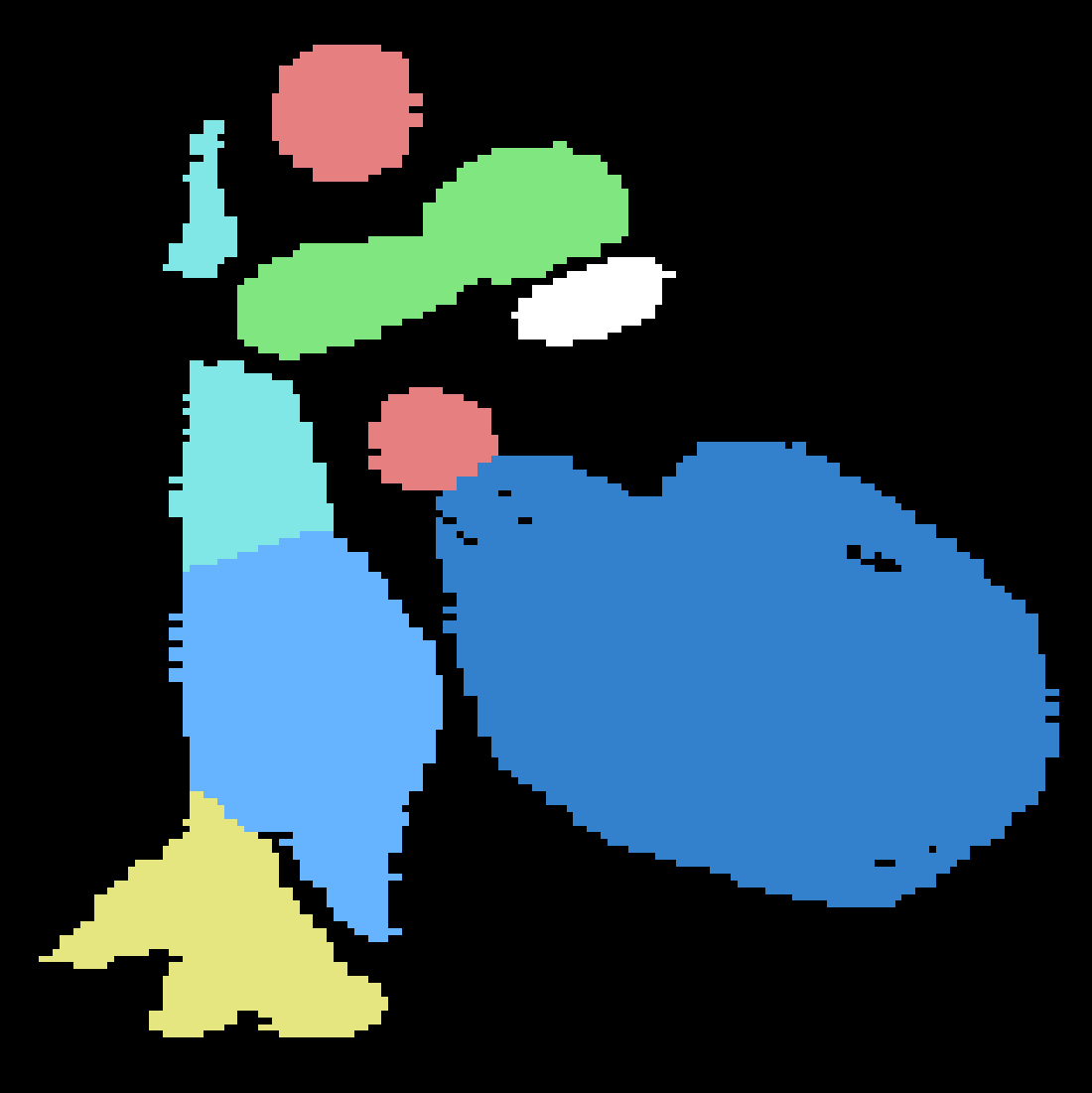}} &
\makecell{\includegraphics[width=0.4\linewidth]{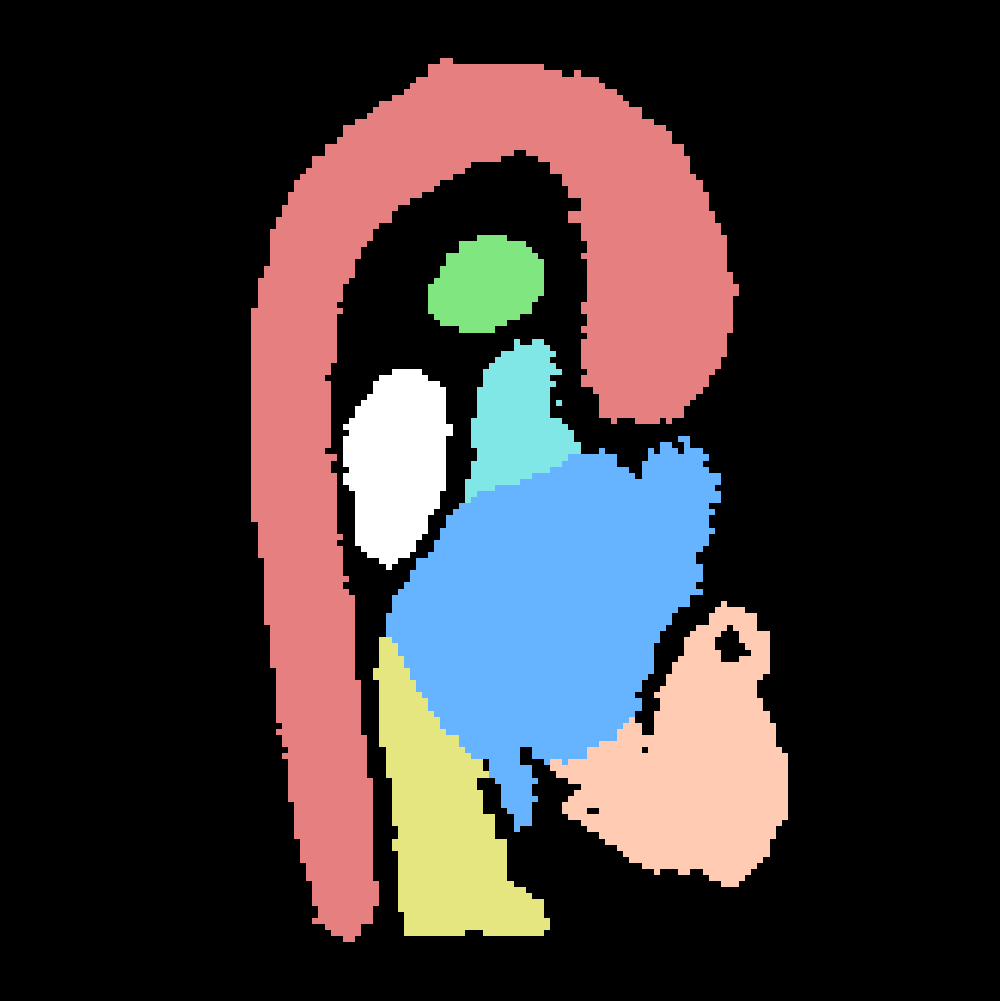}} &
\makecell{\includegraphics[width=0.4\linewidth]{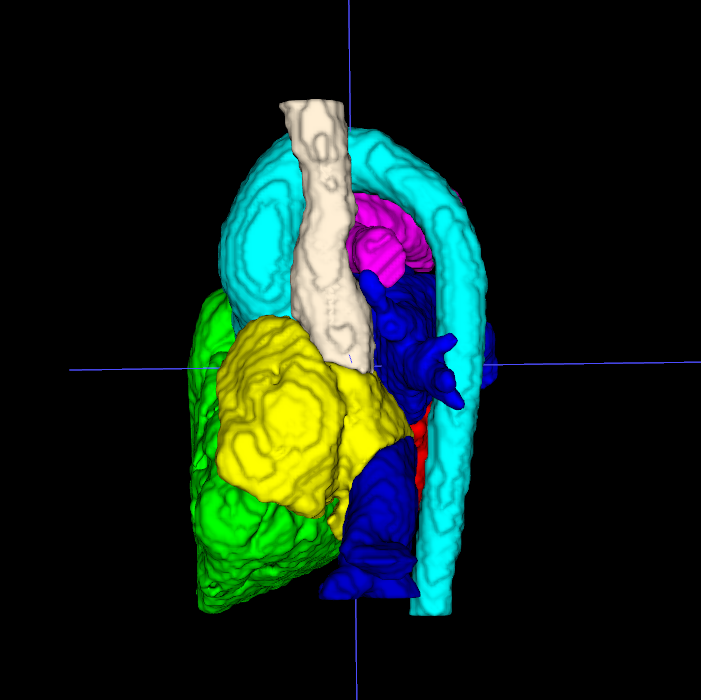}} 
\vspace{-0.03cm}\\

\makecell{MCSeg \\ (0.8737)} & 
\makecell{\includegraphics[width=0.4\linewidth]{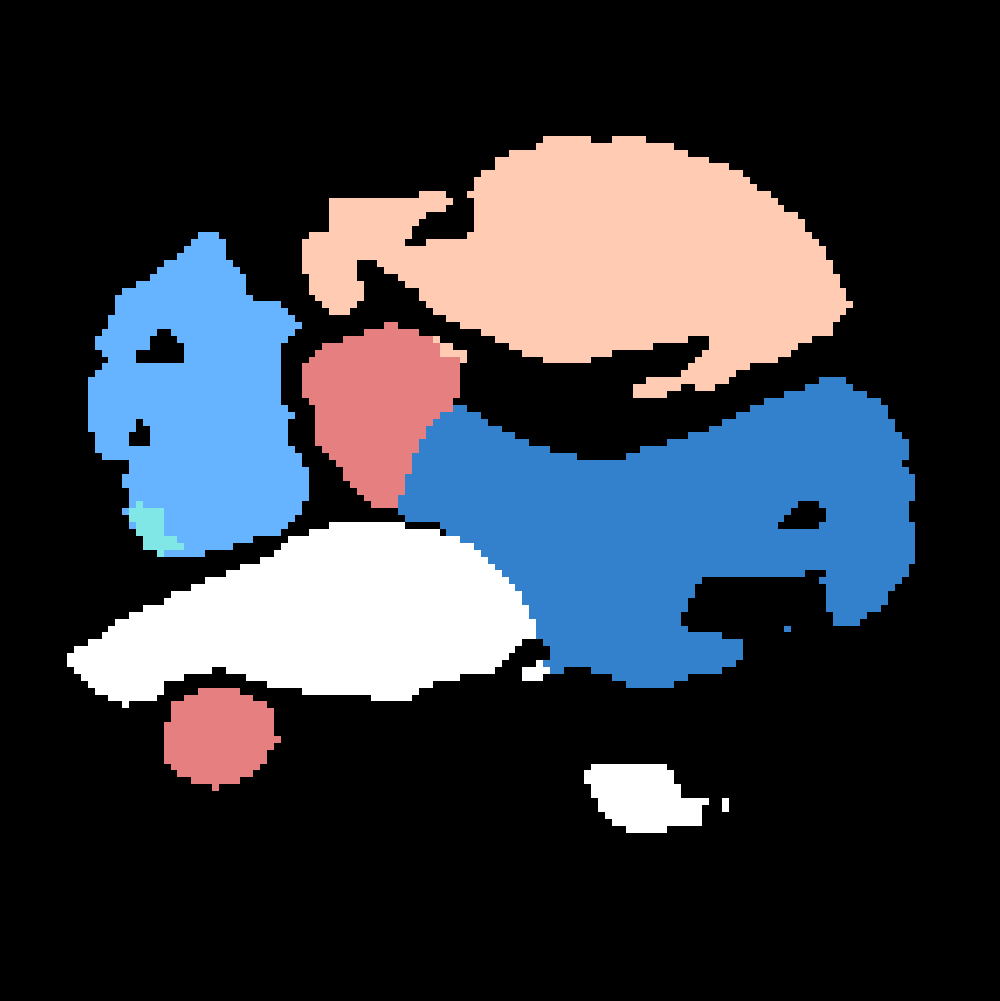}} &
\makecell{\includegraphics[width=0.4\linewidth]{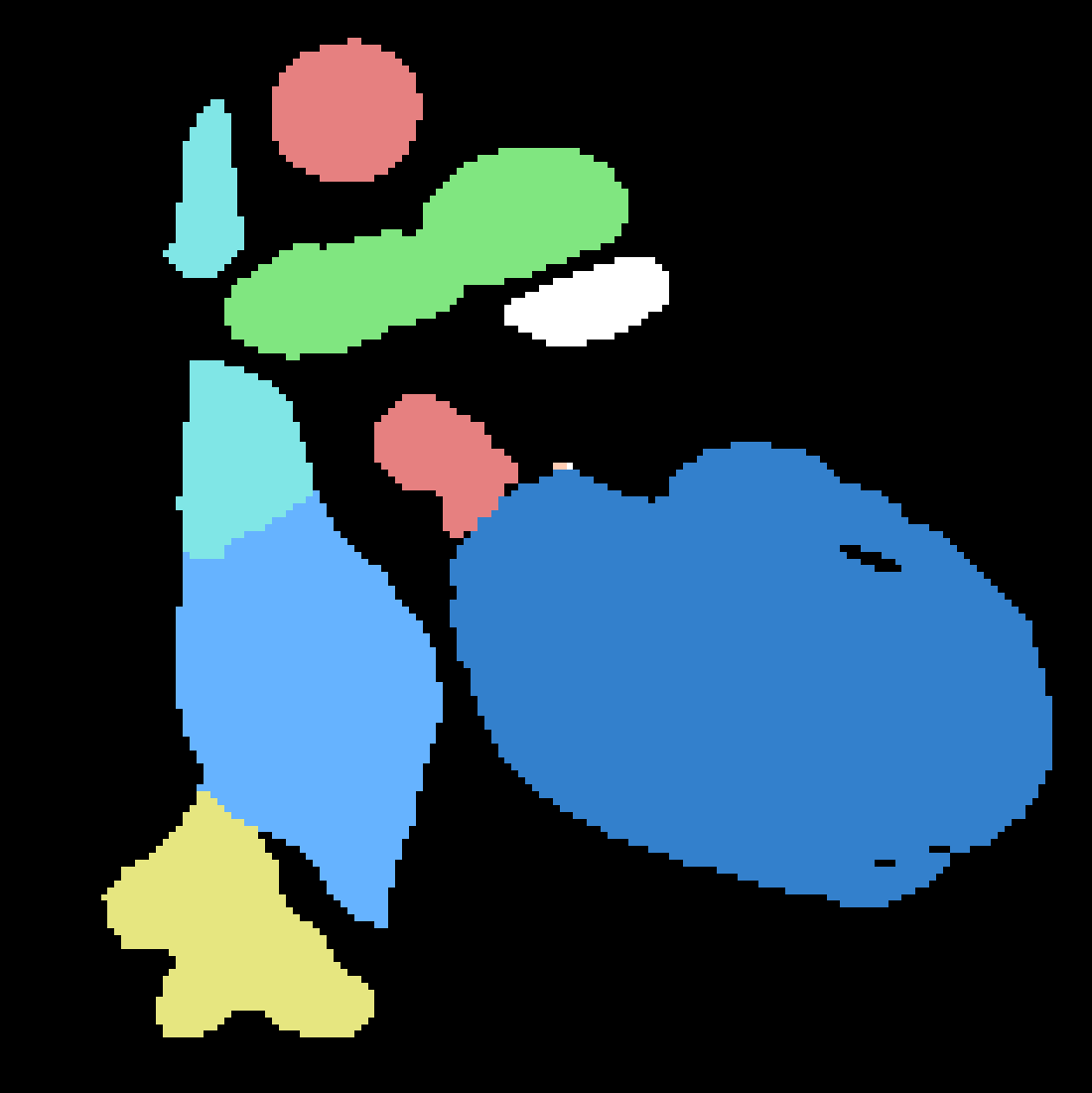}} &
\makecell{\includegraphics[width=0.4\linewidth]{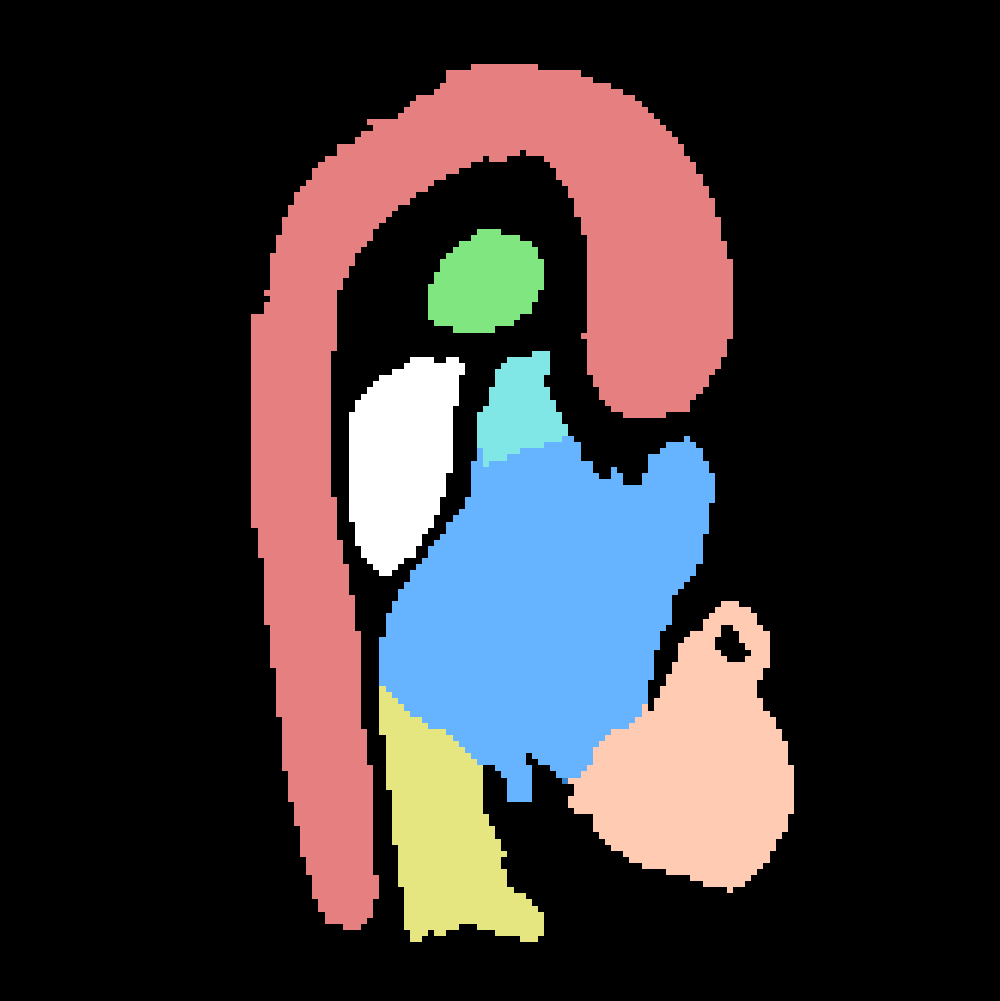}} &
\makecell{\includegraphics[width=0.4\linewidth]{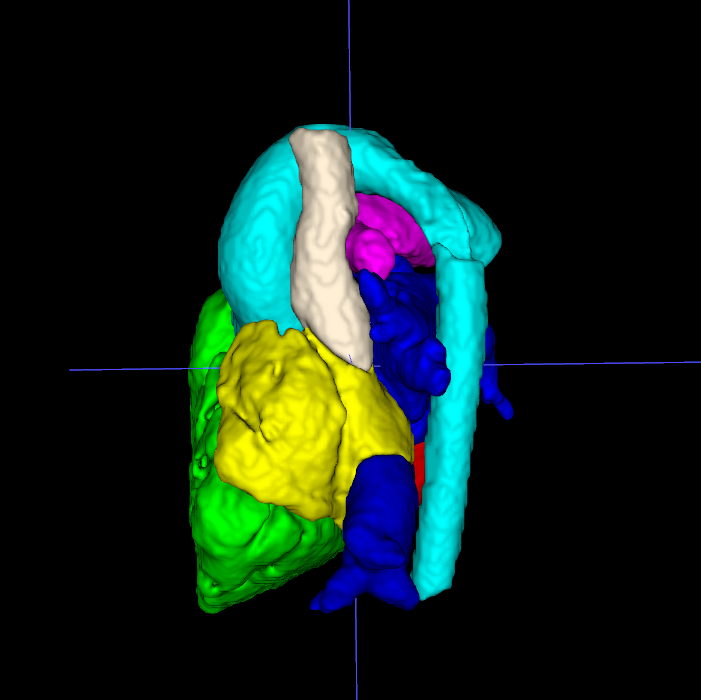}} 
\vspace{-0.03cm}\\

\makecell{CardiacSeg \\ (0.8423)} & 
\makecell{\includegraphics[width=0.4\linewidth]{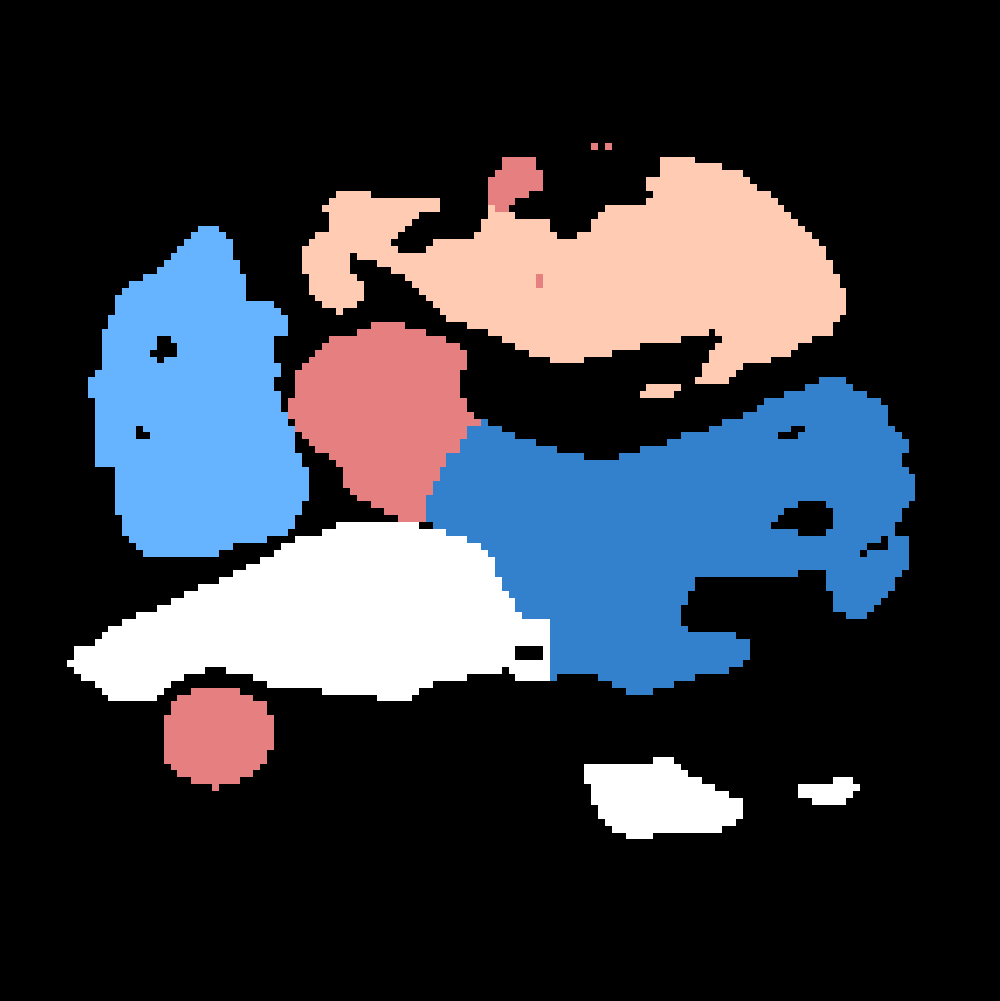}} &
\makecell{\includegraphics[width=0.4\linewidth]{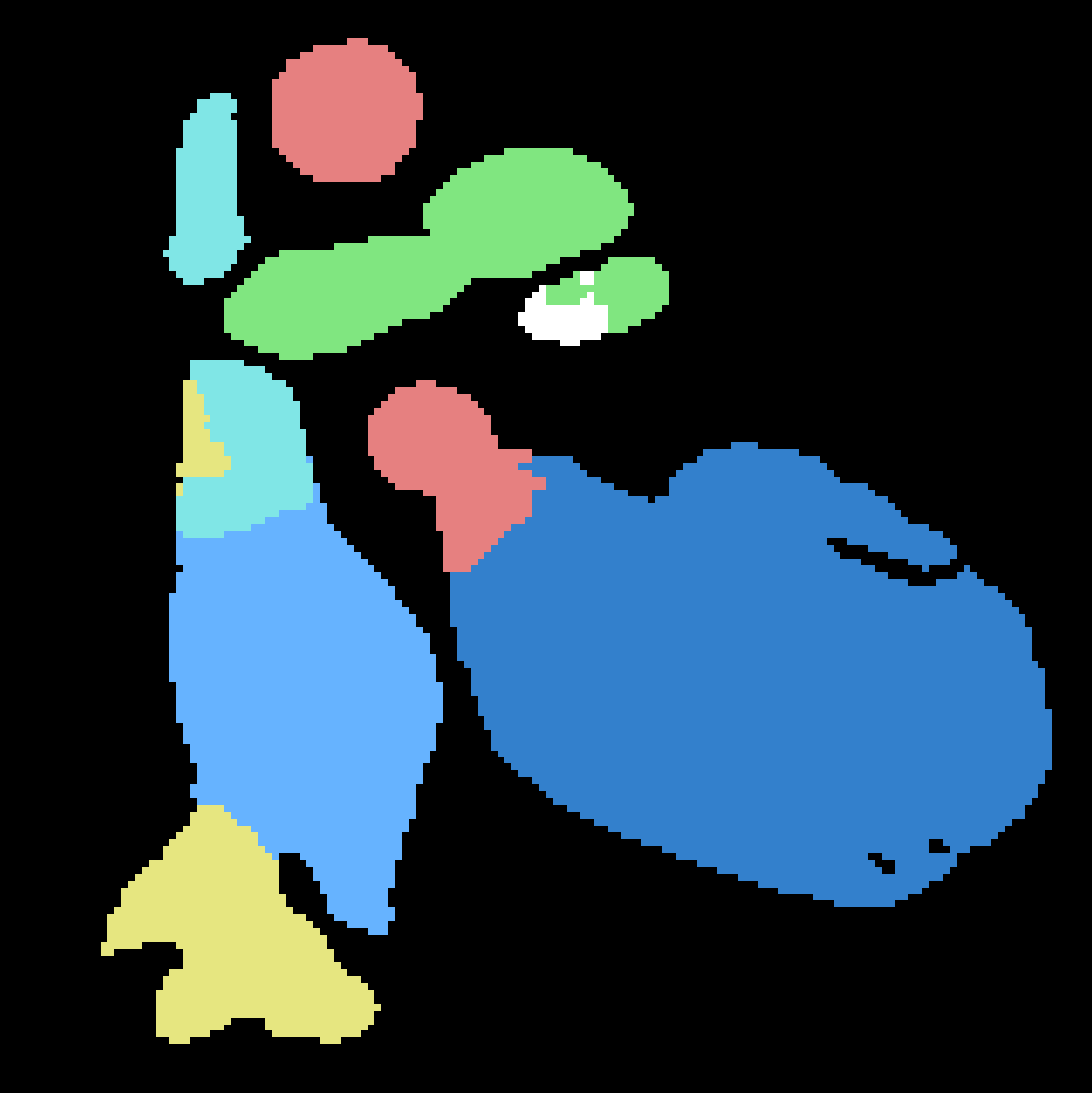}} &
\makecell{\includegraphics[width=0.4\linewidth]{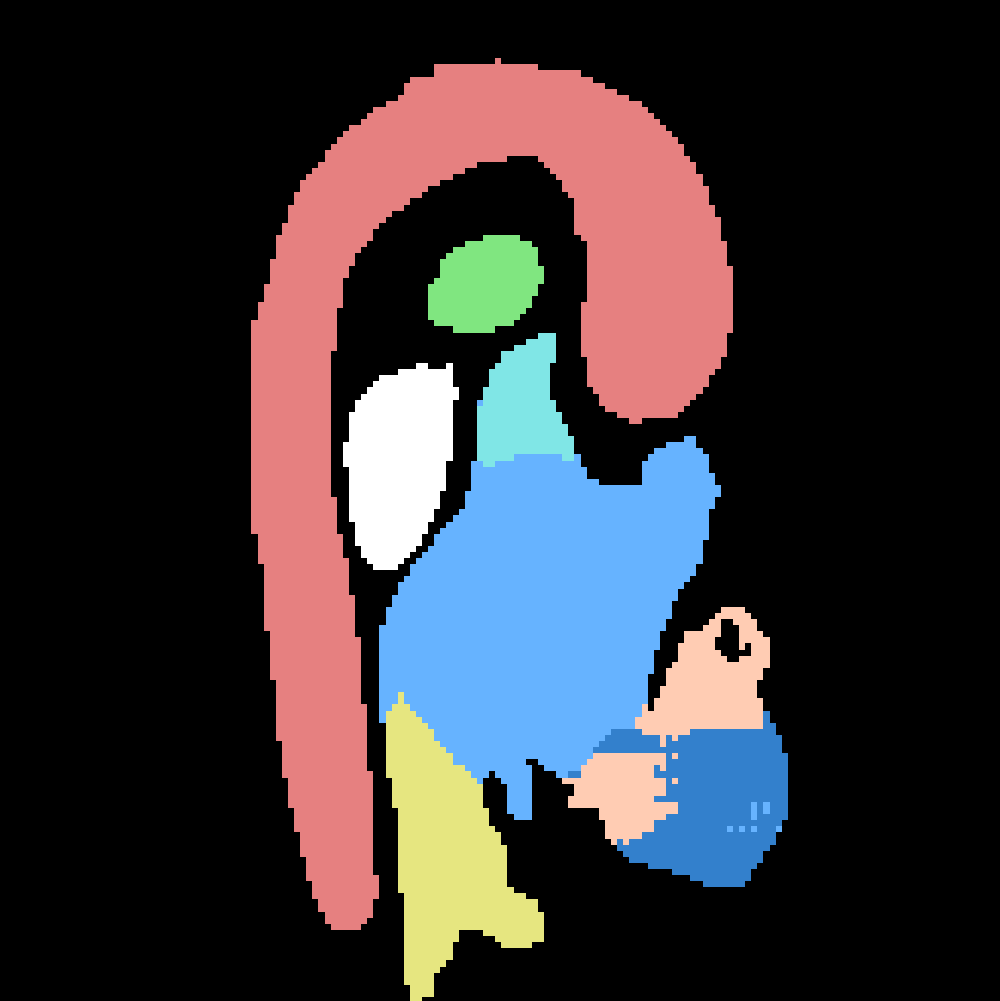}} &
\makecell{\includegraphics[width=0.4\linewidth]{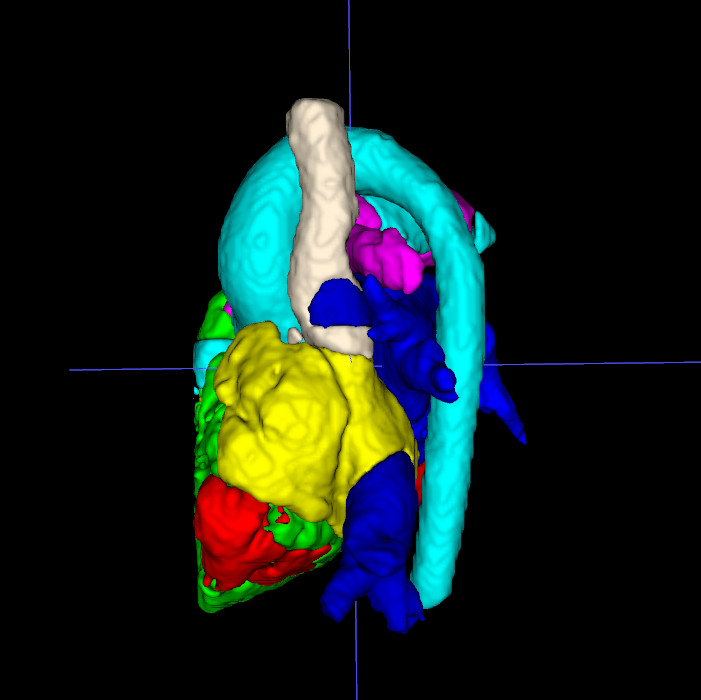}} 
\vspace{-0.03cm}\\

\makecell{MedNeXt \\ (0.8321)} &
\makecell{\includegraphics[width=0.4\linewidth]{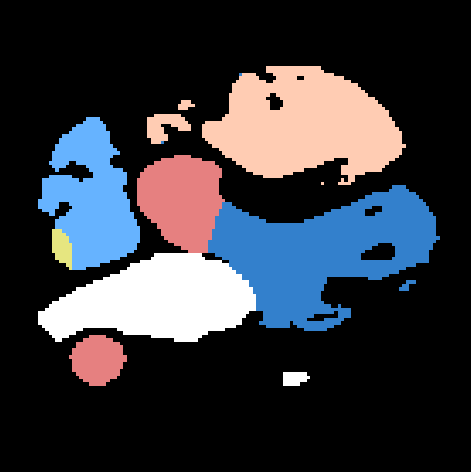}} &
\makecell{\includegraphics[width=0.4\linewidth]{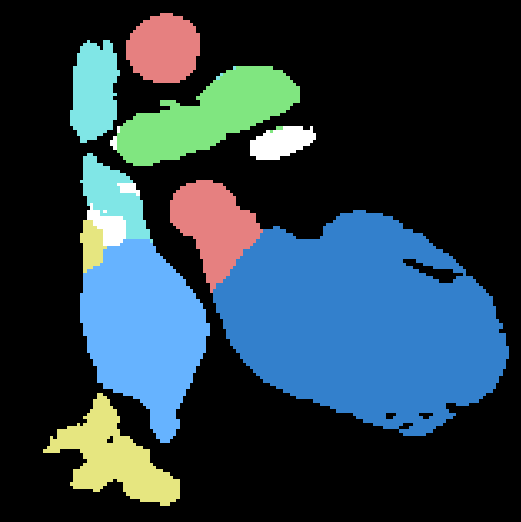}} &
\makecell{\includegraphics[width=0.4\linewidth]{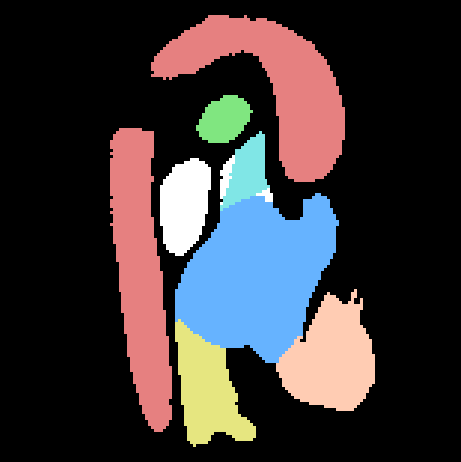}} &
\makecell{\includegraphics[width=0.4\linewidth]{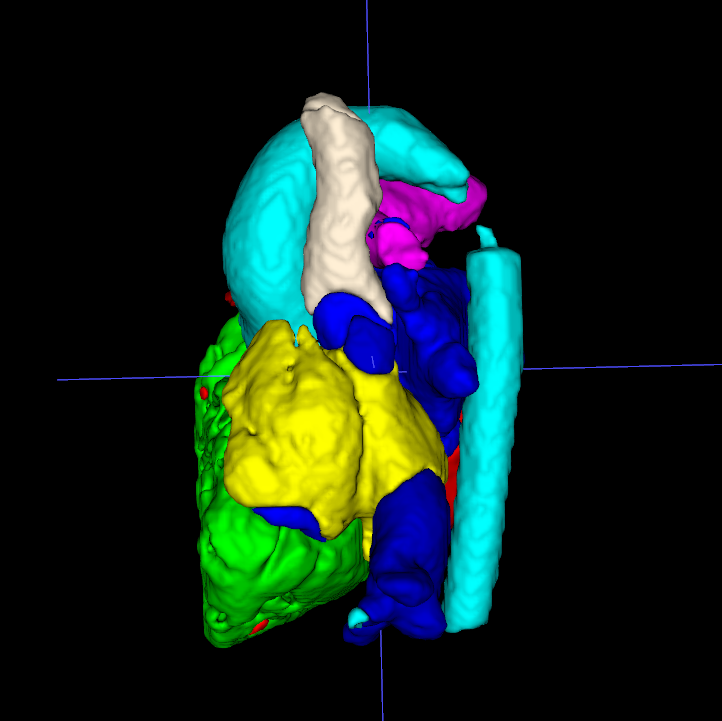}} 
\vspace{-0.03cm}\\

\makecell{SwinUNETR \\ (0.8130)} &
\makecell{\includegraphics[width=0.4\linewidth]{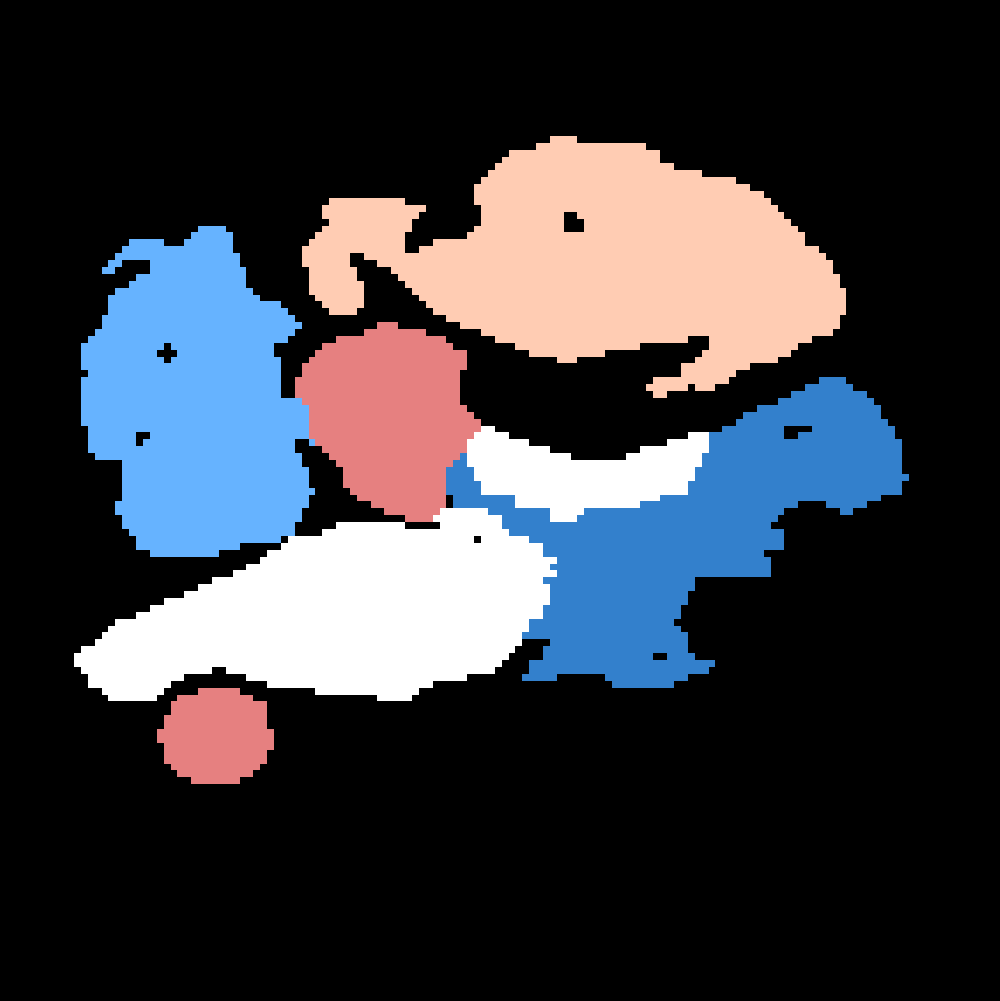}} &
\makecell{\includegraphics[width=0.4\linewidth]{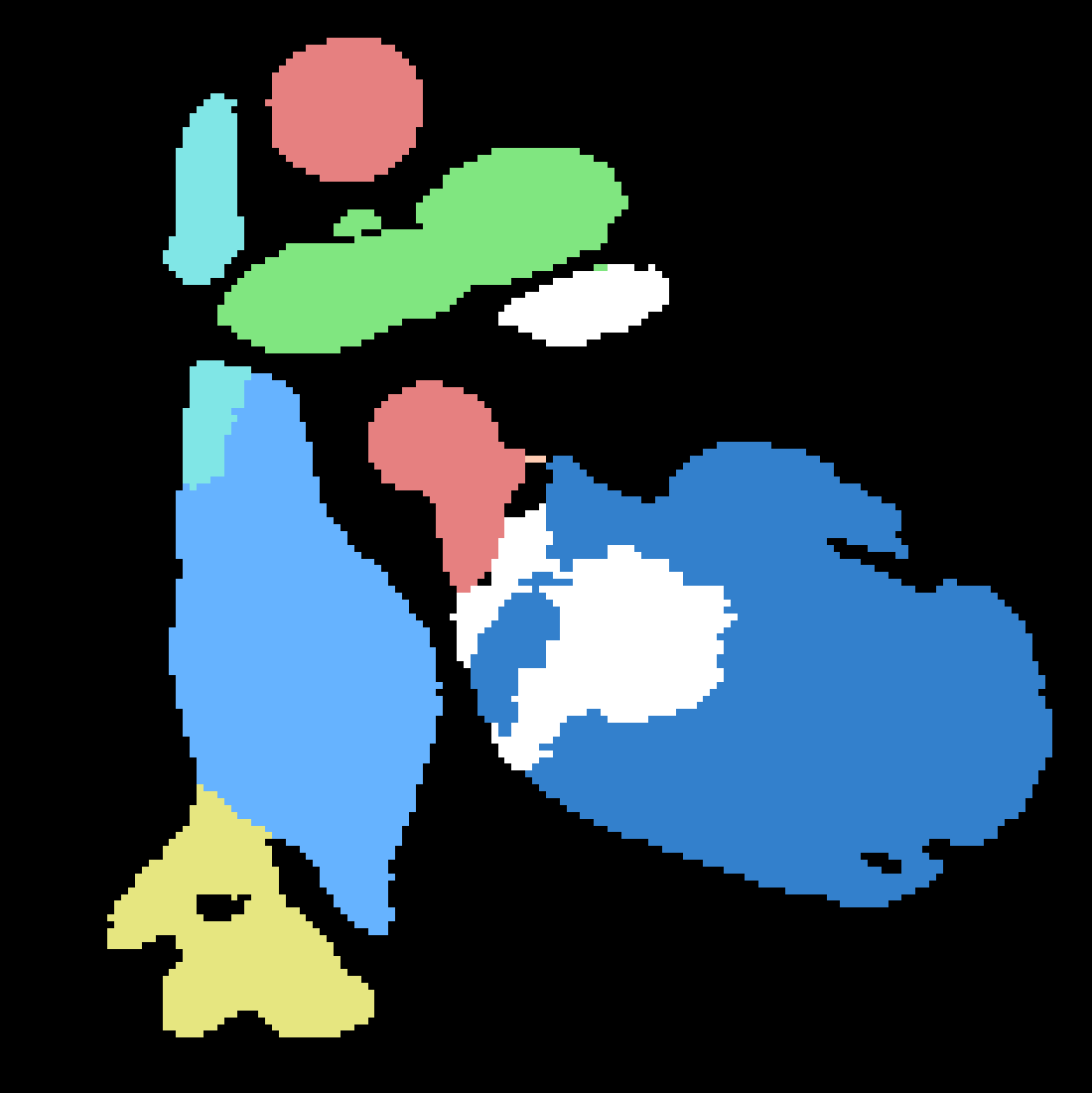}} &
\makecell{\includegraphics[width=0.4\linewidth]{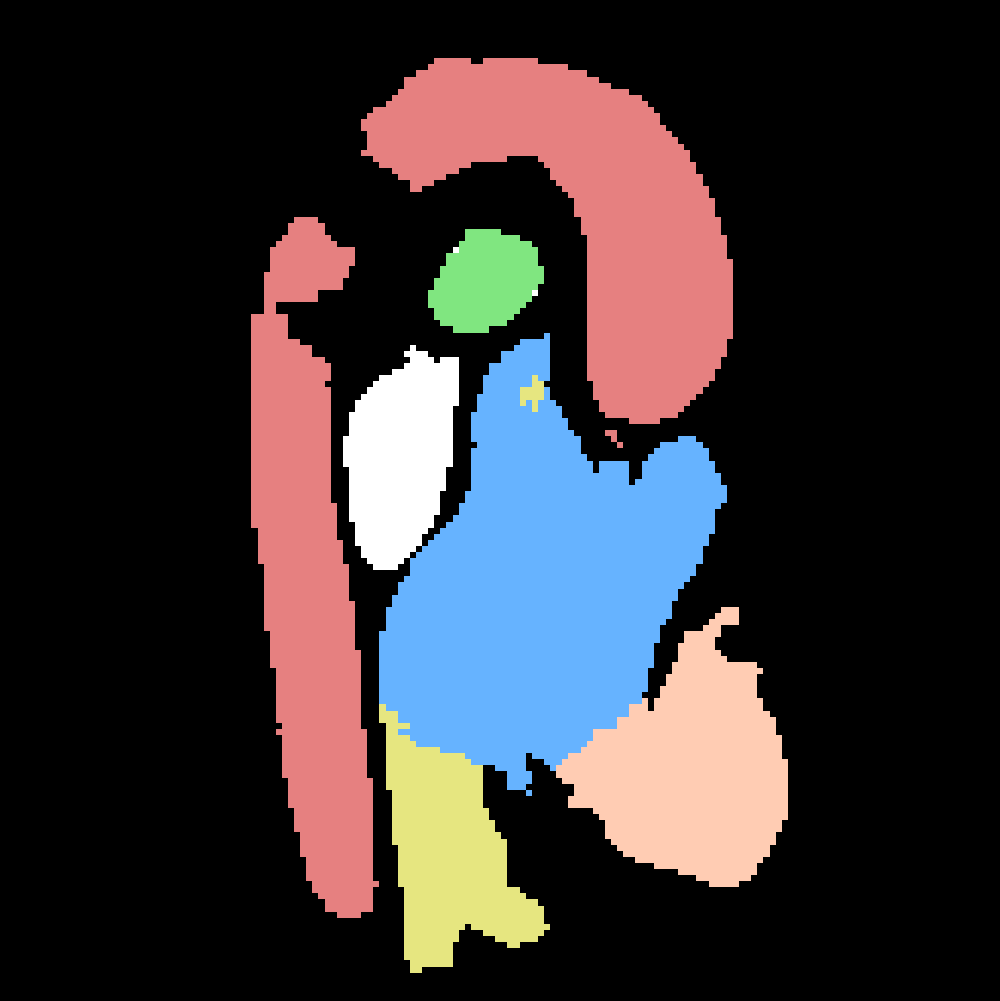}} &
\makecell{\includegraphics[width=0.4\linewidth]{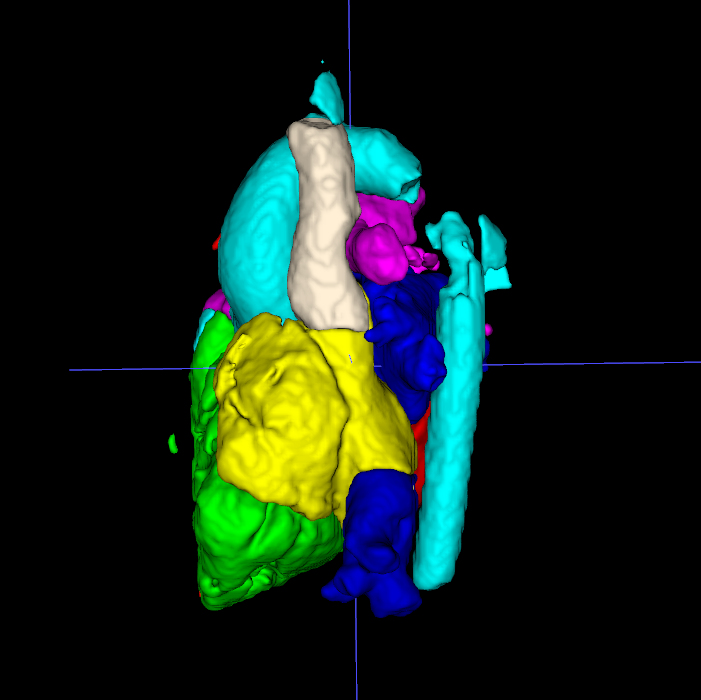}} 
\vspace{-0.03cm}\\

\makecell{CoTr \\ (0.8476)} &
\makecell{\includegraphics[width=0.4\linewidth]{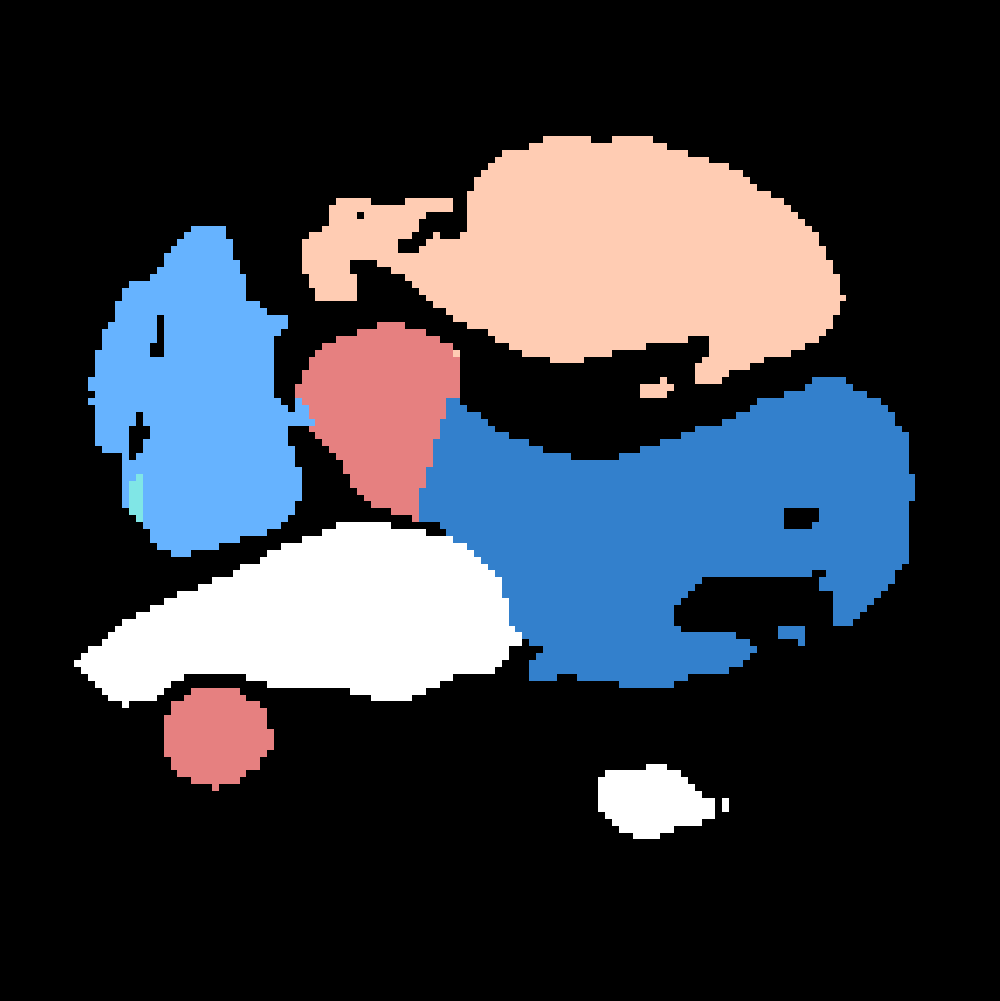}} &
\makecell{\includegraphics[width=0.4\linewidth]{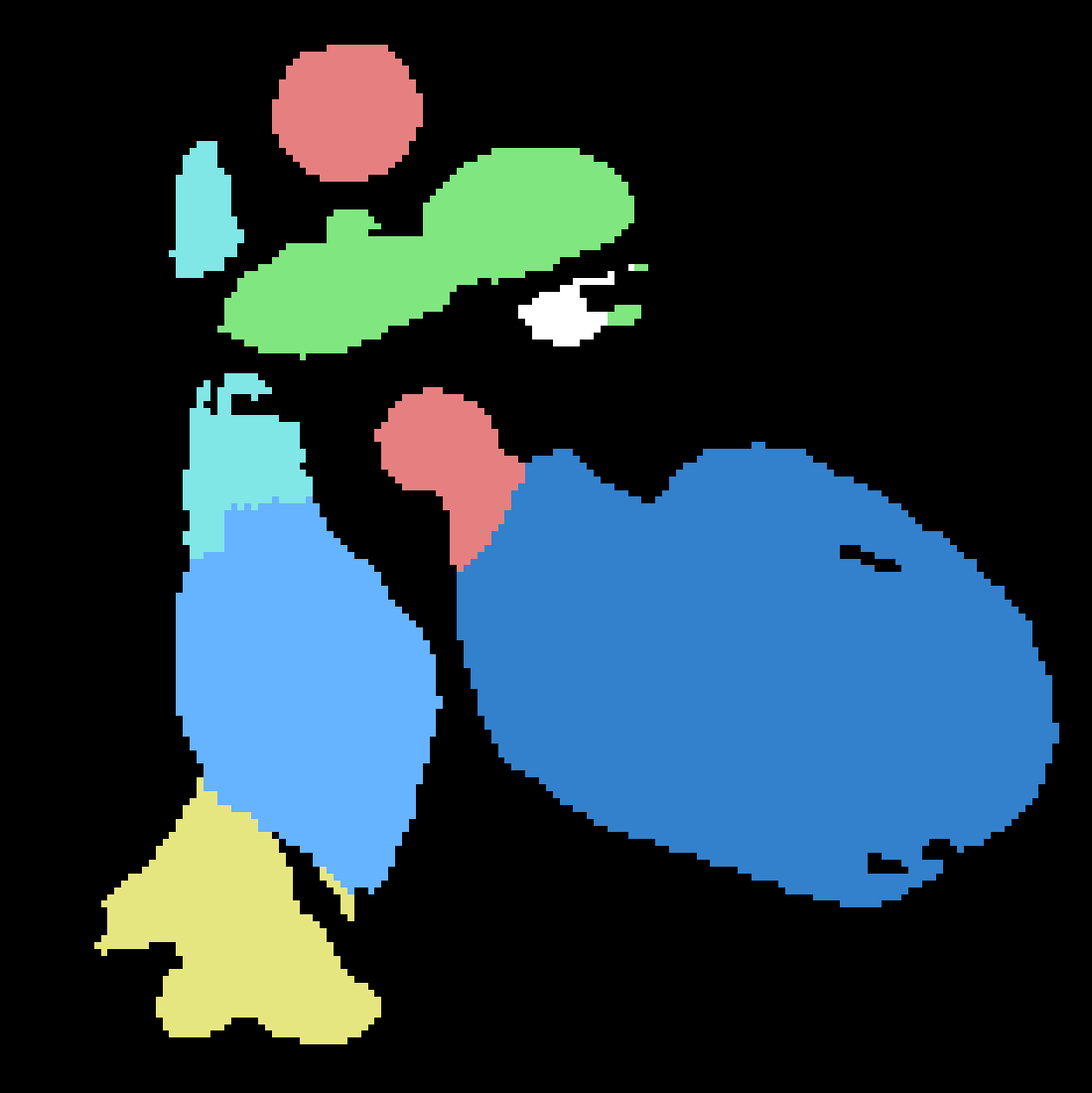}} &
\makecell{\includegraphics[width=0.4\linewidth]{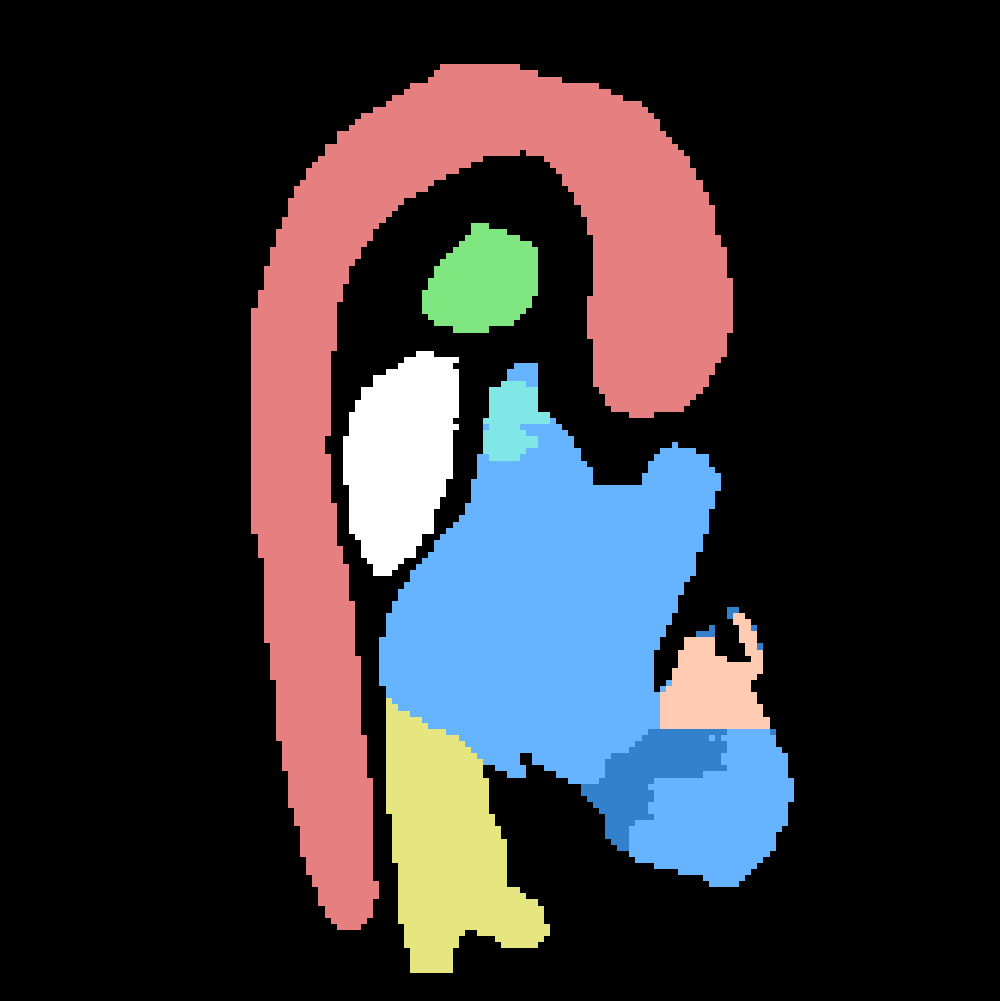}} &
\makecell{\includegraphics[width=0.4\linewidth]{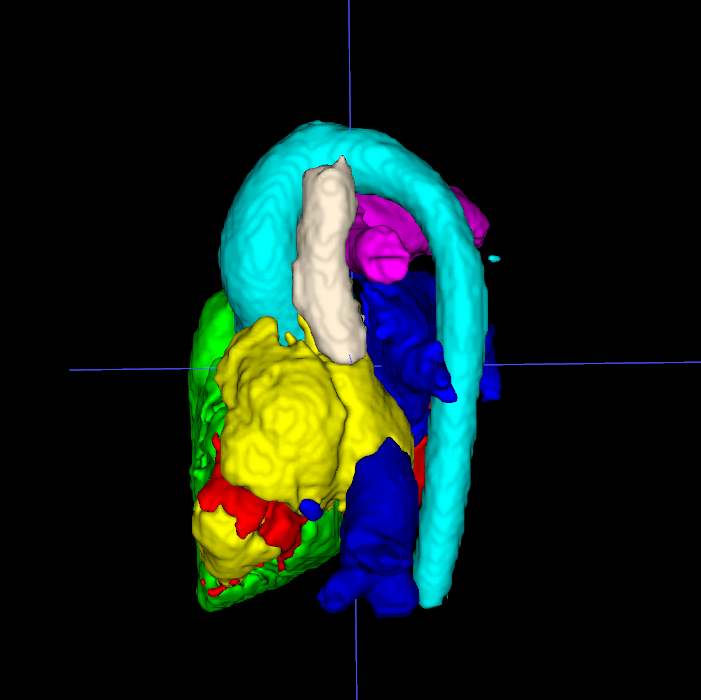}} 
\vspace{-0.03cm}\\

\end{tabular}
\end{minipage}
\vspace{-0.1cm}
\caption{Qualitative comparison for different models on image with ID “pat11” from HVSMR-2.0 dataset.}
\vspace{-0.1cm}
\label{fig:hvsmr_cases}
\end{figure}

\begin{figure}[!t]
\centering
\includegraphics[width=\linewidth]{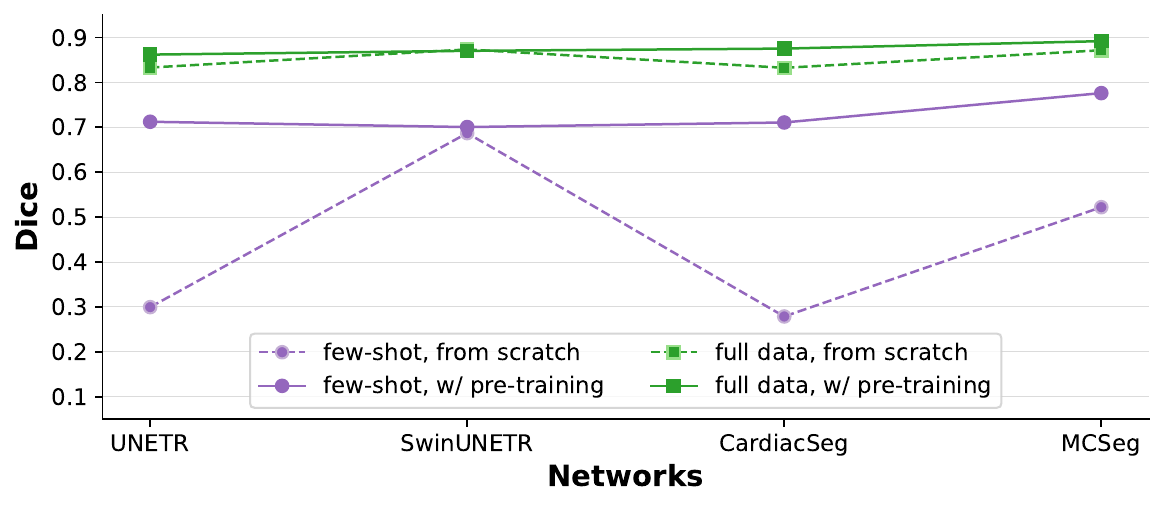}
\vspace{-0.6cm}
\caption{Comparison of models training with and without pre-training weights under few-shot and full-data settings on ImageCHD dataset.}
\label{fig:fewshot}
\vspace{-0.3cm}
\end{figure}

\textbf{Few-shot Experiment.}
The few-shot training set was constructed by randomly selecting eight samples from the original training set, while the test set remained unchanged.
Fig.~\ref{fig:fewshot} compares four transformer-CNN hybrid networks. Among them, only SwinUNETR employs a Swin Transformer as the encoder, while the others utilize ViT encoders. Leveraging pre-trained ViT weights markedly improves few-shot performance: compared to training from scratch, UNETR, CardiacSeg, and MCSeg achieve relative Dice score improvements of 43\%, 46\%, and 58\%, respectively.
MCSeg benefits the most from pre-training, achieving the highest Dice scores in both full-data ($0.896 \pm 0.024$) and few-shot ($0.776 \pm 0.065$) settings, with the smallest performance drop between the two.
Moreover, MCSeg's performance under the few-shot setting is the closest to its full-data result.
A case study is illustrated in Fig. \ref{fig:case_fewshot}, where MCSeg generates segmentations most closely aligned with the ground truth, whereas other models exhibit varying degrees of missegmentation.
These results highlight the efficacy of pre-training in boosting MCSeg's robustness and enhanced adaptability to limited data scenarios.

\subsubsection{Whole-heart Segmentation on HVSMR-2.0 Dataset}
As shown in Table~\ref{tab:exp_hvsmr}, MCSeg achieves the highest overall Dice score ($0.832 \pm 0.071$), outperforming competing methods by 1.4\%. Notably, MCSeg yields the best or second best performance on substructures LV, RV, AO, PA, and IVC, with low standard deviations, indicating high accuracy and consistency. 
Fig.~\ref{fig:hvsmr_cases} presents a case study, in which we observe that while other models exhibit varying degrees of missegmentation across different views, the segmentation result of MCSeg closely aligns with the ground truth. 
These results demonstrate that MCSeg shows a greater advantage over other models on MRI images compared to CT images.

\begin{figure}
\centering
    \begin{subfigure}{\linewidth}
    \centering
        \includegraphics[width=\linewidth]{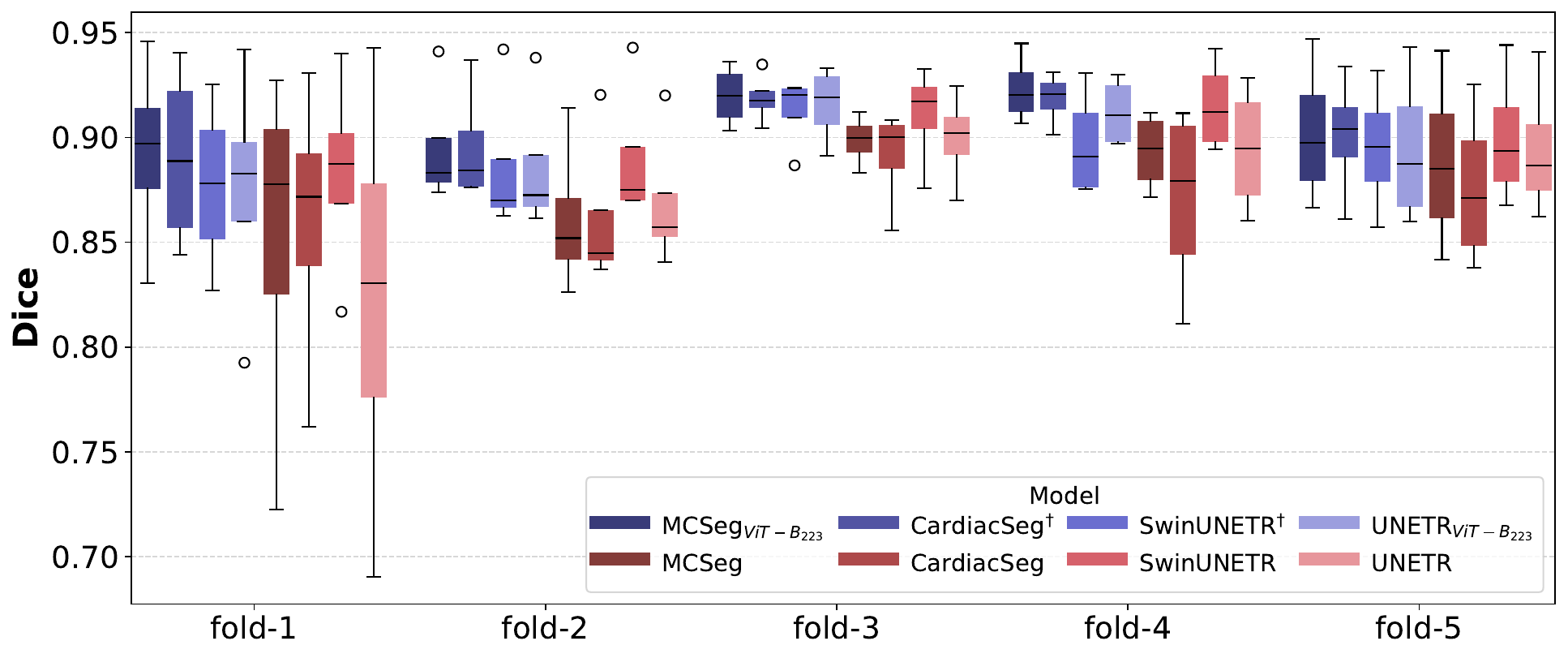}
        \vspace{-0.55cm}
        \caption{Whole-heart segmentation on MM-WHS CT dataset.}
        \label{sfig:5f_mmwhs_ct}
        \vspace{0.15cm}
    \end{subfigure}
    \begin{subfigure}{\linewidth}
    \centering
        \includegraphics[width=\linewidth]{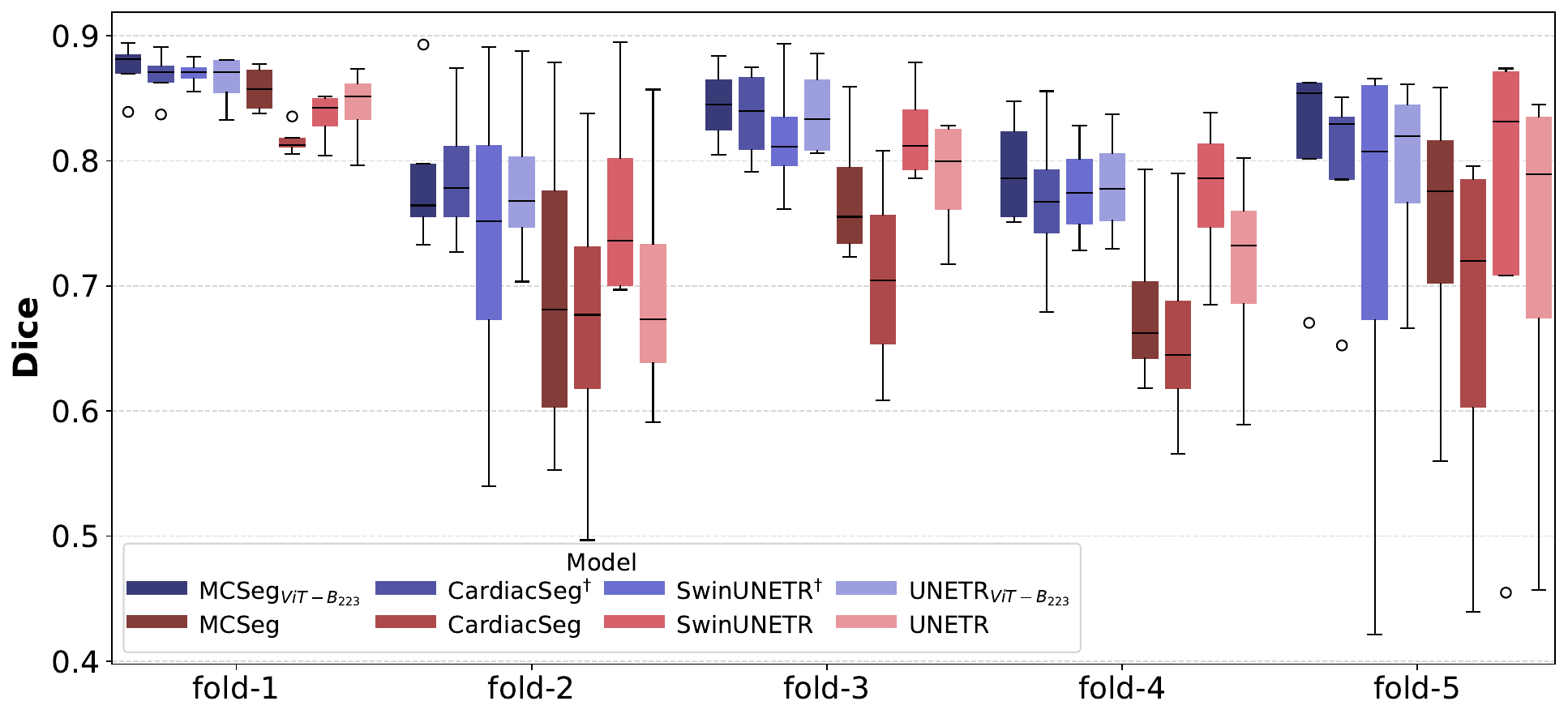}
        \vspace{-0.55cm}
        \caption{Whole-heart segmentation on MM-WHS MRI dataset.}
        \vspace{0.15cm}
        \label{sfig:5f_mmwhs_mri}
    \end{subfigure}
    \begin{subfigure}{\linewidth}
    \centering
        \includegraphics[width=\linewidth]{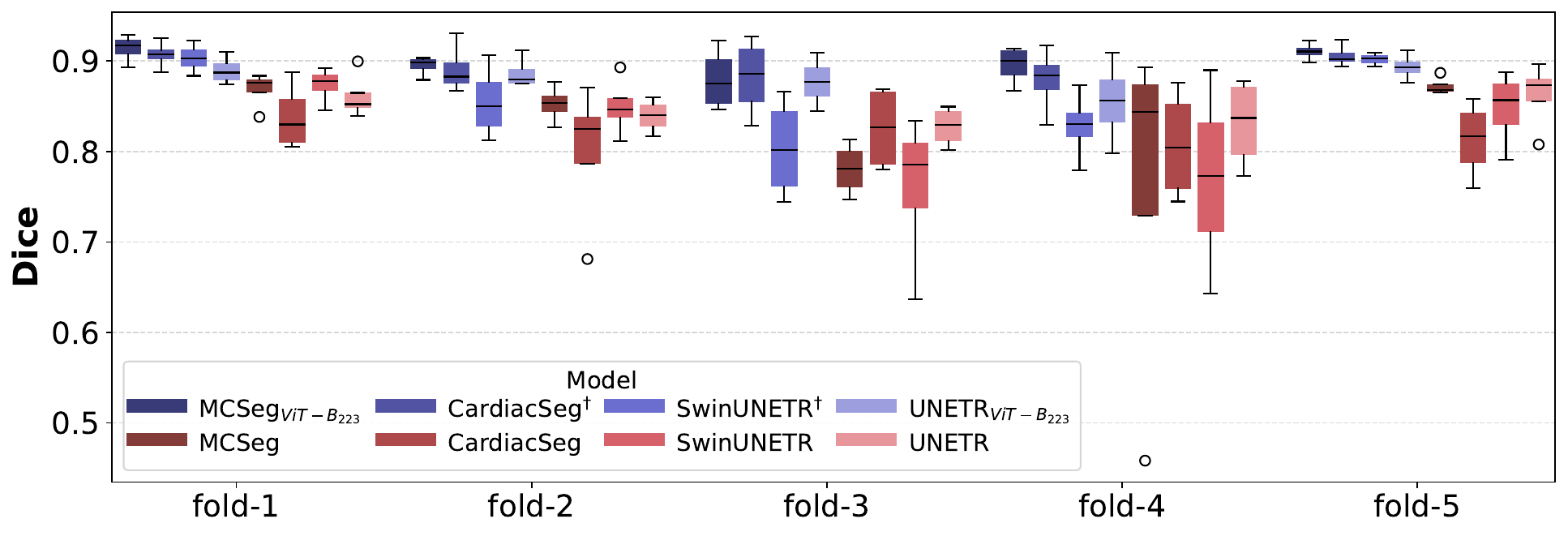}
        \vspace{-0.55cm}
        \caption{LA segmentation on MSD Heart dataset.}
        \label{sfig:5f_msd}
    \end{subfigure}
\caption{Results of 5-fold cross-validation for eight models on various datasets. Models in reds are trained from scratch.}
\end{figure}

\subsubsection{Whole-heart Segmentation on MM-WHS Dataset}
\label{sssec:zeroshot}
MM-WHS dataset includes both CT and MRI modalities. For each modality, we conducted five-fold cross-validation using UNETR, SwinUNETR, CardiacSeg, and MCSeg, trained both from scratch and with pre-trained weights. 
The statistic analyses are shown in Fig.~\ref{sfig:5f_mmwhs_ct}\&\ref{sfig:5f_mmwhs_mri}. As illustrated, MCSeg$_{ViT-B_{223}}$ achieves the highest Dice scores and the lowest intra-group variance in nearly all folds. 
Remarkably, despite the distinct image distributions of CT and MRI images, the ViT-B pre-trained solely on CT images yields even greater performance improvements on MRI segmentation than on CT itself. 
Case studies are provided in supplementary materials.

\begin{figure}[!t]
\centering
\begin{minipage}{\linewidth}
    \centering
    \tabcolsep=0.02cm
    \footnotesize
    \begin{tabular}{ccccc}
    \label{fig: MSD_case}
    GT & MCSeg & CardiacSeg & SwinUNETR & UNETR
    \\ \vspace{-0.03in}
    \makecell[c]{\includegraphics[width=0.17\linewidth]{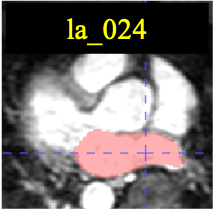}} & \makecell[c]{\includegraphics[width=0.17\linewidth]{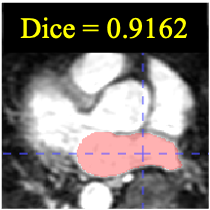}} & \makecell[c]{\includegraphics[width=0.17\linewidth]{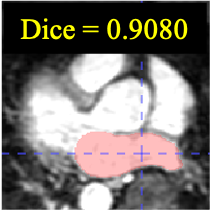}} & \makecell[c]{\includegraphics[width=0.17\linewidth]{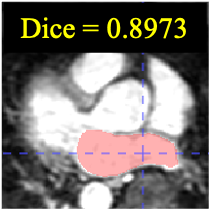}} & \makecell[c]{\includegraphics[width=0.17\linewidth]{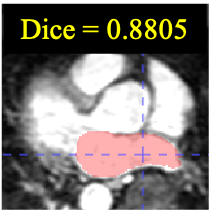}}
    \\ \vspace{-0.03in}
    \makecell[c]{\includegraphics[width=0.17\linewidth]{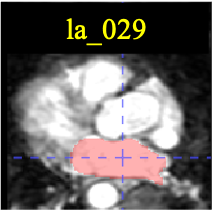}} & \makecell[c]{\includegraphics[width=0.17\linewidth]{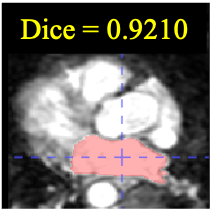}} & \makecell[c]{\includegraphics[width=0.17\linewidth]{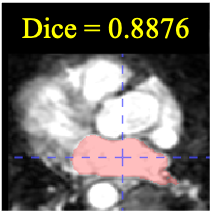}} & \makecell[c]{\includegraphics[width=0.17\linewidth]{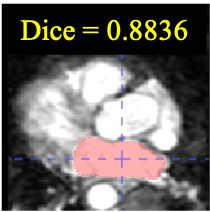}} & \makecell[c]{\includegraphics[width=0.17\linewidth]{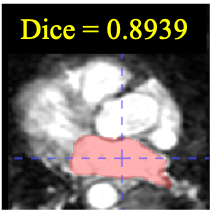}}
    \\
\end{tabular}
\end{minipage}
\caption{Qualitative comparisons for four models on cases ``la\_024" and ``la\_029" from MSD Heart dataset.}
\label{fig:MSD_cases}
\end{figure}

\subsubsection{Left Atrium Segmentation on MSD Heart Dataset}
\label{sssec:la_seg}
Although this task is less challenging than whole-heart segmentation, the statistical analysis presented in Fig.~\ref{sfig:5f_msd} clearly demonstrates the advantages of MCSeg and its pre-trained weights, which are even more apparent compared to the MM-WHS experiments. 
This indicates that MCSeg effectively leverages the capabilities of the ViT backbone obtained during pre-training.
These results highlight the robust performance of MCSeg across tasks of varying difficulty. Furthermore, as shown in the case study Fig.~\ref{fig:MSD_cases}, a qualitative comparison of segmentation results on two representative images further confirms that MCSeg consistently outperforms other models.

\begin{table}[!t]
\caption{Comparison of GPU memory usage during training, numbers of parameters and average inference time per case for various networks in ImageCHD experiments.}
\label{table:Params}
    \centering
    \resizebox{0.48\textwidth}{!}{
    \begin{tabular}{l|c|cc|c}
        \toprule
        \multirow{2}*{\makecell[c]{\textbf{Networks}}} & \multirow{2}*{\makecell{\textbf{GPU Memory} \\ \textbf{Usage (G)}}} &\multicolumn{2}{c|}{\textbf{\#(Params) (M)}}& \multirow{2}*{\makecell{\textbf{Inference} \\ \textbf{Time (s)}}}\\ 
        ~ & ~ & Trainable& Total  & \\
        \midrule
        SAM-Med3D$^{\dag}$~\cite{wang2024sam} & 43.4 & 100.5 & 100.5 & 70.7 \\
        nnFormer~\cite{zhou2023nnformer} & 59.7 & 37.6 & 37.6 & 42.7 \\
        VT-UNet$^{\dag}$~\cite{VT-UNet} & 20.1 & 11.8 & 11.8 & 36.8 \\
        nnUNet~\cite{isensee2021nnu} & 20.2 & 31.2 & 31.2 & 25.4 \\
        TransUNet~\cite{transunet} & 24.6 & 41.4 & 41.4 & 19.3 \\
        MedNeXt~\cite{roy2023mednext} & 59.2 & 12.0 & 12.0 & 14.4 \\
        UNet++~\cite{zhou2019unetplusplus} & 36.1 & 7.0 & 7.0 & 4.3 \\
        SwinUNETR$^{\dag}$~\cite{tang2022self} & 47.9 & 19.6 & 62.2 & 4.0 \\
        CoTr~\cite{cotr} & 31.0 & 41.9 & 41.9 & 1.9 \\
        UNETR$_{ViT-B_{723}}$~\cite{unetr} & 29.5 & 2.7 & 92.8 & 1.1 \\
        MCSeg$_{ViT-B_{723}}$ (ours) & 28.8 & 7.5 & 97.6 & 1.0 \\
        \bottomrule
    \end{tabular}}
\end{table}

\subsubsection{Computational Efficiency}
Table~\ref{table:Params} summarizes the number of network parameters, GPU memory consumption during training, and the average inference time per case for MCSeg and all baseline models. Both GPU memory usage and inference time were measured on ImageCHD with an input size of $(128, 128, 128)$. As shown, MCSeg requires fewer trainable parameters, uses less GPU memory, and achieves the shortest inference time among all compared methods.
\subsection{Ablation Studies}
\label{ssec:Ablations}
To optimize the performance of MCSeg, we conducted a series of ablation experiments towards configuration of SFP, the effects of data scale and modality used for pre-training, fine-tuning strategy, and the combination of loss function.

\begin{table}[!t]
\caption{Ablation studies of (a) pyramid design and (b) SFP settings on ImageCHD dataset.}
\centering
\begin{subtable}{\linewidth}
    \centering
    \vspace{-0.15cm}
    \caption{}
    \vspace{-0.15cm}
    \label{stab:abl_fpn}
    \renewcommand\arraystretch{1.2}
    \begin{tabular}{c|c}
        \toprule
        \textbf{Pyramid Design}&  \textbf{Dice}\\
        \midrule
        None (baseline) & $0.876 \pm 0.037$\\
        FPN~\cite{lin2017fpn} & $0.877 \pm 0.042$\\
        SFP (ours)& $\mathbf{0.892 \pm 0.033}$\\
        \bottomrule
    \end{tabular}
    \vspace{0.15cm}
\end{subtable}
\\
\begin{subtable}{\linewidth}
    \centering
    \caption{}
    \vspace{-0.15cm}
    \label{stab:abl_net}
    \renewcommand\arraystretch{1.4}
    \begin{tabular}{cc|c}
        \toprule
        \textbf{Features}&\textbf{Pyramid Scale}&  \textbf{Dice}\\
        \midrule
        $(z_{3}, z_{6}, z_{9}, z_{12})$ &$\left(\frac{1}{16}, \frac{1}{8}, \frac{1}{4}, \frac{1}{2}\right)$& $0.876 \pm 0.035$\\
        $z_{12}$ &$\left(\frac{1}{16}, \frac{1}{8}, \frac{1}{4}, \frac{1}{2}\right)$& $0.876 \pm 0.045$\\
        $(z_{3}, z_{6}, z_{9}, z_{12})$& $\left(\frac{1}{32}, \frac{1}{16}, \frac{1}{8}, \frac{1}{4}\right)$&$0.885 \pm 0.036$\\
        $z_{12}$ &$\left(\frac{1}{32}, \frac{1}{16}, \frac{1}{8}, \frac{1}{4}\right)$& $\mathbf{0.892 \pm 0.033}$\\
        \bottomrule
    \end{tabular}
\end{subtable}
\end{table}

\begin{table*}[!t]
\caption{Impact of pre-training data scales and modalities. Downstream segmentation performance is reported alongside pre-training costs (GPU hours and peak memory usage). The first row denotes the baseline model trained from scratch.}
\label{tab:pt_mri}
    \centering
    \resizebox{0.9\textwidth}{!}{
    \renewcommand\arraystretch{1.1}
    \begin{tabular}{l|ccc|ccc|cc}
        \toprule
          \multirow{2}*{\textbf{Model}} & \multicolumn{3}{c|}{\textbf{Data}} &   \multicolumn{3}{c|}{\textbf{Dice}} & \multicolumn{2}{c}{\textbf{Pre-training Cost}}\\
          &CT&   CTA&MRI&   ImageCHD (CT) &HVSMR-2.0 (MRI) &  MM-WHS (MRI) & GPU Hours (h) & Peak Memory Usage (GB)\\
         \midrule
          ViT-B&--& --& --& $0.872 \pm 0.053$& $0.743 \pm 0.127$&$0.752 \pm 0.069$ & -- & -- \\
          ViT-B$_{311}$&223&   --&88& $0.896 \pm 0.025$ &$\mathbf{0.838 \pm 0.061}$ & $0.821 \pm 0.043$ & 108.38 & 69.37 \\
          ViT-B$_{223}$&223&  --&--&  $0.895 \pm 0.027$&$0.832 \pm 0.071$ &$\mathbf{0.822 \pm 0.033}$ & 84.87 & 69.37 \\
          ViT-B$_{723}$&223&  500&--&  $\mathbf{0.896 \pm  0.024}$& $0.789 \pm 0.130$ & $0.817 \pm 0.049$ & 252.62 & 69.37 \\
          ViT-B$_{1223}$&223&  1000&--&  $0.895 \pm 0.022$& $0.825 \pm 0.081$ & $0.806 \pm 0.057$ & 423.08 & 69.37 \\
         \bottomrule
    \end{tabular}}
\end{table*}

\begin{table}[!t]
\caption{Comparison of (a) training strategies on multi-modal datasets with ViT-B$_{223}$, and (b) various loss function combinations on ImageCHD dataset.}
\centering
\begin{subtable}{\linewidth}
    \centering
    \vspace{-0.15cm}
    \caption{}
    \vspace{-0.15cm}
    \label{tab:ft_strategy}
    \renewcommand\arraystretch{1.1}
    \begin{tabular}{l|cc}
        \toprule
        \multirow{2}*{\textbf{Training Strategy}} & \multicolumn{2}{c}{\textbf{Dice}}\\
         &  ImageCHD&  HVSMR-2.0\\
         \midrule
         From scratch & $0.872 \pm 0.053$ &  $0.743 \pm 0.127$\\
         Fine-tuning (ViT unfrozen)& $0.888 \pm 0.033$ &  $0.831 \pm 0.058$\\
         Fine-tuning (ViT frozen)& $\mathbf{0.892 \pm 0.033}$ &  $\mathbf{0.832 \pm 0.071}$\\
         \bottomrule
    \end{tabular}
    \vspace{0.15cm}
\end{subtable}
\\
\begin{subtable}{\linewidth}
    \centering
    \caption{}
    \vspace{-0.15cm}
    \label{tab:abl_loss}
    \begin{tabular}{ccc|cc}
         \toprule
         $L_{dice}$& $L_{ce}$&  $L_{rmi}$& \textbf{Dice}$\uparrow$  & \textbf{HD95}$\downarrow$\\
         \midrule
         \Checkmark& \XSolidBrush & \XSolidBrush & $0.854 \pm 0.046$ & $19.0 \pm 10.0$\\
         \Checkmark&  \Checkmark& \XSolidBrush & $0.885 \pm 0.036$ &$14.3 \pm 5.16$\\
         \Checkmark& \XSolidBrush &  \Checkmark& $0.862 \pm 0.052$ &$17.2 \pm 6.17$\\
         \Checkmark&  \Checkmark&  \Checkmark& $\mathbf{0.892 \pm 0.033}$ & $\mathbf{13.4 \pm 5.35}$\\
         \bottomrule
    \end{tabular}
\end{subtable}
\end{table}

\subsubsection{SFP Design}
Table~\ref{stab:abl_fpn} demonstrates that our proposed SFP effectively bridges the encoder and decoder, outperforming the conventional FPN~\cite{lin2017fpn} by 1.5\%. 
Finally, we investigated the feature source and scaling configurations within the SFP (Table~\ref{stab:abl_net}). Constructing the multi-scale pyramid exclusively from the deepest ViT block ($z_{12}$) alongside a scaling configuration of $\left(\frac{1}{32}, \frac{1}{16}, \frac{1}{8}, \frac{1}{4}\right)$ yields the optimal performance. This indicates that the deepest representations ($z_{12}$) contain more comprehensive semantic information, serving as a robust foundation for multi-scale decoding compared to fusing intermediate block outputs.

\subsubsection{Data Scale and Modalities for Pre-training}
Table~\ref{tab:pt_mri} illustrates the impact of varying pre-training data scales and modalities on downstream segmentation performance and computational costs. Utilizing pre-trained weights consistently achieves significant performance gains over training from scratch, improving Dice scores by over 2.3\% on ImageCHD and ranging from 7.0\% to 9.5\% on two MRI datasets. Considering both performance gains and computational costs, ViT-B$_{223}$ provides the optimal trade-off between multi-modal segmentation performance and pre-training efficiency.

\subsubsection{Fine-tuning Strategy}
Table~\ref{tab:ft_strategy} compares three training strategies for MCSeg. Fine-tuning with a frozen ViT encoder achieves the highest Dice scores on both ImageCHD and HVSMR-2.0 datasets, proving to be the most effective strategy for downstream adaptation.

\subsubsection{Loss Function}
Table~\ref{tab:abl_loss} evaluates various loss combinations. The integration of Dice, CE, and RMI losses achieves the highest Dice score (0.892) and the lowest HD95 (13.4), confirming that RMI loss effectively refines boundary accuracy beyond the standard Dice and CE combination.

\subsection{Discussion}
Our experimental results demonstrate that MCSeg outperforms SOTA methods on four datasets. Notably, pre-training solely on CT images yields substantial performance gains on both CT and MRI datasets. This cross-modality transferability stems from their shared anatomical structures. Notably, MRI tasks, which often suffer from high variability due to different scanners and protocols, benefit tremendously from the robust anatomical priors learned from CT. Conversely, incorporating vatious modalities like CTA or limited MRI into pre-training provides negligible or inconsistent benefits, largely due to cross-modality domain gaps and insufficient data scales.

Regarding model architecture, we eliminated the pre-training variant by applying our pre-trained ViT weights to UNETR. The consistent superiority of MCSeg over UNETR strictly isolates and validates the efficacy of our SFP module in bridging the encoder-decoder gap. Furthermore, compared to SwinUNETR and CardiacSeg, our approach validates that a domain-specific, large-scale pre-training paradigm is practical for multi-modal success.

Whereas large-scale 3D MAE pre-training yields substantial improvements, we acknowledge its practical limitations regarding computational and memory demands. As detailed in Table~\ref{tab:pt_mri}, pre-training requires hundreds of GPU hours on advanced hardware (\textit{e.g.}, 80G A100s). To ensure broad accessibility and fully support reproducibility for typical users with constrained resources, we have open-sourced our pre-trained ViT weights, the fine-tuned downstream model weights, and the complete source code, allowing researchers and clinicians to deploy the model out-of-the-box. 

Furthermore, a conceptual limitation is that our pre-trained weights are explicitly cardiac-specific. In an era drawn to ``universal'' medical foundation models designed for all-around tasks, a specialized focus might seem restrictive. Yet, from a practical standpoint, we argue that this anatomical specialization is precisely what makes the model reliable and readily deployable in real-world clinical pathways, where targeted precision typically outweighs broad but shallow capabilities.

\section{Conclusion}
\label{sec:conclusion}
In this study, we propose MCSeg, a volumetric transformer-based framework for multi-modal cardiac image segmentation. By integrating a 3D ViT-B/16 encoder pre-trained via masked autoencoders with a specifically designed SFP, and incorporating RMI loss during fine-tuning, MCSeg effectively captures both global anatomical context and fine-grained boundary details. Whereas large-scale pre-training significantly elevates the performance floor, the SFP and RMI loss further raise the ceiling of segmentation precision.

Extensive evaluations confirm that MCSeg consistently surpasses existing SOTA methods on both CT and MRI datasets. Furthermore, its exceptional performance in few-shot scenarios underscores the model's generalization capabilities and adaptability when faced with scarce annotated data. 

Ultimately, beyond providing an accurate tool for cardiac segmentation, this study establishes a effective clinical pipeline. By coupling organ-specific large-scale pre-training with a tailored downstream architecture, MCSeg offers a robust and transferable paradigm for developing clinical-grade medical image analysis solutions.

\section*{References}
\bibliographystyle{IEEEtran}
\bibliography{IEEEabrv, ref}



\end{document}